\pdfoutput=1
\documentclass[11pt]{article}
\usepackage[utf8]{inputenc}
\usepackage{times}
\usepackage{longtable}
\usepackage{algorithm}
\usepackage[noend]{algorithmic}
\usepackage{xspace}
\usepackage{nicefrac}
\usepackage{booktabs}
\usepackage{footmisc}
\usepackage[shortlabels]{enumitem}
\usepackage[margin=1in]{geometry}
\usepackage{comment}
\usepackage{booktabs,tabularx}
\usepackage{multirow}
\usepackage{textcomp}
\usepackage[numbers]{natbib}
\usepackage{graphicx}
\usepackage{color}
\usepackage[hidelinks]{hyperref}
\usepackage{wrapfig}
\usepackage{amssymb,amsmath,amsthm}
\usepackage[capitalise,noabbrev]{cleveref}

\usepackage{mathtools}
\usepackage{array}
\usepackage{url}
\usepackage[strict]{changepage}
\usepackage{makecell}
\usepackage{titlesec}
\usepackage{setspace}
\usepackage{bm}
\usepackage{mdframed}
\usepackage{bbm}
\usepackage{threeparttable}
\usepackage{subfigure}

\usepackage[T1]{fontenc}

\usepackage{titletoc}
\usepackage{longtable}
\usepackage{pifont} 
\usepackage{wrapfig}

\usepackage{fancyhdr} % for fancy headers and footers
\usepackage{xcolor} % for color definitions
\usepackage[most]{tcolorbox} % for colored boxes

\usepackage[table]{xcolor}  
\usepackage{colortbl}  

\PassOptionsToPackage{table}{xcolor}

\usepackage{wasysym}
\usepackage{booktabs}
\usepackage{algorithm}
\usepackage{algorithmic}
\usepackage[switch]{lineno}
\usepackage{colortbl}
\usepackage{listings}
\usepackage{tikz}
\usepackage{xparse}
\usepackage{enumitem}
\usepackage{fontawesome5}
\usepackage{natbib}
\definecolor{mygreen}{RGB}{30, 128, 20}
\colorlet{shadecolor}{gray!20}

\setcitestyle{round}
\tikzset{chatstyle/.style={text width=2.8in,rounded corners=2pt}}

\definecolor{mygreen}{HTML}{88EABB}
\definecolor{OliveGreen}{HTML}{00693E}

\usepackage{wrapfig}
\usepackage{adjustbox}
\usepackage{xspace}
\usepackage{listings}
\usepackage[most]{tcolorbox}

\usepackage{pifont}

\usepackage{multicol}
\usepackage{multirow}
\usepackage{bbding}
\usepackage{circledsteps}

\usepackage{fancyhdr}
\usetikzlibrary{mindmap,shadows}

\usepackage{colortbl}
\usepackage{xcolor}
\definecolor{LightCyan}{RGB}{232,241,255}
\definecolor{LightRed}{RGB}{255,235,235}
\definecolor{LightPink}{RGB}{255,235,255}
\definecolor{LightGreen}{RGB}{218,255,234}
\definecolor{LightYellow}{RGB}{255,255,235}
\definecolor{LightGray}{RGB}{242,242,242}
\definecolor{Red}{RGB}{253, 239, 242}
\definecolor{Yellow}{RGB}{255, 255, 204}
\definecolor{Pink}{RGB}{255, 243, 254}
\definecolor{Gray}{RGB}{249, 249, 249}
\definecolor{Green}{RGB}{230, 255, 241}
\definecolor{Blue1}{RGB}{218, 232, 245}
\definecolor{Blue2}{RGB}{239, 248, 253}
\definecolor{Blue3}{RGB}{136, 190, 220}
\definecolor{Blue4}{RGB}{83, 157, 204}
\definecolor{Blue5}{RGB}{42, 122, 185}
\definecolor{Blue6}{RGB}{11, 85, 159}
\definecolor{GreenCheck}{RGB}{0, 102, 51}
\definecolor{LightBack}{RGB}{247,249,251}
\definecolor{babyblueeyes}{rgb}{0.63, 0.79, 0.95}
\tcbset{
    colback=LightBack,colframe=black,boxsep=3pt, arc=3pt,boxrule=1pt,
    width=\linewidth,enhanced,frame hidden, overlay={\draw[line width=1pt] (frame.north west)--([xshift=\linewidth]frame.north west);\draw[line width=1pt] (frame.south west)--([xshift=\linewidth]frame.south west);}
}

\usepackage{pifont}
\usepackage{graphicx}
\usepackage{array}
\usepackage{tabularx}

\usepackage[table]{xcolor}   
\usepackage{graphicx}     
\usepackage{longtable} 
\usepackage{booktabs}  
\usepackage{multirow}        

\usepackage{array}
\usepackage{ragged2e}
\usepackage{booktabs}
\usepackage{url}
\usepackage{hyperref}

\hypersetup{
    colorlinks=true,
    linkcolor=black, 
    filecolor=black, 
    urlcolor=blue,  
    citecolor=black 
}

\def\@onedot{\ifx\@let@token.\else.\null\fi\xspace}

\newcommand{\header}[1]{\noindent\textbf{{#1.}}~~}

\titleformat*{\subparagraph}{\itshape}

\newsavebox\actorsfigure

\makeatletter
\def\@maketitle{%
  \newpage
  \null
  \vskip 2em%
  \hrule height 3pt  % upper
  \vskip 1.5em%
  \begin{center}%
    {\LARGE \@title \par}%
  \end{center}%
  \vskip 1em%
  \hrule height 1pt  % lower
  \vskip 2em%
  \begin{center}%
    {\large
      \lineskip .5em%
      \begin{tabular}[t]{c}%
        \@author
      \end{tabular}\par}%
    \vskip 1em%
    {\large \@date}%
  \end{center}%
  \par
}
\makeatother

\title{Digital Twin AI: Opportunities and Challenges from Large Language Models to World Models}

\author{
Rong Zhou$^{1}$\thanks{Major contribution.} \and
Dongping Chen$^{2*}$ \and
Zihan Jia$^{3*}$ \and
Yao Su$^{4*}$ \and
Yixin Liu$^{1*}$ \and
Yiwen Lu$^{5*}$ \and
Dongwei Shi$^{1*}$ \and
Yue Huang$^{6*}$ \and
Tianyang Xu$^{7*}$ \and
Yi Pan$^{8}$ \and
Xinliang Li$^{8}$ \and
Yohannes Abate $^{8}$ \and
Qingyu Chen$^{9}$ \and
Zhengzhong Tu$^{10}$ \and
Yu Yang$^{1}$ \and
Yu Zhang$^{11}$ \and
Qingsong Wen$^{12}$ \and
Gengchen Mai$^{13}$ \and
Sunyang Fu$^{14}$ \and
Jiachen Li$^{15}$ \and
$^{}$ \and
Xuyu Wang$^{16}$ \and
Ziran Wang$^{17}$ \and
Jing Huang$^{5,18}$ \and
Tianming Liu$^{8}$ \and
$^{}$ \and
Yong Chen$^{5}$\thanks{Yong Chen, Lichao Sun and Lifang He are co-corresponding authors.}\and
Lichao Sun$^{1\dagger}$ \and
Lifang He$^{1\dagger}$ \and \\
\small{$^{1}$Lehigh University\quad $^{2}$University of Maryland\quad $^{3}$University of New South Wales\quad} \\
\small{$^{4}$Worcester Polytechnic Institute\quad $^{5}$University of Pennsylvania\quad  $^{6}$University of Notre Dame\quad} \\
\small{$^{7}$Columbia University\quad $^{8}$University of Georgia\quad $^{9}$Yale University\quad} \\
\small{$^{10}$Texas A\&M University\quad $^{11}$Stanford University\quad $^{12}$Squirrel Ai Learning} \\
\small{$^{13}$University of Texas at Austin\quad $^{14}$University of Texas Health Science Center at Houston\quad} \\
\small{$^{15}$University of California, Riverside\quad $^{16}$Florida International University\quad $^{17}$Purdue University}\\
\small{$^{18}$Children’s Hospital of Philadelphia}
}

\date{}

\begin{document}

% This addressed the weird gap in authors list, caused by Ayfer's name.
\begin{spacing}{1.1}
\maketitle
\vspace{-1.2em}
\end{spacing}

\begin{abstract}
Digital twins, as precise digital representations of physical systems, have evolved from passive simulation tools into intelligent and autonomous entities through the integration of artificial intelligence technologies. This paper presents a unified four-stage framework that systematically characterizes AI integration across the digital twin lifecycle, spanning modeling, mirroring, intervention, and autonomous management. By synthesizing existing technologies and practices, we distill a unified four-stage framework that systematically characterizes how AI methodologies are embedded across the digital twin lifecycle: (1) modeling the physical twin through physics-based and physics-informed AI approaches, (2) mirroring the physical system into a digital twin with real-time synchronization, (3) intervening in the physical twin through predictive modeling, anomaly detection, and optimization strategies, and (4) achieving autonomous management through large language models, foundation models, and intelligent agents. We provide an in-depth analysis of the synergy between physics-based modeling and data-driven learning, highlighting the transition from traditional numerical solvers to physics-informed and foundation models for physical systems. Furthermore, we examine how generative AI technologies, including large language models and generative world models, transform digital twins into proactive and self-improving cognitive systems capable of reasoning, communication, and creative scenario generation. Through extensive review across eleven application domains such as healthcare, aerospace, smart manufacturing, robotics, and smart cities, we identify both universal challenges including scalability, explainability, and trustworthiness, as well as domain-specific requirements. This paper reveals how AI-driven digital twins are evolving toward more intelligent, interoperable, and ethically responsible ecosystems, highlighting key directions for future interdisciplinary research and development.
% All resources, including benchmarks and tools referenced in this paper, will be made available at [repository link].
\end{abstract}

\pagebreak

\begin{small}
\tableofcontents
\end{small}

\setlength{\parskip}{0.5em}

\pagebreak

\section{Introduction}
% include google map as an example
% include agent part
% all figure：overview, history, concept, world model 2025, applications
\textit{"What I can’t create, I don’t understand."}

\hfill -- \textit{Richard Feynman}

\noindent Digital twins (DT), as precise digital representations of physical twins (real-world entities or systems), are meticulously designed to maintain a bidirectional connection with their real-world systems, enabling state synchronization for monitoring, prediction, optimization, and decision support~\cite{KREUZER2024102304}. 
Beyond mere replication, digital twins embody a paradigm shift from static digital mirrors to dynamic, continuously learning reflections of reality. 
Due to the inherent advantages of predictive analytics, dynamic system simulation, and operational optimization that digital twins provide, this technology has been widely used for healthcare~\cite{erol2020digital, SiemensHealthineers_2018}, biological domain~\cite{hengelbrock2024digital, cheng2024digital}, urban planning and management~\cite{deng2021creating, liu2020digital}, manufacture~\cite{lee2019digital, twin2020predictive}, and science~\cite{jones2020characterising, Mihai2022Digital, Menon2023Digital,segovia2022design}. 
As NVIDIA's founder and CEO, Jensen Huang, stated in a keynote at the Berlin Summit for the Earth Virtualization Engines initiative, AI and accelerated computing will revolutionize our understanding of complex systems~\cite{jensen}, highlighting a new era where digital twins evolve from analytical tools into intelligent agents that learn, predict, and act upon the physical world.

\noindent Technically, a digital twin operates by integrating comprehensive sensor data from a studied object, such as a wind turbine, where sensors are strategically placed to monitor crucial performance metrics including energy output, temperature, and weather conditions. This data is continuously transmitted to a processing system, which applies it to a virtual replica of the physical object. Utilizing this up-to-date digital model, various simulations can be conducted to analyze performance issues and devise potential improvements. The ultimate goal of this process is to extract insightful knowledge from the simulations, which can then be applied to enhance the real-world object, optimizing its efficiency and functionality. A more familiar example of a digital twin is Google Map, which fuses satellite imagery, GPS data, and real-time traffic to maintain a constantly updated mirror of the physical world ~\cite{liu2021review}. This continuous feedback loop between sensing, modeling, and adaptation forms the conceptual foundation of AI-driven digital twins—systems that not only reflect the world but also learn from it to guide real-world actions.
% . By utilizing satellite imagery, street view data, and other geo-information, Google Maps constructs a virtual replica of the physical world on your mobile device. Furthermore, through sensors like GPS and real-time traffic data, it continuously updates this virtual model. The system then integrates these data inputs to offer optimal route suggestions, which are relayed back to the user via screen display or voice navigation. Although this is a relatively simple implementation compared to full-fledged industrial digital twins, it clearly illustrates the core principle: establishing a bidirectional flow between the physical world and its digital counterpart~\cite{liu2021review}.

\noindent Over the years, the concept of digital twins has evolved significantly, increasingly integrating with artificial intelligence (AI) breakthroughs to transform how we simulate and predict the behaviors of physical systems~\cite{huang2021survey}. Digital twins and machine learning (ML) are closely intertwined, enhancing predictive maintenance and decision-making across various industries. Early systems relied on traditional machine learning algorithms to support predictive maintenance and fault detection~\cite{smith2020predictive, jardine2006review}. However, as data volumes and system complexity expanded, deep learning (DL) emerged as the cognitive core of digital twins, empowering them to extract intricate spatiotemporal patterns and emulate complex dynamics. Architectures such as convolutional~\cite{lecun1998gradient}, recurrent~\cite{hochreiter1997long}, and graph neural networks~\cite{kipf2016semi,gilmer2017neural} have enabled digital twins to move from modeling observed behavior to reasoning about unobserved mechanisms. This integration marks a fundamental transition: AI is no longer merely a component within digital twins but the intelligence that animates them.
\begin{figure}[H]
    \centering
\includegraphics[width=\linewidth]{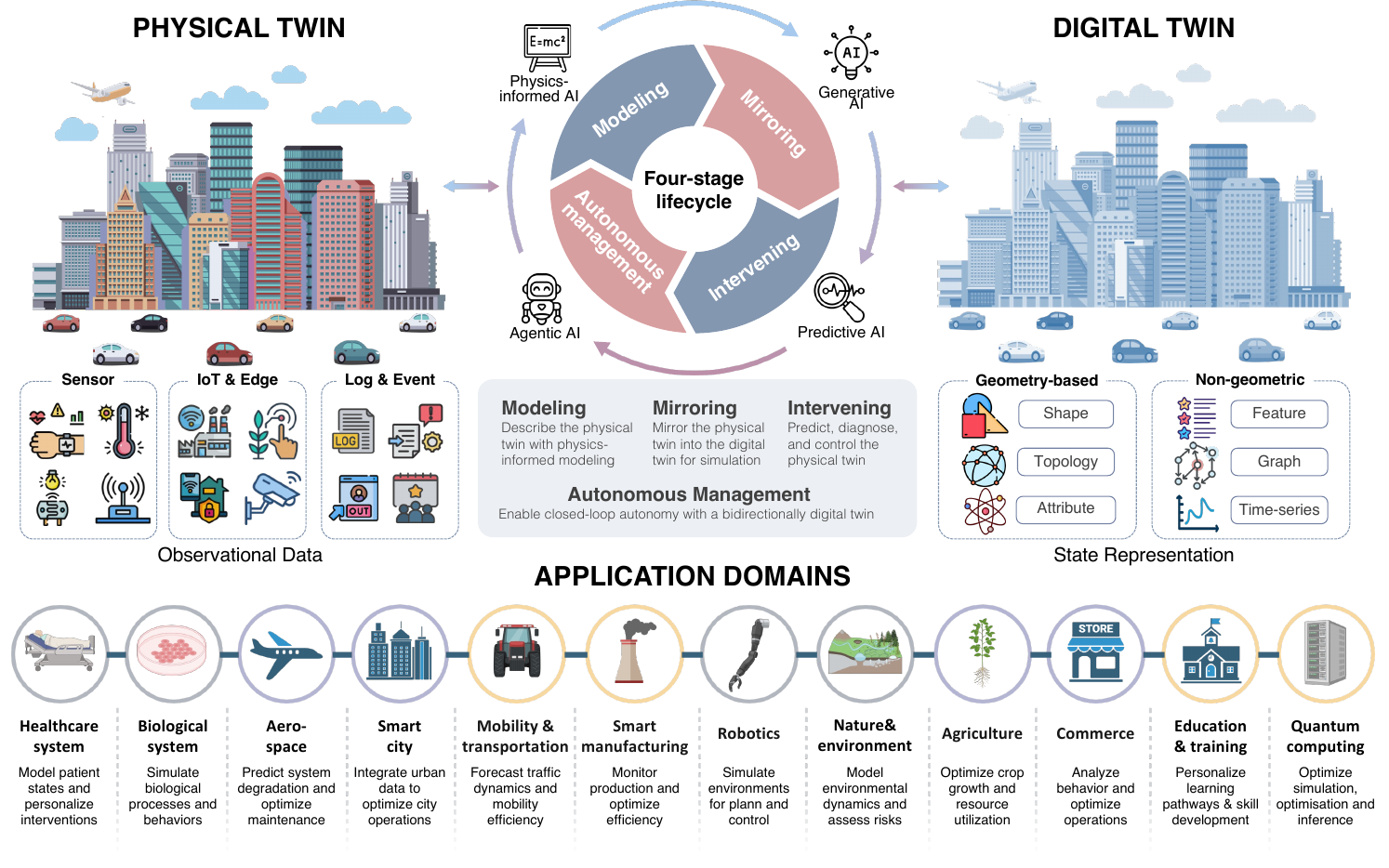}
    \caption{\textbf{AI-driven digital twin framework and application landscape.}
The four-stage lifecycle conceptualizes digital twins as evolving intelligent systems:
\textbf{First}, describing the world (\textbf{physical twin}) via physics-informed AI and observational data.
\textbf{Second}, mirroring the world into synchronized digital simulators (\textbf{digital twin}) through generative AI.
\textbf{Third}, intervening in the world with predictive AI for forecasting, diagnosis, and optimization.
\textbf{Ultimately}, achieving autonomous management of the world via agentic AI powered by large language models and foundation models.
This conceptual framework generalizes across a wide range of application domains.}
    \label{fig:intro}
\end{figure}

\noindent With the advent of large-scale AI models and foundation architectures, the synergy between AI and digital twins has entered an unprecedented phase~\cite{cao2023comprehensive, liu2023deidgpt}. Recent breakthroughs, from LLM-based autonomous agents~\cite{huang2024metatool, liu2023agentbench} to world models~\cite{liu2024sora}, demonstrate how AI can emulate, reason, and even imagine complex physical systems.
For instance, a multi-agent system framework that applies an LLM is utilized to automate the parametrization of process simulations in digital twins~\cite{xia2024llmexperimentssimulationlarge}. Moreover, NVIDIA Cosmos provides a world foundation model that generates photorealistic synthetic environments for digital-twin simulations in robotics and autonomous systems~\cite{agarwal2025cosmos}.

\noindent The convergence of AI technology and digital twins promises to not only enhance the fidelity and responsiveness of these virtual models but also to redefine the boundary between simulation and intelligence. Augmented by adaptive learning and generative reasoning, AI-driven digital twins can anticipate faults before they occur, personalize interventions, and autonomously manage complex systems. Such capabilities herald a future where digital twins evolve into trustworthy, explainable, and human-aligned partners in science, industry, and healthcare.~\cite{emmert2023role}.

\noindent Given this transformation, there is an urgent need to consolidate knowledge across the rapidly diversifying landscape of AI-powered digital twins. 
This paper provides a comprehensive, AI-centered overview of digital twin technologies. We begin by tracing the history of digital twins to establish the conceptual foundation. We then present a four-stage lifecycle that organizes how AI empowers digital twins: modeling the physical twin through physics-based methods and data integration, mirroring it into executable simulators, intervening through prediction, anomaly detection, and optimization, and ultimately achieving autonomous management via large language models and intelligent agents.
Given the diverse applications of digital twins, we will examine how AI technology enhances their implementation across different domains such as healthcare, biological systems, and industry. Lastly, we will discuss the existing challenges and issues in using AI technology for digital twins and offer insightful recommendations for future research directions.

\subsection{Major Contributions}

To the best of our knowledge, this paper provides the AI-centered conceptual synthesis of digital twins as evolving intelligent systems. Unlike prior domain-specific reviews, we present a unified framework that connects the physical, digital, and cognitive layers of this emerging paradigm. The contributions of this paper are summarized as follows.

\begin{itemize}[noitemsep,topsep=0pt,leftmargin=0.13in]

\item We conceptualize digital twins as evolving AI systems, distilling a four-stage lifecycle: describing the physical twin, mirroring the physical twin to digital twin, intervening in the physical twin, and autonomously managing the physical twin. This layered perspective reveals how AI continuously enhances the fidelity, intelligence, and autonomy of digital twins.

\item We provide an in-depth analysis of the integration between physics-based modeling and data-driven learning, highlighting the transition from traditional numerical methods to physics-informed neural networks, neural operators, and foundation models for physical systems. This synthesis clarifies how physical principles and learning algorithms can jointly improve interpretability, generalization, and reliability in digital twin modeling.

\item We analyze the rapid development of generative AI, including large language models, diffusion models, and world simulators, and examine their role in enabling reasoning, communication, and imagination within digital twins. These technologies transform digital twins from passive simulation tools into proactive, self-improving cognitive systems capable of autonomous understanding and creative scenario generation.

\item Through extensive review across eleven application domains, we identify both universal challenges such as scalability, explainability, and trustworthiness, as well as domain-specific requirements in areas including healthcare, aerospace, energy, and education. These observations reveal how AI-driven digital twins are evolving toward more intelligent, interoperable, and ethically responsible ecosystems, highlighting key directions for future exploration and interdisciplinary collaboration.
\end{itemize}

\subsection{Organization}
% polisg
To guide readers through this interdisciplinary synthesis, this paper is organized into seven main sections. Section~\ref{sec:2}, \textit{History of Digital Twins}, reviews the conceptual origins and technological evolution of digital twin systems. Sections~\ref{sec:3}–\ref{sec:6} form the methodological core of this work, presenting a progressive framework that models, mirrors, intervenes, and autonomously manages the physical world through its digital counterpart. Specifically, Section~\ref{sec:3}, \textit{Modeling the Physical Twin}, describes how physical systems are represented through physics-based and data-driven modeling. Section~\ref{sec:4}, \textit{Mirroring the Physical Twin into the Digital Twin}, explains how these models are instantiated and visualized within virtual simulators. Section~\ref{sec:5}, \textit{Intervening in the Physical Twin via the Digital Twin}, focuses on predictive modeling, anomaly detection, and optimization techniques that enable human-in-the-loop decision-making. Section~\ref{sec:6}, \textit{Towards Autonomous Management of the Digital Twin}, advances this paradigm toward AI-driven autonomy, highlighting large language models, foundation models, and intelligent agents as enablers of self-managing digital twins. Finally, Section~\ref{sec:7}, \textit{Applications}, demonstrates how these methodological principles are applied across diverse domains such as healthcare, aerospace, smart manufacturing, and robotics, illustrating the broad impact of digital twin technologies in real-world systems. Finally, Section~\ref{sec:8}, \textit{Open Challenges and Future Directions}, discusses the key open problems and outlines future research directions for building scalable, trustworthy, and autonomous digital twin systems.

\section{History of Digital Twins}\label{sec:2}
\noindent The concept of the digital twin was formally introduced in 2002 by Michael Grieves during a presentation at the University of Michigan. This presentation emphasized the establishment of a product lifecycle management center that integrated both real and virtual spaces, along with data flows to enhance efficiency and innovation in product development and management~\cite{Grieves2006, Gonzalez2021, Wikipedia2024}. While the terminology around digital twins has evolved over the years, the fundamental idea of merging digital and physical twins has remained constant.

\noindent Interestingly, the practice of digital twinning dates back to the 1960s, long before the term was coined. NASA was among the early adopters, using basic forms of digital twins for space missions. One notable example is the Apollo 13 mission, where simulations using digital twin concepts played a critical role in bringing the crew safely back to Earth~\cite{nasa_digital_twins}. These early applications demonstrated the potential of digital twins in enhancing design, maintenance, and operational efficiency across various industries.

\noindent Companies like Rolls-Royce have been pioneers in adopting digital twin technology. They have utilized digital twins to customize repair processes for engine parts, automating and optimizing maintenance practices based on the specific geometries of these components~\cite{RollsRoyceDigitalTwin}. Similarly, in the aerospace industry, Boeing employed digital twins for the 787 Dreamliner’s battery systems to improve safety and manage risks more effectively~\cite{Boeing787DigitalTwin}. Airbus has also embraced this technology, using digital twins for their A350 XWB aircraft to enable real-time performance monitoring, which has led to substantial improvements in fuel efficiency and reductions in emissions~\cite{AirbusA350DigitalTwin}.

\noindent Recent advancements in digital twin technology have continued to drive significant innovations across multiple industries. In manufacturing, healthcare, construction, automotive, and urban planning, digital twins are becoming indispensable tools. For instance, Tesla has harnessed digital twins to accelerate vehicle development, while the Mayo Clinic has used them to advance personalized medicine~\cite{startus2024, mckinsey2024, matterport2024, hcltech2024}. Additionally, architects and city planners are leveraging digital twins to enhance project management and urban development, integrating AI and IoT to optimize resource use and improve outcomes.

% \section{Artificial Intelligence Technology in Digital Twins}\label{sec:3}
\section{Modeling the Physical Twin}\label{sec:3}
% world model (genie2 feifei lab)
% agent AI
% physical AI
In today’s technological landscape, the integration of Artificial Intelligence (AI) with digital twins is gaining widespread attention and driving diverse applications. 
% The integration of AI technologies enhances digital twins with greater intelligence and autonomy. 
AI technologies have the potential to enhance the intelligence and autonomy of digital twins. 
For example, compared to traditional methods based on physics, the physics-informed AI system can significantly improve digital twins in numerous ways, such as automating the modeling process~\cite{wang2023simultaneous} and improving computational efficiency~\cite{yang2024data}. Furthermore, by learning from sensor data of physical systems and simulation data, AI can provide more effective and efficient predictions~\cite{huang2021survey} and fault detection~\cite{booyse2020deep}. In addition, in the past two years, generative AI~\cite{bordukova2024generative,mariam2024unlocking}  and Large Language Models (LLMs)~\cite{yang2024llm, xia2024llm, vsturm2024enhancing} have profoundly impacted digital twins, particularly in tasks related to simulation. The ongoing advancements in AI continue to push the boundaries of what digital twins can achieve, heralding a new era of smart, interconnected systems across various industries.

\subsection{Physical System Modeling}
% solve PDE with numerical methods and uncertainty quant：1. discover pde 2. solve pde (pinn & neural operators)
% Table Physics-Based Methods and AI Systems
Physics knowledge has long been a foundation in traditional digital twins~\cite{boschert2016digital,liu2021review,tao2022digital}, providing essential tools for understanding and predicting complex systems through mathematical representations of physical laws. Physics-informed AI effectively addresses a range of limitations inherent in traditional physics-based methods by combining AI techniques with physical knowledge~\cite{kantaros20213d,cheatham2000molecular,agalianos2020discrete}. In this section, we review physics-based methods used in digital twins, highlight their limitations, and summarize new insights from recent physics-informed AI models.

% Figure 2 Physics-Based Methods and AI Systems. Three figures should be explained.
\begin{figure}[H]
    \centering
\includegraphics[width=0.9\linewidth]{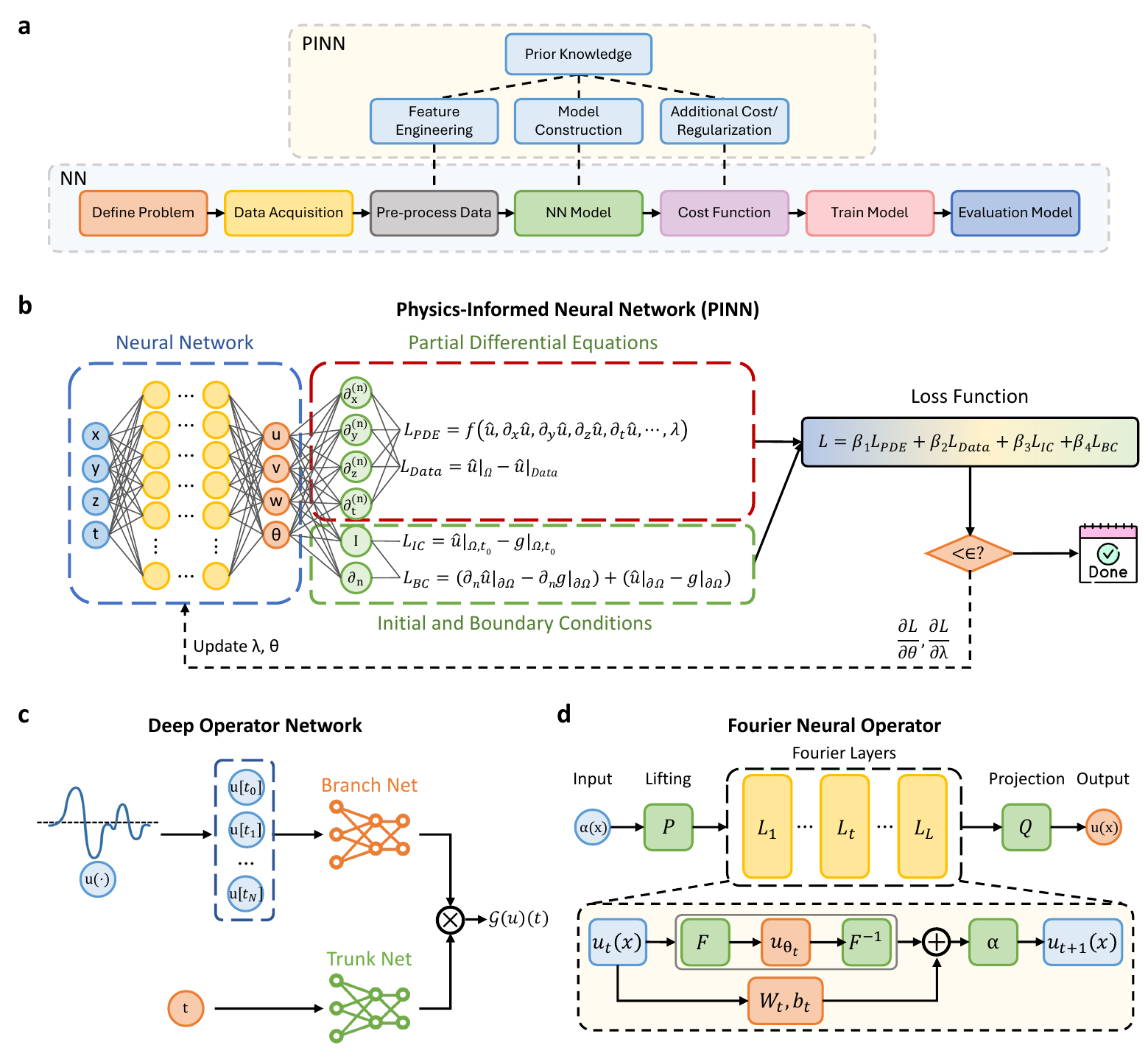}
    \caption{Physics-Based Methods and AI Systems.\textbf{(a)} Workflow of Physics-Informed Neural Networks (PINNs), integrating prior knowledge into the learning pipeline through regularization and domain constraints. \textbf{(b)} Architecture of PINNs incorporating data, PDEs, and boundary/initial conditions into a unified loss function. \textbf{(c)} Deep Operator Network (DeepONet) structure, modeling nonlinear operators via separate branch and trunk networks. \textbf{(d)} Fourier Neural Operator (FNO) framework, leveraging Fourier transforms for efficient learning of solution operators in PDE problems.}
    \label{fig:PM}
\end{figure}

\subsubsection{Fundamental Physics-Based Methods}
Before the advent of AI, researchers predominantly relied on physics-based methods for tasks such as simulation, prediction, analysis, and control in digital twins, with computational outcomes derived from numerical methods. 
In this section, we will introduce several important fields in these paradigms, beginning with constructing Partial Differential Equations (\textbf{PDEs})~\cite{molinaro2021embedding,thomas2021cfd}, followed by solving these PDEs through \textbf{numerical methods}~\cite{tuegel2012airframe,gockel2012challenges}, and finally evaluating the error through \textbf{uncertainty quantification}~\cite{wang2015use,zhang2017backward}.

\header{Partial Differential Equations (PDE)}
Computational physics has long been a foundational discipline in building traditional simulations for digital twins, focusing primarily on numerical solutions for PDEs. For instance, in constructing simulators for fluid phenomena, it is often necessary to solve the Navier-Stokes equations~\cite{slotnick2014cfd}. These equations have been extensively utilized in simulators designed for aerospace~\cite{li2021digital}, aircraft~\cite{tuegel2012airframe,gockel2012challenges}, weather~\cite{voosen2020europe}, and oil pipelines~\cite{wanasinghe2020digital}, all of which involve fluid dynamics. Similarly, in solid physics and materials science, heat equations are used to simulate the heat conduction process, and plasticity equations are used to simulate stress variations within materials. These simulations are then applied in the creation of virtual twins for architecture~\cite{al2021digital}, manufacturing~\cite{knapp2017building}, and other industries~\cite{yu2022energy,debroy2017building,wagg2020digital}. In addition, reaction-diffusion equations~\cite{britton1986reaction} are commonly used to model diffusion processes of substances or signals, such as biochemical reactions~\cite{dobrzynski2007computational} and tumor growth~\cite{gatenby1996reaction}. These equations play a crucial role in the simulation of digital twins, especially within the health industry and for biological entities~\cite{gerach2021electro,wu2022integrating,moller2021digital}. In addition to describing natural phenomena, PDEs are also commonly used to model various social phenomena. For instance, the spread of infectious diseases is often described by the Susceptible-Infectious-Recovered model~\cite{alrashed2022covid}; traffic flow models are employed to represent traffic conditions and manage congestion prediction~\cite{karafyllis2022analysis}. In economics and finance, investment strategies~\cite{hu2023application} and option pricing~\cite{barles1998option} are frequently described using the Hamilton-Jacobi-Bellman equation and the Black-Scholes equation, respectively.

\header{Numerical Methods} 
To accurately obtain numerical solutions for equations in computational physics, it is necessary to perform a discretization process. This is done by dividing the domain into a discrete grid and subsequently approximating the solutions at these grid points~\cite{dahlquist2003numerical}. Based on this strategy, several numerical methods have been developed, each tailored to meet specific simulation challenges. The finite difference method~\cite{liszka1980finite} simplifies implementation by using the difference of function values to approximate derivatives at grid points; however, it often lacks precision and efficiency~\cite{verzicco1996finite}. Spectral methods~\cite{shen2011spectral}, on the contrary, represent the solution globally using different basis functions, offering high resolution (i.e., spectral accuracy~\cite{trefethen2000spectral}). Despite their high precision, spectral methods are generally suitable only for relatively regular domains, limiting their applicability to other downstream tasks~\cite{montanari2015limitation}. The finite element method (FEM)~\cite{huebner2001finite} segments the domain into various small elements, such as triangles or tetrahedrons, and approximates the solution within each using low-order polynomials. Due to its flexible meshing strategy, FEM can adapt to a wide range of scenarios and complex real-world conditions~\cite{rao2017finite, hinchy2020using, tao2022digital, moussa2018insights}. A notable example of FEM applied in simulation is the ICOsahedral Nonhydrostatic (ICON) model~\cite{jungclaus2022icon}, which initially discretizes the Earth into an icosahedron consisting of 20 triangular faces, facilitating the calculation of numerical solutions. The ICON model demonstrates the power of FEM by converting a set of partial differential equations, which describe various weather conditions, into algebraic equations and solving them using a supercomputer. This approach has been integrated into the Earth2 system by NVIDIA.

\header{Uncertainty Quantification}
Simulation in digital twins aims to achieve a precise one-to-one correspondence between a physical system and its virtual representation~\cite{wagg2020digital}, demonstrating the importance of quantifying the uncertainties inherent in both physical measurements~\cite{hausmann2021managing} and computational models~\cite{kennedy2001bayesian,thelen2023comprehensive,lin2021uncertainty}. The primary techniques used for quantifying uncertainty and its propagation in simulators include Monte Carlo methods~\cite{rubinstein2016simulation,leira2019utilization}, Bayesian Inference~\cite{box2011bayesian,nguyen2023direct}, and Sensitivity Analysis~\cite{christopher2002identification,casal2023sensitivity}. Monte Carlo methods involve random sampling of input parameters to create a distribution of possible outcomes, enabling estimation of output uncertainties. It is particularly effective in handling complex and high-dimensional problems, which are further enhanced with variance reduction techniques such as importance sampling~\cite{tokdar2010importance} and stratified sampling~\cite{neyman1992two}, focusing on the most critical parts of the input space~\cite{rubinstein2016simulation, owen2013monte}. Bayesian Inference, on the other hand, uses Bayes' Theorem~\cite{joyce2003bayes} to update the probability distribution of model parameters based on prior knowledge and new data. This approach provides a systematic framework for incorporating uncertainty into model predictions, making it possible to refine the model as more data becomes available. Techniques like Markov Chain Monte Carlo~\cite{fernandez2016ggmcmc} are crucial in this context, as they allow the practical application of Bayesian inference by approximating posterior distributions in high-dimensional spaces~\cite{gelman2013bayesian}. Moreover, Sensitivity Analysis evaluates how variations in input parameters affect model outputs, and identifies the key parameters that significantly influence the results. Global sensitivity analysis methods~\cite{iooss2015review}, such as the Sobol index~\cite{owen2014sobol}, offer a comprehensive understanding of how input uncertainties propagate through the model, thus highlighting the most influential parameters and guiding efforts to reduce uncertainty~\cite{saltelli2008global}.

\subsubsection{Physics-Informed AI Models}
In recent years, the rapid advancement of AI technologies has introduced innovative methodologies to address modeling and computation challenges in traditional physics-based methods. One approach involves using AI to explicitly extract underlying PDEs from data for further modeling process~\cite{brunton2020machine,long2018pde,cranmer2020discovering,kaheman2021pde}. Alternatively, another strategy involves embedding some or all of the known physical knowledge directly into AI models~\cite{raissi2019physics,lu2019deeponet,li2020fourier}. Additionally, AI-assisted computation can reduce computational costs by enhancing the numerical computation steps with AI technologies~\cite{zhang2020machine,rojek2021ai}. This integration streamlines processes and boosts efficiency in modeling and computation tasks.

\header{PDEs Discovery for Modeling} In the traditional digital twins modeling process, a significant challenge arises when certain physical laws are unclear or only partially understood, making it difficult to establish accurate mathematical models~\cite{rasheed2020digital,tao2019digital,glaessgen2012digital}. This situation is particularly common in complex systems, such as turbulence~\cite{duraisamy2019turbulence}, multiphase flows~\cite{prosperetti2009computational}, and materials science~\cite{curtarolo2013high}. To address these modeling challenges, researchers have begun to explore the use of AI techniques to directly learn underlying physical knowledge and models from data. One of the earlier and highly influential works is by Brunton et al.~\cite{brunton2016discovering}, who introduced the Sparse Identification of Nonlinear Dynamics (SINDy) method, which can discover governing equations from time-series data. This algorithm is based on the assumption that physical laws are often simple, leading to sparsity in the indices. Using the concept of the Koopman Operator~\cite{brunton2021modern}, SINDy transforms the problem of finding low-dimensional governing equations into a high-dimensional linear regression problem. A series of subsequent studies have expanded the applicability of SINDy to a wider range of scenarios, including the discovery of PDEs~\cite{rudy2017data}, handling noisy data~\cite{schaeffer2017sparse}, handling multiscale physics~\cite{champion2019data}, and jointly processing control inputs~\cite{kaiser2018sparse}. In particular, recent research has begun to utilize SINDy to construct simulators for digital twins in various application domains, such as manufacturing~\cite{wang2022time}, chemical engineering~\cite{wang2023simultaneous}, and other industrial scenarios. 
More recently, researchers have continued to advance the field of PDE discovery from data by utilizing deep learning. Early influential works include the numerical method-based PDE-Net~\cite{long2018pde} and the symbol regression-based approaches~\cite{bar2019learning} followed two parallel trajectories. Further studies, such as PDE-Net 2.0~\cite{chen2022pde}, have mixed numerical methods with symbolic approaches. Additionally, a series of works~\cite{kaheman2021pde,champion2019data,kaheman2020sindy} have combined deep learning techniques to further enhance the applicability and dimensionality of the SINDy algorithm.

\header{Solving PDEs for Simulation} Besides utilizing AI to explicitly uncover the underlying physical knowledge and mathematical models from data, another approach is to embed some or all of the known physical knowledge based on PDEs into AI models, aiming to combine the flexibility of data-driven methods with the interpretability of physical models to solve PDEs. Following this guidance, the most well-known work is the Physics-Informed Neural Networks (PINNs)~\cite{raissi2019physics}. By directly incorporating PDEs into the loss function as penalty terms, PINNs allow physical information to constrain the neural network outputs to some extent.
\begin{equation}
\mathcal{L}_{\text{PINN}} = \lambda_{\text{data}} \mathcal{L}_{\text{data}} + \lambda_{\text{PDE}} \mathcal{L}_{\text{PDE}}
\end{equation}
where \(\lambda_{\text{data}}\), \(\lambda_{\text{physics}}\) are weights that balance the contributions of each term. The loss term for PDEs often consists of two independent parts: one part satisfies the physical equations, and the other part satisfies the boundary conditions.
\begin{equation}  \mathcal{L}_{\text{PDE}}=\mathcal{L}_{\text{physics}}+\mathcal{L}_{\text{boundary}}=\frac{1}{M} \sum_{j=1}^{M} \left( \mathcal{N}[u](x_j, t_j) \right)^2+\frac{1}{P} \sum_{k=1}^{P} \left( \mathcal{B}[u](x_k, t_k) - g(x_k, t_k) \right)^2
\end{equation}
where \( \mathcal{N}[u](x_j, t_j) \) is the residual of the PDE at collocation points \((x_j, t_j)\). And \( \mathcal{B}[u] = g \) is the boundary condition. %(M and P should be defined). 
Subsequently, researchers developed PINN variants, each enhancing PINNs' performance and applicability from different perspectives. Variational PINNs~\cite{kharazmi2019variational} introduced a variational formulation to improve training stability and accuracy, Conservative PINNs~\cite{jagtap2020conservative} ensured the conservation of physical quantities, and Adaptive PINNs~\cite{jagtap2020adaptive} employed adaptive activation functions to enhance learning efficiency. Probabilistic PINNs~\cite{meng2020ppinn} incorporated probabilistic models to quantify uncertainty in the predictions. PINNs and their variants have been applied in a wide range of simulation domains, such as fluid dynamics~\cite{cai2021physics}, structural mechanics~\cite{haghighat2021physics}, and biomedical engineering~\cite{kadeethum2020physics}.

Currently, Neural Operators have emerged as another class of methods for embedding physical knowledge. The earliest work is DeepONet~\cite{lu2019deeponet}, which leveraged the universal approximation theorem to directly learn differential operators in PDEs, rather than the solutions to PDEs. This approach sparked the concept of Neural Operators. Further advancements led to the Fourier Neural Operator (FNO), which significantly improved performance by filtering out high-frequency information through multiple Fourier layers. An even more advanced development based on the FNO structure is the FourCastNet~\cite{pathak2022fourcastnet}, which is used for weather forecasting and has been integrated into the Earth2 system. Other variants based on FNO include Adaptive FNO~\cite{guibas2021adaptive}, Multiwavelet Fourier Feature Operators (MWFF)~\cite{gupta2021multiwavelet}. Both Neural Operators and PINNs focus the majority of computational effort during the model training phase. Once trained, these physics-informed AI models can generate predictions at remarkable speeds, often orders of magnitude faster than conventional numerical solvers~\cite{karniadakis2021physics}. This shift in computational paradigm offers significant advantages in scenarios requiring repeated simulations or real-time predictions. For instance, Lu et al.~\cite{lu2021learning} demonstrated that their physics-informed DeepONet could solve partial differential equations up to 1000 times faster than traditional numerical methods. Similarly, Hennigh et al.~\cite{hennigh2021nvidia} showed that AI-based turbulence models could accelerate CFD simulations by up to two orders of magnitude. This dramatic reduction in inference time not only enables rapid what-if analyses and design optimizations but also opens up new possibilities for real-time control and decision-making in complex systems~\cite{willard2020integrating}. However, it is important to note that the training process for these AI models can be computationally intensive and may require significant amounts of data or carefully designed loss functions incorporating physical constraints. However, the potential for rapid and accurate predictions makes physics-informed AI models an increasingly attractive option for a wide range of tasks for digital twins.

\subsection{Observational Data Integration}

A key challenge in modeling the physical twin is ensuring that real-world observations can be effectively incorporated into the model. This process unfolds in two steps. Section~3.2.1 addresses \textit{acquisition and alignment}, where heterogeneous data from sensors, IoT devices, or logs are cleaned, synchronized, and transformed into consistent observational evidence. Section~3.2.2 then focuses on \textit{data assimilation}, the stage where these prepared observations are combined with the model to update its states and parameters. In other words, acquisition and alignment ensure that the data are trustworthy and comparable, while assimilation ensures that the model itself adapts to the incoming evidence. Together, they form the methodological bridge that keeps the physical twin closely tied to the evolving reality.

\subsubsection{Acquisition and Alignment}

The construction of digital twins begins with the acquisition of observational data, which forms the bridge between physical systems and computational models. In practice, data originate from diverse sources, and their heterogeneity introduces challenges of noise, missing values, inconsistent sampling, and semantic mismatches. Acquisition and alignment methods aim to transform these disparate inputs into clean, reliable, and interoperable sequences that can serve as evidence for downstream modeling and assimilation.

\header{Sensor Data}
The most direct form of observational data is produced by physical sensors that measure variables such as temperature, pressure, vibration, voltage, location, audio, or video. Raw sensor signals are rarely usable without processing, as they are often affected by noise, baseline drift, or intermittent dropout. Signal processing techniques such as low-pass filtering and detrending are standard approaches for mitigating these issues \cite{chen2016signal}. Anomaly detection is also necessary to identify faulty measurements or unexpected spikes, which may otherwise corrupt the dataset. A comprehensive survey of statistical and machine learning methods for anomaly detection is provided by Hodge and Austin \cite{hodge2004survey}. Missing values caused by sensor dropout are typically reconstructed through interpolation or imputation methods, ranging from classical statistical approaches to probabilistic models \cite{little2019statistical}. Together, these preprocessing steps convert raw sensor feeds into stable sequences suitable for integration.

\header{IoT and Edge Data}
Beyond dedicated sensors, observational data are increasingly collected through the Internet of Things (IoT), which refers to distributed devices connected via wireless communication networks such as Wi-Fi, Bluetooth, Zigbee and LTE/5G. IoT infrastructures generate vast quantities of heterogeneous data, ranging from environmental readings to user interactions \cite{atzori2010internet}. Their communication is usually based on lightweight protocols such as MQTT or CoAP, designed for constrained environments. However, large-scale IoT deployments raise challenges including latency, packet loss, and inconsistent device configurations. To mitigate these problems, edge computing has emerged as a complementary paradigm: instead of sending all raw data to the cloud, computation is partially performed on gateways or embedded processors close to the data source \cite{shi2016edge}. This strategy reduces bandwidth usage and response time while allowing local preprocessing such as compression or anomaly detection. More recently, the integration of AI models into edge devices has enabled real-time feature extraction and adaptive decision support, a direction often referred to as edge intelligence \cite{zhou2019edge}.

\header{Log and Event Data}
In addition to continuous signals, digital twins incorporate discrete logs and event records that capture system-level behaviors. Logs may include textual records from control systems, operational software, or user interactions. Event data typically represent discrete occurrences, such as failures or maintenance actions. Unlike sensor and IoT data, logs are semi-structured or unstructured and require dedicated parsing methods. Log parsing frameworks such as Drain convert free-text records into structured templates suitable for analysis \cite{he2016experience}. A recurring challenge is inconsistent timestamps across distributed systems, which necessitates synchronization and normalization \cite{li2021log}. Once standardized, logs and events can be serialized and aligned with continuous data streams, enabling their integration into digital twin workflows. These data forms highlight the importance of sequence modeling, anomaly detection, and temporal reasoning in AI-based integration approaches.

\header{Alignment of Heterogeneous Data}
Regardless of source, acquired data streams must be aligned to ensure interoperability. Temporal alignment is the first requirement: sensors may operate at different sampling rates, IoT devices may report intermittently, and logs may record discrete events. Synchronization protocols such as the Network Time Protocol (NTP) establish a common reference clock for distributed devices \cite{mills1991internet}. When perfect synchronization is not possible, interpolation and resampling methods are applied to harmonize signals on a shared timeline. Sliding-window aggregation can further reconcile high-frequency sensor data with lower-frequency records. Beyond time, semantic and unit normalization is necessary to prevent conflicts across heterogeneous datasets. Ontology-based mapping and dictionary-driven label alignment are commonly used for harmonizing variable names, while unit conversion ensures comparability of physical quantities such as temperature or flow rate \cite{noy2001ontology}. Together, these procedures transform heterogeneous sources into coherent datasets ready for integration with digital twin models.

Acquisition and alignment provide the methodological foundation for digital twin data integration. Sensor data, IoT and edge platforms, and log or event systems all contribute valuable but heterogeneous evidence. Through preprocessing, synchronization, and semantic normalization, raw observations are transformed into harmonized sequences that preserve fidelity while achieving interoperability. These steps also highlight opportunities for AI researchers, such as anomaly detection for noisy signals, lightweight models for edge-level feature extraction, and semantic alignment driven by representation learning.

\subsubsection{Data Assimilation}

Once observational data have been collected and aligned, the next step is to integrate them with the model in order to estimate the true system state and calibrate unknown parameters. This process, known as \textit{data assimilation} (DA), provides a principled framework for merging model predictions with observational evidence. In the context of digital twins, DA is critical: without it, the model would drift away from reality due to imperfect initial conditions, incomplete parameterization, or unmodeled disturbances; with DA, the model remains dynamically consistent with the physical twin. Put simply, acquisition and alignment make the data usable, while assimilation makes the model responsive to the data.
Formally, DA can be seen as a Bayesian estimation problem, where the model forecast provides a prior, the observations provide a likelihood, and the assimilation step produces a posterior estimate of the system state. Beyond updating the state, DA also delivers estimates of uncertainty, which are essential for downstream tasks such as prediction, optimization, and decision-making. Over the past decades, DA methods have evolved along three major lines: sequential approaches, variational approaches, and more recently, hybrid and learning-based approaches.

\header{Sequential Data Assimilation}
Sequential methods update the system state step by step as new data arrive. The classical example is the Kalman Filter (KF), which provides an optimal linear estimator under Gaussian assumptions \cite{kalman1960new}. To handle nonlinear dynamics, the Extended Kalman Filter (EKF) approximates the system by local linearization, while the Unscented Kalman Filter (UKF) employs deterministic sampling to capture nonlinear effects more robustly \cite{julier1997new}. For large-scale, high-dimensional systems such as geophysical models, the Ensemble Kalman Filter (EnKF) has become the method of choice \cite{evensen1994sequential}. By representing error statistics with an ensemble of model trajectories, EnKF provides a scalable solution that is widely used in weather forecasting and ocean modeling. Sequential methods are attractive for digital twins because of their ability to process data streams in real time, although they may suffer from sampling errors or loss of variance in very high-dimensional settings.

\header{Variational Data Assimilation}
Variational approaches cast DA as an optimization problem. The objective is to minimize a cost function that balances model fidelity with observational fit over a given time window. In 3D-Var, the optimization is performed at a single analysis time, combining the background forecast with new observations \cite{lorenc1986analysis}. In 4D-Var, the assimilation extends over a temporal window, allowing the system dynamics to constrain the analysis \cite{le1994variational}. These methods exploit all available observations within the assimilation window and yield dynamically consistent state trajectories. They are particularly effective for large-scale applications such as numerical weather prediction. However, their reliance on adjoint models and high-dimensional optimization makes them computationally demanding, which may limit their direct applicability in some digital twin scenarios.

\header{Hybrid and Learning-Based Approaches}
Recent developments have sought to combine the strengths of sequential and variational schemes with machine learning. One active direction is to learn or approximate components of the assimilation pipeline. For example, neural networks can serve as surrogate observation operators, mapping model state variables to observation space when the true operator is highly nonlinear or computationally expensive \cite{fang2021using}. Another line of work integrates machine learning to improve error covariance modeling, a long-standing challenge in EnKF and variational schemes \cite{bocquet2020bayesian}. The concept of \textit{Neural Data Assimilation} has also emerged, where deep learning architectures are trained to emulate the update step directly, providing a data-driven approximation of Bayesian inference. Furthermore, differentiable programming frameworks have enabled the formulation of \textit{differentiable DA}, where the assimilation process itself is embedded into a computation graph, facilitating end-to-end learning of model and assimilation parameters. PINNs, although already introduced in Section~3.1.2, can also be incorporated into assimilation frameworks as constraints that guide state estimation \cite{raissi2019physics}. These hybrid methods highlight a promising synergy: traditional DA offers a principled statistical framework, while AI methods bring flexibility, scalability, and the ability to exploit large datasets.

Data assimilation represents the methodological bridge between data and models in digital twins. Sequential methods emphasize real-time, step-by-step updating, variational methods leverage optimization over temporal windows, and hybrid methods incorporate machine learning to overcome long-standing limitations. Together, they ensure that the physical twin is not merely simulated but continuously synchronized with reality. In doing so, data assimilation transforms the digital twin into a dynamic mirror of its physical counterpart, capable of supporting reliable prediction, optimization, and decision-making in complex environments.

\section{Mirroring the Physical Twin into the Digital Twin}\label{sec:4}
% simulator building and visualization
Learning a distribution based on real data observations is a fundamental challenge in generative AI. Digital twins, which vary in their representation dimensions, require different frameworks for modeling. This section will go through some fundamental approaches and then cover recent state-of-the-art methods built upon for learning digital twins in 2D and 3D dimensions with dynamic modeling.
\begin{figure}[t]
    \centering
\includegraphics[width=0.95\linewidth]{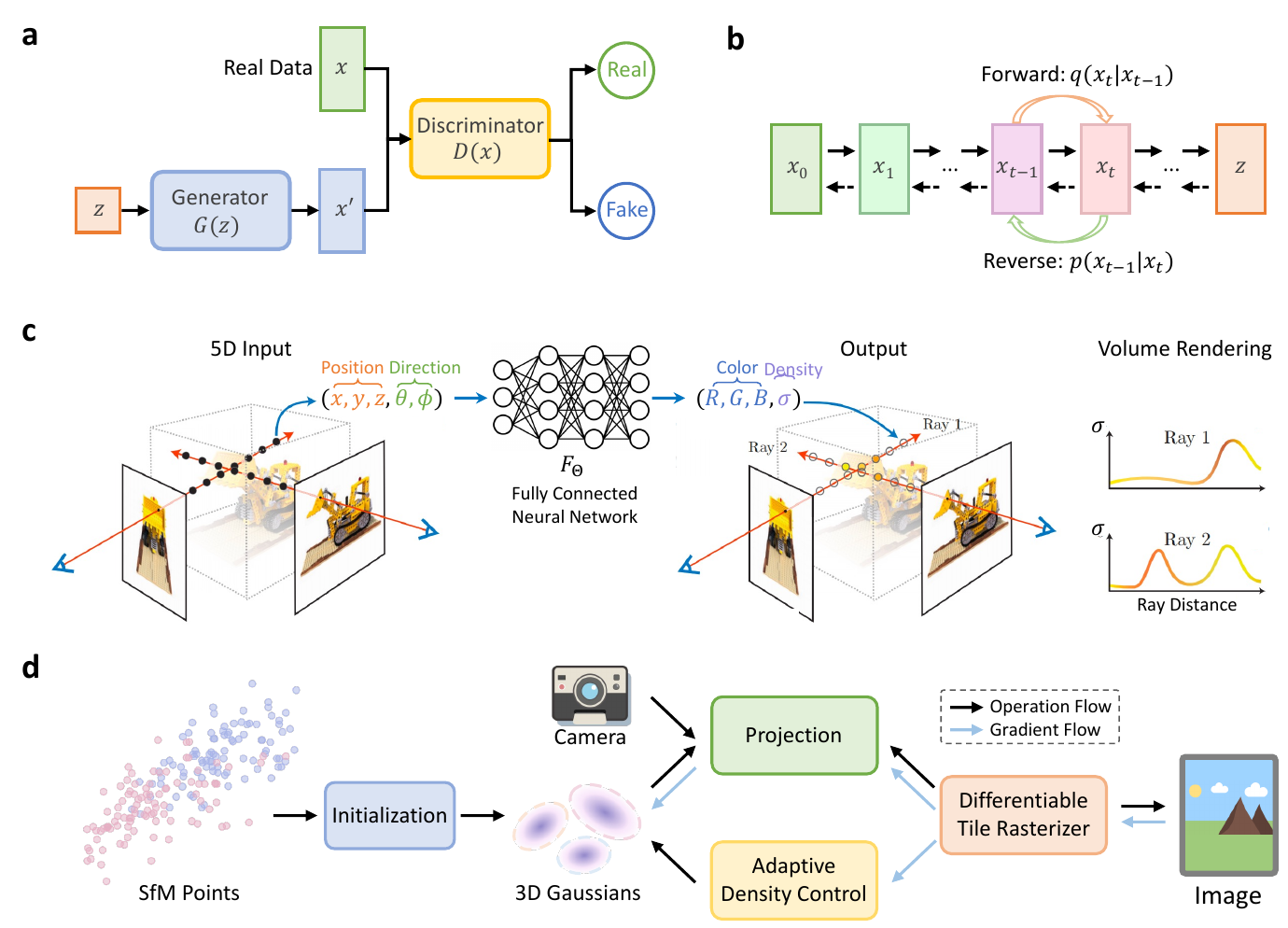}
    \caption{Generative AI models. \textbf{(a)} The framework of Generative Adversarial Networks (GANs). \textbf{(b)} The framework of Denoising diffusion probabilistic models (DDPMs). \textbf{(c)} The framework of Neural Radiance Field (NeRF). \textbf{(d)} The framework of 3D Gaussian Splatting.}
    \label{fig:PM}
\end{figure}

\header{Variational Autoencoders (VAEs)}
% \label{sec:vae}
VAEs~\cite{rezende2014stochastic,doersch2016tutorial,kingma2019introduction} are designed to learn an encoder-decoder pair that maps input data into a continuous latent space with two Gaussian proxy distributions. In such a framework, we train the encoder model to match the prior distribution of the latent variable $\mathbf{z}$ and train the decoder model to reconstruct the original image. In the encoder part, we employ a Gaussian proxy distribution $q_\phi(\mathbf{z}\mid\mathbf{x})$ parameterized by a neural network $\phi$ to approximate the intractable posterior distribution $q(\mathbf{z}\mid\mathbf{x})$. We seek to predict the mean and variance of the latent variable $\mathbf{z}$ from the input image $\mathbf{x}$, i.e., $\mu_\phi(\mathbf{x})$ and $\sigma_\phi(\mathbf{x})$, to align the encoded distribution closer to pre-defined prior distribution ($p(\mathbf{z})=\mathcal{N}(\mathbf{z}; \mathbf{0}, \mathbf{I})$). In the decoder part, assuming the images follow a Gaussian distribution, we essentially train the decoder $p_\theta(\mathbf{x}\mid\mathbf{z})$ to reconstruct the original image from the latent variable $\mathbf{z}$. To achieve these two objectives, we can maximize the evidence lower bound (ELBO)~\cite{kingma2019introduction}, which is defined as
\begin{align}
\mathcal{L}(\boldsymbol{\phi}, \boldsymbol{\theta}; \mathbf{x})
&= \mathbb{E}_{q_{\boldsymbol{\phi}}(\mathbf{z} \mid \mathbf{x})}\left[ \log p_{\boldsymbol{\theta}}(\mathbf{x},\mathbf{z}) - \log q_{\boldsymbol{\phi}}(\mathbf{z}\mid\mathbf{x}) \right], \\
&= \mathbb{E}_{q_{\boldsymbol{\phi}}(\mathbf{z} \mid \mathbf{x})}\left[\log p_{\boldsymbol{\theta}}(\mathbf{x} \mid \mathbf{z})\right] - \mathbb{D}_{\mathrm{KL}}\left(q_{\boldsymbol{\phi}}(\mathbf{z} \mid \mathbf{x}) \| p(\mathbf{z})\right), 
\label{eq:elbo}
\end{align}
with the constraint $\mathcal{L}(\phi, \theta; \mathbf{x}) \leq \log p_\theta(\mathbf{x})$. Since the loss function in Eq. \ref{eq:elbo} is differentiable, we can train the encoder $\phi$ and the decoder $\theta$ using gradient descent in an end-to-end manner. During inference time, we can sample latent variables from $p(\mathbf{z})$ and then feed them into the decoder to generate new images.
% /Volumes/SanDisk/top-papers/basic-paper/diffusion-models/chan.pdf
% /Volumes/SanDisk/top-papers/basic-paper/diffusion-models/lingyang.pdf

\header{Normalizing Flows} Normalizing flows~\cite{0f899b92b7fb03b609fee887e4b6f3b633eaf30d,729b18d8d91035f4bb84bf2e61b0517824e5d31b,f31dc980804ad32eb82993524af8ca790ab17c82,d81edc320369c9c26e26a6801e08d68c4daf766c,c8b25a128f4bfd0c79de82c174dd403b2ef6eeb1} are a powerful class of generative models that enable flexible and tractable density estimation by transforming a simple probability distribution into a more complex one through a series of invertible transformations. The core idea of normalizing flows is to start with a simple distribution, typically a multivariate Gaussian~\cite{kingma2019introduction}, and apply a sequence of bijective (invertible and differentiable) mappings to transform this simple distribution into one that matches the target data distribution. Specifically, let $\mathbf{z}_0 \sim p_{\mathbf{z}_0}(\mathbf{z}_0)$ represent a random variable drawn from a simple base distribution $p_{\mathbf{z}_0}$, such as a standard normal distribution. A normalizing flow applies a sequence of invertible transformations $f_i: \mathbb{R}^d \rightarrow \mathbb{R}^d$ for $i = 1, 2, \dots, K$, resulting in the transformed variable $\mathbf{z}_K = f_K \circ f_{K-1} \circ \dots \circ f_1 (\mathbf{z}_0)$. The probability density function of $\mathbf{z}_K$ under the transformation is given by the change of variables formula:
\begin{equation}
    \label{eq:normalizing_flow}
    p_{\mathbf{z}_K}(\mathbf{z}_K) = p_{\mathbf{z}_0}(\mathbf{z}_0) \left| \det \left( \frac{\partial f^{-1}}{\partial \mathbf{z}_K} \right) \right| = p_{\mathbf{z}_0}(\mathbf{z}_0) \prod_{i=1}^{K} \left| \det \left( \frac{\partial f_i}{\partial \mathbf{z}_{i-1}} \right) \right|^{-1},
\end{equation}
where $\det \left( \frac{\partial f_i}{\partial \mathbf{z}_{i-1}} \right)$ is the determinant of the Jacobian matrix of the $i$-th transformation. This formulation allows for exact likelihood computation, making normalizing flows highly effective for density estimation.

While normalizing flow offers several advantages, such as exact likelihood estimation and invertibility, they also have some limitations. One of the primary challenges is the trade-off between the expressiveness of the transformations and the computational complexity of calculating the Jacobian determinants~\cite{3c8a110d71ef3132431288aee05b189e206d6ea9,489fab3d30e6e35c51de010f78a78cecd3d758a0,533e651177354d6687e8d0a24217a6bf65692bcc}. Simple transformations lead to efficient computation but may lack the flexibility to model complex distributions, whereas more complex transformations can capture intricate structures in the data but at the cost of increased computational burden. To tackle these challenges, recent researches attempt to enhance the expressiveness of normalizing flows, such as Neural Spline Flows~\cite{1eeb265595e250cf66751ef9032524386d7a9b32} and Residual Flows~\cite{ef2b790aae2c1e1c7eb8a8777515266b5094de88}, which balance expressiveness and computational efficiency. 

\header{Generative Adversarial Neural Networks (GANs)}
GANs~\cite{goodfellow2014generative,creswell2018generative,gui2021review} is a kind of powerful framework (as shown in Figure \ref{fig:PM}) for learning data distributions, consisting of two main components: a generator ($G$) and a discriminator ($D$). These components are typically implemented as differentiable neural networks that map input data from one space to another. The optimization of GANs can be formulated as a min-max game between the generator and discriminator, with the following objective:
\begin{equation}
\min_G \max_D \mathbb{E}_{\mathbf{x} \sim p{\text{data}}(\mathbf {x})}[\log D(\mathbf{x})] + \mathbb{E}_{\mathbf{z} \sim p{\mathbf{z}}(\mathbf{z})}[\log (1 - D(G(\mathbf{z})))].
\end{equation}
The generator's goal is to create new examples that closely resemble the real data distribution, while the discriminator aims to accurately distinguish between real and generated examples. The training process reaches an equilibrium point, known as the Nash equilibrium~\cite{ratliff2013characterization}, where the generator has effectively captured the true data distribution.
However, GANs often face training instability issues due to the non-overlapping support between the real and generated data distributions. One approach to mitigate this problem is to introduce noise into the discriminator's input, thereby widening the support of both distributions. Wang et al. (2022)~\cite{wang2022diffusion} propose an adaptive noise injection scheme based on a diffusion model to stabilize GAN training. Due to the one-shot generation nature, GAN can be an efficient alternative compared to the more capable diffusion models, which require iterative multi-step and time-consuming denoising during inference. 

\header{Denoising Diffusion Probabilistic Models (DDPMs)} DDPMs use two Markov chains: a forward chain adding noise and a reverse chain removing it. The forward chain transforms data into a simple prior distribution, while the reverse chain, using neural networks, inverts this process. Given a data distribution $\mathbf{x}_0 \sim q(\mathbf{x}_0)$, the forward process generates a sequence $\mathbf{x}_1, \mathbf{x}_2, \dots, \mathbf{x}_T$ with transition kernel $q(\mathbf{x}_t\mid\mathbf{x}_{t-1})$. The joint distribution is the product of these transitions. The transition kernel is typically a Gaussian perturbation: 
% \begin{equation}
% \end{equation}
$q(\mathbf{x}_t\mid\mathbf{x}_{t-1}) = \mathcal{N}(\mathbf{x}_t; \sqrt{1-\beta_t} \mathbf{x}_{t-1}, \beta_t \mathbf{I}),$
where $\beta_t \in (0,1)$ is a hyperparameter. This allows for analytical marginalization:
% \begin{equation}
% \end{equation}
$q(\mathbf{x}_t\mid\mathbf{x}_0) = \mathcal{N}(\mathbf{x}_t; \sqrt{\bar{\alpha}_t} \mathbf{x}_0, (1-\bar{\alpha}_t) \mathbf{I}), $
with $\alpha_t = 1 - \beta_t$ and $\bar{\alpha}_t = \prod_{s=0}^{t} \alpha_s$. The reverse process uses a learnable transition kernel:
\begin{equation}
  p_\theta(\mathbf{x}_{t-1}\mid\mathbf{x}_t) = \mathcal{N}(\mathbf{x}_{t-1}; \mu_{\theta}(\mathbf{x}_t, t), \Sigma_{\theta}(\mathbf{x}_t, t)),
\end{equation}
where $\theta$ denotes model parameters. It's trained to match the time reversal of the forward process by minimizing KL divergence. The simplified loss function is:
\begin{equation}
\ell^{\text {simple }}_t(\theta)=\mathbb{E}_{\mathbf{x}_0, t, \epsilon_t}\left\|\epsilon_\theta\left(\mathbf{x}_t, t\right)-\epsilon\right\|_2^2. 
\end{equation}

The denoising process proceeds step-by-step using a formula involving the learned noise prediction $\epsilon_\theta$ and Gaussian noise $\mathbf{z}$.

Sampling efficiency was enhanced through techniques like DDIM~\cite{014576b866078524286802b1d0e18628520aa886}, which enables faster sampling by constructing non-Markovian diffusion processes. The EDM framework~\cite{2f4c451922e227cbbd4f090b74298445bbd900d0} further improved efficiency and quality by refining the design space. Score-based models~\cite{633e2fbfc0b21e959a244100937c5853afca4853} provided a unified view of diffusion models through stochastic differential equations. For discrete data, D3PMs~\cite{91b32fc0a23f0af53229fceaae9cce43a0406d2e} extended diffusion models to discrete state spaces. Practical applications were advanced by works like GLIDE~\cite{7002ae048e4b8c9133a55428441e8066070995cb} and Stable Diffusion~\cite{rombach2022high}, which enabled high-quality text-to-image generation. Classifier guidance~\cite{64ea8f180d0682e6c18d1eb688afdb2027c02794,de18baa4964804cf471d85a5a090498242d2e79f} and cross-attention control~\cite{04e541391e8dce14d099d00fb2c21dbbd8afe87f} further improved conditional generation and editing.

\header{Radiance Field Modeling with Implicit and Explicit Methods} A radiance field provides a three-dimensional light distribution model that describes the interaction of light with surfaces and materials within an environment~\cite{Ansh_Mittal_2023}. It can be mathematically expressed as a function \( L \!:\! \mathbb{R}^5 \!\mapsto\! \mathbb{R}^+ \), where \( L(x, y, z, \theta, \phi) \) denotes the mapping of a spatial point \((x, y, z)\) and a direction given by spherical coordinates \((\theta, \phi)\) to a non-negative radiance value. Radiance fields are typically represented in two forms: implicit or explicit, each offering distinct advantages for scene depiction and rendering~\cite{Stefan_Zeisl_2023}. An implicit radiance field models the light distribution within a scene indirectly without defining the scene's geometry explicitly~\cite{Hu_Zhan_2022}. In the context of deep learning, this often involves employing neural networks to learn a continuous representation of the volumetric scene~\cite{Shude_Chen_2023}. A notable example is NeRF~\cite{mildenhall2020nerf}, where a neural network, often a multi-layer perceptron (MLP), maps spatial coordinates \( (x,y,z) \) and viewing directions \( (\theta, \phi) \) to corresponding color and density values~\cite{Jianlin_Liu_2023}. The radiance at any point is dynamically computed by querying the MLP rather than being stored directly. This approach provides a compact and differentiable representation of complex scenes but typically requires significant computational resources during rendering due to the need for volumetric ray marching~\cite{Nikola_Popović_2022}.
\textit{Conversely}, an explicit radiance field explicitly encapsulates the light distribution using discrete spatial structures like voxel grids or point sets~\cite{Lingzhi_Li_2022}. Each component of this structure encodes the radiance data for its specific spatial location, facilitating faster and more direct radiance retrieval, albeit at the expense of increased memory demands and reduced resolution~\cite{Yuedong_Chen_2023}. 

\header{3D Gaussian Splatting (3DGS)} 3DGS ~\cite{2cc1d857e86d5152ba7fe6a8355c2a0150cc280a} integrates the benefits of both implicit and explicit radiance fields through the use of adjustable 3D Gaussians. This method, optimized by multi-view image supervision, provides an efficient and flexible representation capable of accurately depicting scenes. It merges neural network-based optimization with structured data storage, targeting real-time, high-quality rendering with efficient training, especially for intricate and high-resolution scenes. The 3DGS model is described as:
\begin{equation}
    L_{\text{3DGS}}(x, y, z, \theta, \phi) = \sum_{i} G(x, y, z, \bm{\mu}_i, \bm{\Sigma}_i) \cdot c_i(\theta, \phi),
\end{equation}
where \( G \) symbolizes the Gaussian function, defined by the mean \( \bm{\mu}_i \) and covariance \( \bm{\Sigma}_i \), and \( c \) indicates the view-dependent color.
% \fixme{then how to learn 3D Gaussian, the basic process description, and also those important loss functions}
Specifically, learning 3D GS involves two main processes as depicted in Figure \ref{fig:PM}: 
\begin{enumerate}
    \item \textit{Rendering}: The rendering process in 3D GS differs significantly from the volumetric ray marching used in implicit methods like NeRF. Instead, it employs a splatting technique that projects 3D Gaussians onto a 2D image plane. This process involves several key steps. First, in the frustum culling step, Gaussians outside the camera's view are excluded from rendering. Then, in the splatting step, 3D Gaussians are projected into 2D space using a transformation involving the viewing transformation and the Jacobian of the projective transformation. In the final step, the color of each pixel is computed using alpha compositing, which blends the colors of overlapping Gaussians based on their opacities.
    \item \textit{Optimization}: To achieve real-time rendering, 3D GS employs several optimization techniques, including the use of tiles (patches) for parallel processing and efficient sorting of Gaussians based on depth and tile ID. The learning process in 3D GS involves optimizing the properties of each Gaussian (position, opacity, covariance, and color) as well as controlling the density of Gaussians in the scene. The optimization is guided by a loss function combining L1 and D-SSIM losses:
    \begin{equation}
    \mathcal{L} = (1-\lambda)\mathcal{L}_1 + \lambda\mathcal{L}_{\text{D-SSIM}},
\end{equation}
where $\lambda$ is a weighting factor. The density of Gaussians is controlled through point densification and pruning processes. Densification involves cloning or splitting Gaussians based on positional gradients, while pruning removes unnecessary or ineffective Gaussians.
\end{enumerate}
 
Since the introduction of 3D Gaussian Splatting for real-time radiance field rendering~\cite{2cc1d857e86d5152ba7fe6a8355c2a0150cc280a}, several significant advancements have been made in the field. Plenoxels~\cite{e91f73aaef155391b5b07e6612f5346dea888f64} introduced a neural network-free approach for photorealistic view synthesis using a sparse 3D grid with spherical harmonics. Dynamic 3D Gaussians~\cite{8ba2c82fe675ede1d5c12fc4f97cf8ca3ebf1ca3} extended the concept to dynamic scenes, enabling six-degree-of-freedom tracking and novel-view synthesis. Animatable and Relightable Gaussians~\cite{ef951d12930bc0bfce46c0e10a0f291c87455551} focused on high-fidelity human avatar modeling from RGB videos. Other notable works include Floaters No More~\cite{18c73ceb30817c836978cd648687a765829a4b69}, which addressed background collapse in NeRF acquisition, and 3DGS-Avatar~\cite{fa35524739d5c60d94befc3f8e77488b4dd810db}, which achieved real-time rendering of animatable human avatars. These advancements highlight the ongoing evolution and diversification of techniques in the realm of 3D scene representation and rendering.

\subsection{Simulator Building}
\subsubsection{Foundational Modeling and State Representation}
A digital twin simulator must first decide how to represent the physical system it seeks to emulate. In some cases, the simulator requires an explicit geometric description to reproduce spatially governed behaviors such as deformation, fluid flow, or molecular interactions. In other cases, geometry is unnecessary, and the system can be modeled through relational or temporal dependencies that capture how components interact or evolve over time. These two complementary perspectives lead to two types of state representation: geometry-based, which encodes the physical structure of the system, and abstract, which focuses on the data-driven or relational aspects of system behavior.

\header{Geometry-Based State Representation}
In geometry-based representations, the physical system is modeled through explicit spatial constructs that define its shape, structure, and material properties. These representations form the backbone of digital twins in domains where spatial configuration directly governs system dynamics and behavior, such as engineering design, manufacturing, and biomedical modeling. They answer fundamental questions about the physical world: \textit{What does the object look like? How are its components connected? What are its physical properties?}

\textit{Shape and Geometry.} “What is the shape of an object?” - This question defines the geometric configuration of the physical entity in space. Shape representation establishes the foundation for all subsequent modeling, determining how the object occupies and interacts with its environment. Common formulations include meshes used in finite element and finite volume methods (FEM/FVM), which discretize continuous domains for numerical analysis~\cite{huebner2001finite,rao2017finite}; point clouds capturing dense surface samples for 3D reconstruction and inspection~\cite{rusu20113d}; voxel or grid representations used in volumetric modeling and medical imaging~\cite{cciccek20163d}; and parametric models such as CAD and Building Information Models (BIM) that encode both geometry and semantics~\cite{eastman2011bim}. Neural fields, including Neural Radiance Fields (NeRF)~\cite{mildenhall2021nerf}, Signed Distance Fields (SDF)~\cite{curless1996volumetric}, and occupancy networks~\cite{mescheder2019occupancy}, further generalize geometry into continuous implicit functions, offering differentiable and compact scene representations. Together, these formulations provide the means to reconstruct, visualize, and simulate the spatial state of the physical world.

\textit{Topology and Connectivity.} “How are its components connected?” Beyond local geometry, topology describes the relationships and connectivity that define structural integrity and motion constraints. In mechanical systems, mesh connectivity specifies adjacency relations between elements and supports stress or deformation analysis~\cite{rao2017finite}. In articulated systems such as robotic manipulators, kinematic chains describe hierarchical dependencies and degrees of freedom among joints and links~\cite{siciliano2016springer}. Assembly graphs further encode how individual parts interface or move relative to one another, enabling simulation of multi-body dynamics and structural coupling~\cite{du2021assembly}. Accurate topological modeling ensures that the digital twin preserves the structural logic of the physical system, allowing analyses such as load propagation, collision detection, and deformation tracking under physical constraints.

\textit{Physical Attributes.} “What are its physical properties?” Geometry and topology alone cannot determine the system’s behavior without describing its intrinsic material and boundary properties. These attributes define how the physical entity responds to external forces, heat, or other environmental stimuli. Key descriptors include material properties such as density, elasticity, and viscosity; boundary conditions specifying loads, fixed supports, or fluid interfaces; thermal parameters including conductivity and specific heat; and initial conditions that define the starting state of fields such as temperature or velocity. Together, these quantities enable accurate numerical simulation and predictive modeling through governing physical equations. The inclusion of such physical attributes transforms geometric models from static visualizations into dynamic computational entities that mirror the real-world system’s mechanical, thermal, or electromagnetic responses.

\header{Non-Geometric State Representation}
Not all simulators require explicit geometry to capture system behavior. In many digital twins, the key dynamics arise from relationships, temporal patterns, or statistical dependencies rather than spatial form. Non-Geometric State Representations therefore describe systems through symbolic, relational, or feature-based structures, enabling modeling of processes where geometry is unavailable, irrelevant, or computationally unnecessary. Such representations are widely used in cyber-physical, social, and biological systems to support large-scale reasoning and prediction.

\textit{Feature-based Representation.}  
Feature-based representations encode complex system states into compact numerical vectors, allowing efficient computation and integration across modalities. These embeddings can be extracted from text, images, and sensor measurements, or learned through self-supervised and multimodal models~\cite{bengio2013representation}. Large-scale foundation models~\cite{radford2021learning} have further generalized this concept by learning unified latent spaces capable of representing diverse system behaviors and attributes. Within digital twin simulators, embeddings provide scalable interfaces for high-dimensional inference, supporting fast querying, cross-domain adaptation, and intelligent decision-making.

\textit{Time-Series Representation.}  
Time-Series representations describe how system states evolve over time. They are central to digital twins that monitor and forecast behaviors from continuous data streams, such as sensor signals, physiological measurements, or environmental variables~\cite{lipton2015learning}. Recurrent and attention-based models, including RNNs, TCNs, and Transformers~\cite{vaswani2017attention}, have shown strong ability to capture both short- and long-term dependencies. In healthcare twins, these representations enable dynamic patient monitoring and disease progression modeling~\cite{pierson2021digital}; in industrial twins, they support predictive maintenance and anomaly detection based on telemetry data~\cite{wen2022industry}.

\textit{Graph-based Representation.}  
Graph-based representations model a system as a collection of entities and their interactions, providing a natural framework for describing relational dynamics. This formulation has proven effective for networked infrastructures such as transportation systems~\cite{yu2017spatio}, supply chains~\cite{ivanov2020viability}, and energy grids~\cite{palensky2022digital}, as well as for semantic and biomedical knowledge graphs~\cite{wang2022review}. Graph Neural Networks (GNNs) extend these ideas by learning representations over structured dependencies, enabling simulation of flow, fault propagation, and system-level optimization.

Non-Geometric State Representations thus complement geometry-based models by focusing on relationships and dynamics rather than spatial fidelity. Together, they define two foundational paradigms for simulator construction: one grounded in physical form, the other in data and interaction. Modern digital twins often integrate both perspectives. For example, structural components can be modeled geometrically, while control, communication, or biological processes are represented through graphs or embeddings. This hybrid approach enables simulators to capture both the physical dynamics and semantic interactions of complex systems.

\subsubsection{Behavior and Process Simulation}

\header{State-Space Simulation}
State-space simulation represents one major direction for building simulators. The goal is to learn system behavior directly from data, treating simulation as \textit{function approximation} that maps inputs to outputs or as \textit{sequence prediction} that forecasts temporal evolution. Rather than solving governing equations explicitly, these models infer the underlying dynamics from observational evidence, allowing digital twins to emulate complex processes efficiently when analytical formulations are unavailable or computationally prohibitive~\cite{willard2020integrating}.

Feedforward models such as MLPs, CNNs, and ResNets learn steady-state mappings between design or control variables and resulting performance indicators. They are often used as differentiable surrogates for engineering optimization—for instance, predicting aerodynamic lift from airfoil geometry, estimating heat dissipation from material parameters, or forecasting drug release rate from formulation properties~\cite{yang2024data}. By replacing expensive finite-element simulations, such neural surrogates enable rapid design iterations and sensitivity analysis.
When temporal dependencies dominate, recurrent and attention-based networks including LSTMs, GRUs, and Transformers~\cite{vaswani2017attention} capture the dynamic evolution of system states. These models treat simulation as sequence forecasting, learning how a process unfolds over time. In industrial digital twins, they are applied to predict sensor trajectories for fault detection~\cite{wen2022industry}; in healthcare, they model disease progression using patient time-series data~\cite{pierson2021digital}; in transportation, they forecast multi-step traffic flow and congestion propagation across urban networks~\cite{yu2017spatio}.

For systems with explicit spatial or relational structure, GNNs~\cite{battaglia2018relational} extend this paradigm by learning interactions among interconnected components. Each node represents a physical or logical entity, and edges describe dependencies such as force transmission or resource exchange. This allows GNN-based simulators to reproduce mesh deformation in mechanical systems, voltage propagation in power grids, and flow redistribution in transportation networks. Through message passing, GNNs enable digital twins to model how local perturbations collectively shape global dynamics.
Beyond discrete representations, neural operators such as DeepONet~\cite{lu2019deeponet} and the Fourier Neural Operator (FNO)~\cite{li2020fourier} generalize state-space learning to functional mappings. Instead of approximating a single trajectory, they learn solution operators of partial differential equations, directly mapping boundary conditions or source terms to entire solution fields. Once trained, these operators provide fast and high-fidelity surrogates for simulating fluid flow, heat transfer, and material deformation~\cite{gupta2021multiwavelet,lu2021learning,hennigh2021nvidia}, achieving speedups of several orders of magnitude over traditional solvers.

In summary, state-space simulation focuses on learning how systems respond and evolve, either by approximating steady mappings or by predicting temporal sequences. From predicting equipment failure to optimizing engineering designs and accelerating physical simulations, this approach forms a central computational pathway for building intelligent digital twins that learn, adapt, and generalize across physical domains.

\header{Visual World Simulation}
Recent advances in generative AI have introduced a new paradigm for simulation by directly generating realistic visual observations of the world rather than explicitly modeling its internal states. In this view, predicting future frames or synthesizing dynamic scenes becomes equivalent to simulating the evolution of the physical environment. Such \textit{world simulators} enable embodied agents, including robots and autonomous vehicles, to learn, plan, and interact within controllable, data-driven virtual worlds without relying solely on costly physical experiments.

Video diffusion models have emerged as one of the most powerful frameworks for generative simulation. Leveraging large-scale video datasets and diffusion-based architectures, these models can synthesize high-fidelity and temporally coherent videos that approximate real-world physics and dynamics. For instance, VideoComposer~\cite{f02ea7a18f00859d9ea1b321e3385ae7d0170639} introduces a compositional video synthesis framework conditioned on text, spatial layout, and temporal cues, allowing precise control over generated motion and scene composition. DynamiCrafter~\cite{083bab4a967c2221d9f4da9110fe37d8ca679078} extends this capability by animating still images through motion priors learned from text-to-video diffusion models, effectively turning static scenes into dynamic simulations. Most notably, Sora~\cite{videoworldsimulators2024} demonstrates the remarkable potential of text-to-video diffusion models as universal world simulators, capable of generating physically consistent, photorealistic videos from natural language descriptions.

While realism is critical, \textit{controllability} remains essential for simulation-driven learning. Several recent studies have focused on integrating structured control into video generation models to produce task-relevant, interactive environments. Seer~\cite{46a97c83626132db81602becab3379c1cc4edf44} introduces a frame-sequential text decomposer that translates global instructions into temporally aligned sub-instructions, enhancing fine-grained control over the resulting video trajectories. Video Adapter~\cite{7820f9e98c9d064a0402685be2cf875a916edd27} provides a lightweight adaptation mechanism for large pre-trained video diffusion models, enabling efficient domain customization without full finetuning. At a larger scale, the Cosmos World Foundation Model Platform~\cite{agarwal2025cosmos} offers an integrated framework for constructing controllable video-based world models, including video curation pipelines, tokenizers, and pre-trained foundation simulators for physical AI applications such as robotics and autonomous driving.

Beyond pure video generation, interactive world models like Genie~\cite{bruce2024genie} and its successor Genie 2~\cite{parker-holder2024genie2} extend generative simulation into embodied, action-controllable 3D environments. These systems can generate playable, open-ended virtual worlds conditioned on text, sketches, or other multimodal prompts, enabling autonomous agents to learn and act within dynamic visual environments.

Video generation as world simulation thus represents a paradigm shift from explicit physics-based modeling to observation-driven synthesis. By learning the visual and temporal structure of reality, such models offer controllable, scalable, and photorealistic environments for training, testing, and reasoning. Within digital twin systems, they bridge simulation and perception, allowing virtual agents to both observe and interact with realistic representations of the physical world.

\subsection{Simulator Visualization}

\subsubsection{Scene Modeling}
In digital twins, scene modeling supports visualization by defining how the simulated world appears in three-dimensional space. Once the simulator engine has been built to model system behavior, visualization focuses on reconstructing or synthesizing the world’s visible structure. Scene modeling determines what the twin “looks like.” Scene modeling aims to represent the spatial configuration and visual appearance of real or virtual environments by learning how light interacts with matter through radiance field representations. With advances in neural rendering, data-driven methods can now recover highly detailed and spatially consistent 3D and even 4D scenes directly from multi-view images or sensor data. This section introduces two main directions in this field: large-scale static scene reconstruction and dynamic scene modeling.

\header{Static Scene Reconstruction} 
Scaling neural rendering techniques to large urban environments has been a significant focus of recent research. These methods aim to capture the complexity of city-scale scenes while maintaining high visual fidelity and efficient rendering. Block-NeRF~\cite{d7d1bbade9453f0348fac8a5c60d131528b87fcf} introduces a variant of NeRF that can represent large-scale environments by decomposing the scene into individually trained NeRFs. This approach decouples rendering time from scene size, enabling rendering to scale to arbitrarily large environments. Urban Radiance Fields~\cite{102d29870ba101004afce311823df85a9f304be7} extends NeRF to handle asynchronously captured lidar data and address exposure variation between captured images, producing state-of-the-art 3D surface reconstructions and high-quality novel views for street scenes. TensoRF~\cite{b4a9437302411abde0c9de784a14bfc6a5d950cf} proposes modeling the radiance field as a 4D tensor and introduces vector-matrix decomposition to achieve fast reconstruction with better rendering quality and smaller model size compared to NeRF. NeRF in the Wild~\cite{691eddbfaebbc71f6a12d3c99d5c155042459434} addresses the challenges of unconstrained photo collections, enabling accurate reconstructions from internet photos of famous landmarks. More recent works have focused on further improving the scalability and quality of large-scale scene reconstruction. K-Planes~\cite{2b4de48703d5278afbce69844d5ed92b5a699ee1} introduces a white-box model for radiance fields using planes to represent d-dimensional scenes, providing a seamless way to go from static to dynamic scenes. BungeeNeRF~\cite{795fecd949592a89c88cc96d22478df04519d4f8} achieves level-of-detail rendering across drastically varied scales, addressing the challenges of extreme multi-scale scene rendering.
Global-guided Focal Neural Radiance Field~\cite{cc7bdcf19d64246eea7d5172905f9697734d3714} proposes a two-stage architecture to achieve high-fidelity rendering of large-scale scenes while maintaining scene-wide consistency. CityGaussian~\cite{d0aaeb36661475a6cc840c1353adf25f5890d27a} employs a novel divide-and-conquer training approach and Level-of-Detail strategy for efficient large-scale 3D Gaussian Splatting training and rendering.

\header{Dynamic Scene Modeling}
Extending static representations to model dynamic scenes with moving objects has been another important direction in neural rendering research. These methods aim to capture both the spatial and temporal aspects of complex real-world environments.
4D Gaussian Splatting~\cite{716754c9f228a4f63d8e6edb81e200d6f4ea0c3a} approximates the underlying spatio-temporal 4D volume of a dynamic scene by optimizing a collection of 4D primitives, enabling real-time rendering of complex dynamic scenes. Scalable Urban Dynamic Scenes (SUDS)~\cite{2643f90ed0b3e3ee08e7c30a8775465a5d217a47} introduces a factorized scene representation using separate hash table data structures to efficiently encode static, dynamic, and far-field radiance fields.
Street Gaussians~\cite{fca783b3f5c8e68135c8b8bd0b63a5c6e4d628f6} proposes a novel pipeline for modeling dynamic urban street scenes, using a combination of static and dynamic 3D Gaussians with optimizable tracked poses for moving objects. 
Deformable 3D Gaussians~\cite{1ce81d64eefe5915c6ef9719915efa5f4079a6c1} introduces a method that reconstructs scenes using 3D Gaussians and learns them in canonical space with a deformation field to model monocular dynamic scenes.
DynMF~\cite{a082a0df25b7166706e81ade83bc917dd7367edc} presents a compact and efficient representation that decomposes a dynamic scene into a few neural trajectories, allowing for real-time view synthesis of complex dynamic scene motions. Multi-Level Neural Scene Graphs~\cite{d78d69caac2aee35a27e4fdf6f5bbe44f6ced95f} proposes a novel, decomposable radiance field approach for dynamic urban environments, using a multi-level neural scene graph representation that scales to thousands of images with hundreds of fast-moving objects.

\subsubsection{Interactive Visualization and Interfaces}

While scene modeling establishes the structural and visual substrate of the virtual environment, interactive visualization determines how humans and AI agents \emph{see}, \emph{explore}, and \emph{make sense of} that environment. This layer transforms simulated or reconstructed worlds into perceivable, analyzable views, supporting situational awareness, hypothesis testing, and collaborative understanding through intuitive visual interfaces.

\header{Immersive and Real-Time Visualization}
Advances in rendering pipelines and GPU acceleration now allow digital twins to achieve photorealistic, real-time visualization of large-scale environments. Neural rendering techniques such as 3D Gaussian Splatting~\cite{kerbl20233d} make it possible to maintain interactive frame rates without sacrificing fidelity, while compact scene representations like Instant-NGP~\cite{mueller2022instant} further improve efficiency for view synthesis. Immersive visualization systems, including AR/VR headsets and CAVE displays—bring depth perception and spatial presence to users~\cite{li2022surveyvr}, enabling intuitive exploration through gestures, motion tracking, or gaze-based control. Such immersive systems enhance human perception and understanding in domains such as surgical simulation, smart city management, and industrial training~\cite{huang2024mixed}.

\header{Interactive Dashboards and Visual Analytics}
Beyond immersion, digital twins rely on interactive dashboards and visual analytics tools to organize and interpret simulation results and live sensor data~\cite{grieves2020digital}. These interfaces integrate predictive models, streaming information, and diagnostic views into a unified visual layer that supports real-time monitoring and reasoning~\cite{tao2022digital}. In practice, they appear as 3D monitoring dashboards for smart factories~\cite{qi2021digital}, preoperative visualization systems for surgical planning~\cite{fang2022surgical}, or operational control centers that display traffic flow and energy distribution for situational analysis~\cite{chen2023digital}. Recent cloud-based visualization frameworks further allow distributed teams to collaboratively explore, annotate, and interpret digital twin environments through synchronized visual interfaces.

\section{Intervening in the Physical Twin via the Digital Twin}\label{sec:5}

\subsection{Predicting Physical Behavior}
Prediction modeling is a fundamental aspect of digital twin systems, allowing for the forecasting of future conditions and behaviors based on current and historical data. In digital twin systems, which create virtual replicas of physical entities, predictive models can forecast various scenarios such as equipment malfunctions~\cite{Luo2020ah}, performance degradation~\cite{Li2022ano}, and system anomalies~\cite{Xu2021dig}. These predictions are essential for optimizing performance, preventing unexpected issues, and ensuring the efficient operation of physical assets~\cite{Zhu2020rea, Francisco2020sma, Pan2020dig}. Leveraging advanced AI techniques such as machine learning and deep learning methods, digital twins can analyze extensive data to predict trends~\cite{Guc2022sma, Liu2018the}, detect anomalies~\cite{Lian2023ano, Hao2021hyb}, and make decisions with real-time~\cite{Li2021tra, Liang2023spa, Kong2022dyn}. This section will delve into two parts, focusing on: 1) Prediction Modeling in Digital Twin Systems; 2) Types of Prediction in Digital Twin Systems.

\subsubsection{Prediction Modeling Fundamentals}
Prediction modeling is a cornerstone of digital twin systems, providing the capability to forecast future states and behaviors of physical entities based on current and historical data. This capability is critical for optimizing operations, planning maintenance, and improving the overall performance and reliability of the system. This section introduces the concept of prediction modeling, provides a high-level mathematical definition, and presents relevant examples within digital twin systems.

\header{Definition of Prediction Modeling} Prediction modeling in digital twin system involves forecasting future states \( \mathbf{x}_{t+T} \in \mathcal{X} \) using observed data \( \mathbf{X}_t \subseteq \mathcal{X} \) and a prediction function \( \mathcal{F} \). 
Formally, given a set of observed states \( \mathbf{X}_t \) up to the current time \( t \), the goal is to estimate the future state \( \hat{\mathbf{x}}_{t+T} \) at a future time \( t+T \) as \( \hat{\mathbf{x}}_{t+T} = \mathcal{F}(\mathbf{X}_t, T) \), where \( \mathbf{X}_t = \{\mathbf{x}_{t_1}, \mathbf{x}_{t_2}, \ldots, \mathbf{x}_t\} \) represents the historical data up to the current time \( t \), and \( T \) is the prediction horizon, representing the time interval into the future for which the prediction is made.
The state space \( \mathcal{X} \) includes all possible states of the system, encompassing various sensor readings, operational conditions, and performance metrics. The observed data \( \mathbf{X}_t \) are the recorded states used to train the prediction models. The function \( \mathcal{F} \) employs machine learning, deep learning or statistical methods to predict future states based on historical and current data.
To optimize the predictive function \( \mathcal{F} \), the objective is to minimize a loss function \( \mathcal{L} \) that captures the difference between the predicted state \( \hat{\mathbf{x}}_{t+T} \) and the actual future state \( \mathbf{x}_{t+T} \) as:
\[ \mathcal{L} = \mathbb{E}[\ell(\mathbf{x}_{t+T}, \hat{\mathbf{x}}_{t+T})], \]
where \( \ell \) is the task-specific loss function (\emph{e.g.,} mean squared error \( \ell(\mathbf{x}, \hat{\mathbf{x}}) = (\mathbf{x} - \hat{\mathbf{x}})^2 \) in regression tasks). 

\subsubsection{Predictive Tasks}

In digital twin systems, prediction models play a crucial role in maintaining the reliability~\cite{Mubarak2022dig}, efficiency~\cite{Francisco2020sma}, and performance~\cite{Zhang2024dig} of both physical and cyber components. By leveraging advanced AI techniques, these models enable continuous monitoring and analysis of complex data~\cite{Guc2022sma, Liu2018the}, identifying patterns and forecasting potential issues~\cite{Zhu2020rea, Zhang2022bui, Zhu2020rea, Wang2022phy} before they manifest. The primary types of prediction in digital twin systems can be broadly categorized into two main areas: 1) Real-time Decision Making; 2) Predictive Maintenance. The following sub-sections will delve into these categories, outlining the key methods and tasks to highlight their importance and effectiveness in digital twin systems.

\header{Real-time Decision Making}
Real-time decision-making is crucial for digital twin systems, which create virtual replicas of physical entities to enable continuous monitoring and simulation. This capability allows for immediate analysis and response using current and historical data, supporting rapid issue identification, operational optimization, and improved system efficiency. Real-time decision-making in digital twin system spans various domains, including traffic management~\cite{li2019generic,li2020evolvegraph,cai2020traffic, Li2021tra, Liang2023spa, Kong2022dyn,li2019conditional,li2021spatio,cao2021spectral}, industrial logistics~\cite{Wu2022ind, Wang2022phy, Zhang2024dig}, healthcare~\cite{Pan2022tem, subramanian2022digital, elayan2021digital}, fire safety~\cite{Zhang2022bui, ding2023intelligent, zohdi2020machine}, structural health monitoring~\cite{Zhu2020rea, Francisco2020sma, Pan2020dig}, energy management~\cite{Li2023enh, Yi2023dig, Milton2020con}. The following sections explore key methodologies and tasks in these areas, highlighting how advanced AI techniques facilitate real-time decision making. a) \textit{For traffic management}, Cai et al.~\cite{cai2020traffic} developed a hybrid encoder-decoder neural architecture namely Traffic Transformer for traffic forecasting. A graph convolutional network is used to model spatial dependencies and a transformer is utilized to model temporal dependency in which novel temporal positional encoding strategies are proposed to encode the continuity and periodicity of time series.  
Li et al.~\cite{Li2021tra} developed a Multisensor Data Correlation Graph Convolution Network (MDCGCN) to address the challenges of real-time traffic flow prediction. This model effectively captures dynamic temporal and spatial correlations in traffic patterns, significantly improving the prediction accuracy. Similarly, Liang et al.~\cite{Liang2023spa} proposed the Spatial-Temporal Aware Data Recovery Network (STAR), which uses graph neural networks to recover missing entries in spatial-temporal traffic data. This approach ensures accurate data recovery critical for real-time traffic monitoring and decision making in Intelligent Transportation Systems (ITS). Kong et al.~\cite{Kong2022dyn} introduced the Dynamic Graph Convolutional Recurrent Imputation Network (DGCRIN), which models dynamic spatial dependencies and utilizes diverse data for imputing missing traffic data, thus enhancing real-time traffic data analysis. b) \textit{In industrial logistics}, Wu et al.~\cite{Wu2022ind} presented a system using the industrial Internet of Things and long short-term memory (LSTM) networks for real-time tracking of manufacturing resources. This system improves operational efficiency by accurately locating product trolleys and enabling location-based services. Wang and Ma~\cite{Wang2022phy} designed the PhysiQ framework to monitor physical therapy exercises at home. This system uses a multi-task spatiotemporal Siamese Neural Network to measure exercise quality, ensuring patients perform exercises correctly in real time. c) \textit{In fire safety field}, Zhang et al.~\cite{Zhang2022bui} proposed the Artificial-Intelligence Digital Fire (AID-Fire) system, which uses convolutional LSTM neural networks to identify and monitor fire evolution in real time, significantly aiding in firefighting and evacuation processes. d) \textit{In the medical field}, Pan et al.~\cite{Pan2022tem} developed the Temporal-based Swin Transformer Network (TSTNet) for real-time surgical video workflow recognition, achieving high accuracy by modeling temporal information and multi-scale visual data. e) \textit{In battery management field}, Li et al.~\cite{Li2023enh} presented a framework combining convolutional neural networks and LSTM for real-time degradation prediction of lithium-ion battery degradation in real-time, ensuring accurate battery health monitoring. Similarly, Yi et al.~\cite{Yi2023dig} proposed a method for real-time temperature prediction and degradation analysis of lithium-ion batteries, using LSTM networks to maintain battery safety and performance.

\header{Predictive Maintenance} Predictive maintenance is essential for ensuring the reliability and longevity of machinery and equipment across various industries. By utilizing advanced AI techniques, predictive maintenance enables accurate forecasting of equipment failures~\cite{Li2021hie, Aivaliotis2019the, Yang2022sup, Zhang2023fau, Peng2021ad, Liu2018the, Tuegel2011ree} and optimized maintenance schedules~\cite{Luo2020ah, Altun2019soc, Mubarak2022dig,Guc2022sma}.
 a) \textit{In the manufacturing sector}, Li et al.~\cite{Li2021hie} introduced a hierarchical attention graph convolutional network (HAGCN) that combines spatial and temporal dependencies to predict the remaining useful life (RUL) of machinery. Similarly, Aivaliotis et al.~\cite{Aivaliotis2019the} used physics-based simulations to estimate RUL for industrial robots, demonstrating the practicality of integrating digital models with real-time data. Moreover, Luo et al.~\cite{Luo2020ah} proposed a hybrid method that blends model-based and data-driven techniques for accurate CNC machine tool life prediction. 
b) \textit{For systems and equipment diagnostics}, several methods have been introduced. Yang et al.~\cite{Yang2022sup} presented SuperGraph, a spatial-temporal graph-based feature extraction method for rotating machinery fault diagnosis, which demonstrates significant advantages in handling complex data. Similarly, Zhang et al.~\cite{Zhang2023fau} implemented a fault prediction system for electromechanical equipment using multivariate spatial-temporal graph neural networks, which enhances predictive accuracy. In the realm of power electronics, Peng et al.~\cite{Peng2021ad} developed a non-invasive health indicator estimation method for DC–DC converters, employing particle swarm optimization to monitor key components effectively. Addressing challenges within the IoT ecosystem, Altun and Tavli~\cite{Altun2019soc} explored distributed ledger technologies to propose a model that improves security and scalability in predictive maintenance applications. Mubarak et al.~\cite{Mubarak2022dig} combined machine learning and advanced analytics to create a comprehensive predictive maintenance framework for Industry 4.0, optimizing maintenance decisions and improving cost-effectiveness.
c) \textit{In the aerospace industry}, Liu et al.~\cite{Liu2018the} highlighted the integration of various data sources to support decision-making processes, enhancing the efficiency of predictive maintenance. Similarly, Tuegel et al.~\cite{Tuegel2011ree} focused on reengineering aircraft structural life prediction using high-fidelity models ensuring greater accuracy and safety in assessing structural integrity.

\subsection{Detecting and Diagnosing Anomaly}

Anomaly detection (\emph{i.e.,} fault detection) is the process of identifying and responding to unusual patterns or behaviors within a system that deviates from the norm~\cite{chandola2009anomaly}. In digital twin systems, which create virtual replicas of physical entities, anomalies can manifest in various forms such as equipment malfunctions~\cite{Lv2023saf, Ghosh2019hid}, unexpected changes in operational performance~\cite{Shi2024dat, nashivochnikov2020system}, or irregular patterns in sensor data~\cite{Darvishi2023dee, Hasan2023was}. These anomalies can indicate underlying issues that may lead to significant network failures~\cite{Li2022ano, Yun2017dat}, safety hazards~\cite{Wang2022rea, Liu2024tow, Balta2024dig}, and financial losses~\cite{lu2020digital, chatterjee2024digital} if not addressed promptly. Therefore, anomaly detection is crucial for maintaining the integrity~\cite{Kumar2023blo}, reliability~\cite{Maleh2019mac}, and efficiency~\cite{trauer2021improving, schneider2018high} of the digital twin systems. By leveraging advanced anomaly detection algorithms (\emph{e.g.,} Autoencoders, GANs, and RNNs), digital twin systems can continuously monitor and analyze real-time sensor data and historical operational records to identify irregularities~\cite{Castellani2020rea, Shi2018ear, Lian2023ano}, predict potential failures~\cite{Xu2021dig,Hao2021hyb,Hu2023am}, and prescribe maintenance actions~\cite{Wang2021ano,Xu2019ad,Goh2017ano}, enabling timely interventions and preventive maintenance before costly breakdowns occur~\cite{Wu2023ad,russo2018anomaly}. This section will delve into three aspects, focusing on:  1) Anomalies in Digital Twin Systems; 2) Types of Anomaly Detection; 3) Anomaly Detection Methods.

\subsubsection{Anomaly Characterization}

Anomalies in digital twin systems are deviations from expected behavior that can indicate underlying issues in the physical or virtual components of the system~\cite{Xu2021dig, qin2024machine}. Understanding these anomalies is essential for developing effective detection and monitoring methods This section will introduce the concept of anomalies, provide mathematical definitions, and give examples relevant to digital twin systems.

\header{Definition of Anomalies in Digital Twin Systems} An anomaly in a digital twin system is any deviation from the expected behavior of the system, which can encompass various aspects such as data, processes, and performance metrics. Formally, given a system state space \( \mathcal{S} \), an anomaly is defined as a state \( s \in \mathcal{S} \) that significantly deviates from the expected behavior \( \mathcal{E} \subset \mathcal{S} \). This can be expressed as:
\[ s \in \mathcal{A} \iff \mathbb{P}(s \mid \mathcal{E}) < \epsilon, \]
where \( s \) represents a state of the system, such as a specific temperature reading or a performance metric like the speed of a manufacturing line. The state space \( \mathcal{S} \) includes all possible states the system can be in, encompassing every possible temperature, pressure, or performance metric. The expected behavior \( \mathcal{E} \) represents the subset of \( \mathcal{S} \) that includes normal operating states, such as temperature ranges between 50°C and 80°C. The set of anomalies \( \mathcal{A} \) includes states outside this expected range, such as a temperature reading above 90°C. The probability \( \mathbb{P}(s \mid \mathcal{E}) \) measures how likely \( s \) is to occur under normal conditions, and if this probability is lower than a predefined threshold \( \epsilon \), the state \( s \) is flagged as an anomaly.
This broad definition encompasses various types of anomalies including data anomalies, process anomalies, and performance anomalies, each indicating different potential issues within the digital twin system.

\subsubsection{Anomaly Detection Types and Methods}

In digital twin systems, anomalies can be categorized into several types based on their characteristics and impact on the system. The following subsections describe the main types of anomaly detection required in digital twin systems: 1) Data-driven Anomaly Detection; 2) System-based Anomaly Detection.

\header{Data-driven Anomaly Detection} Data-driven anomaly detection is essential in digital twin systems, where massive amounts of data from various sensors and systems are continuously transmitted and monitored. Anomalies in data can arise from numerous sources, including sensor malfunctions~\cite{Hu2023am, Darvishi2023dee, Hasan2023was,Goh2017ano}, network issues~\cite{Li2022ano, Yun2017dat, schneider2018high, russo2018anomaly}, and operational faults~\cite{Lian2023ano, Hao2021hyb, Shi2024dat, nashivochnikov2020system}. a) \textit{Sensor data anomalies} are often caused by sensor failures or inaccuracies in data collection. To address this, various anomaly detection techniques have been employed. For instance, Hu et al.~\cite{Hu2023am} proposed a masked one-dimensional convolutional autoencoder (MOCAE) for bearing fault diagnosis. The model leverages deep learning methods to enhance feature extraction and improve fault detection accuracy. Similarly, Darvishi et al.~\cite{Darvishi2023dee} introduced a deep recurrent graph convolutional architecture for sensor fault detection, isolation, and accommodation. This approach constructs virtual sensors to refurbish faulty data and uses a classifier to detect and isolate faults. Moreover, Hasan et al.~\cite{Hasan2023was} developed a Wasserstein GAN-based model for early fault detection in wireless sensor networks. This model uses Gramian angular field encoding to convert time series data into images, which are then processed by a GAN to detect anomalies, achieving high fault detection accuracy. b) \textit{Network data anomalies} arise from issues in data transmission and network performance (\emph{e.g.,} high latency), affecting the reliability and timeliness of the data received by digital twin systems. Li et al.~\cite{Li2022ano} addressed this challenge by detecting anomalies in internet service quality over fixed access networks. Their system aggregates data from multiple network elements and employs real-time simulations to detect service quality degradation and network faults. Moreover, the importance of data-centric middleware in large-scale digital twin platforms is emphasized in~\cite{Yun2017dat}. Their proposed architecture supports efficient data communication within digital twin systems, which is critical for accurate anomaly detection and system reliability. 
c) \textit{Operational data anomalies} occur due to irregularities or faults in the operational processes of systems. These anomalies can be particularly challenging to detect and diagnose due to the complex interactions between various system components. To address this challenge, many anomaly detection methods have been proposed. For example, Lian et al~\cite{Lian2023ano} proposed a method for detecting anomalies in multivariate time series data from oil and gas stations. Their MTAD-GAN approach combines knowledge graph attention with temporal Hawkes attention to accurately identify and interpret operational anomalies. In the context of power systems, Shi et al.~\cite{Shi2024dat} leverage random matrix theory and free probability theory for anomaly detection, accurately characterizing data correlations and effectively identifying anomalies in complex operations.
Moreover, Hao et al.~\cite{Hao2021hyb} developed a hybrid statistical-machine learning model that integrates SARIMA and LSTM for detecting anomalies in industrial cyber-physical systems. This model effectively identifies cyberattacks, malicious behaviors, and network anomalies with high accuracy and low computational complexity, making it suitable for real-time applications.

\header{System-based Anomaly Detection}
System-based anomaly detection in digital twin systems focuses on identifying and diagnosing faults and irregularities that arise from the overall system's operational processes~\cite{Wu2023ad, Lv2023saf, Ghosh2019hid, Xu2019ad, Wang2021ano} and network communications~\cite{Wang2022rea, Liu2024tow, Balta2024dig}.
These anomalies can stem from various sources, such as hardware malfunctions and cyber-attacks. We categorize these anomalies into two main types: a) Operational fault diagnosis; b) Cybersecurity and network communications. a) \textit{Operational fault diagnosis} involves detecting and diagnosing faults within the operational processes of systems, which can include machinery failures, process disruptions, and component wear and tear. For instance, Wu et al.~\cite{Wu2023ad} developed a multi-layer convolutional neural network for real-time fault diagnosis in high-speed train bogies, enhancing safety and reducing maintenance costs. Similarly, in the manufacturing sector, Lv et al.~\cite{Lv2023saf} designed a fault identification algorithm based on active learning and domain adversarial neural networks (DANN), which significantly improved fault diagnosis accuracy and stability under varying operational conditions. Ghosh et al.~\cite{Ghosh2019hid} constructed a system using hidden Markov models to encapsulate system dynamics and enhance fault diagnosis in manufacturing processes through improved understanding, prediction, and decision-making capabilities. Additionally, Xu et al.~\cite{Xu2019ad} proposed a two-phase fault diagnosis method using deep transfer learning, facilitating real-time monitoring and predictive maintenance by migrating trained models from virtual to physical spaces, thus maintaining operational continuity in dynamic production environments. Furthermore, Wang et al.~\cite{Wang2021ano} explored various classification models for anomaly detection in smart manufacturing, demonstrating the effectiveness of decision trees in achieving high fault classification accuracy and preventing operational disruptions. 
b) \textit{Cybersecurity and network communications} anomalies involve issues related to data transmission, network performance, and security threats, which can compromise the reliability and integrity of digital twin systems. To address these challenges, researchers have developed various approaches.
For instance, Wang et al.~\cite{Wang2022rea} utilized hidden Markov models and transfer learning to identify faulty components in virtual machines within NFV environments, enhancing fault recovery and ensuring secure network operations. Similarly, Liu et al.~\cite{Liu2024tow} proposed an autonomous trusted network framework that integrates data aggregation, security models, and intelligent configuration models to proactively detect and mitigate threats, thereby preventing security breaches and data loss. Moreover, Balta et al.~\cite{Balta2024dig} developed a robust defense mechanism for cyber-physical manufacturing systems. By distinguishing between expected anomalies and cyber-attacks using a combination of data-driven machine learning and physics-based models, they ensure the integrity of manufacturing processes and protect against malicious activities.

% \subsubsection{Anomaly Detection Methods}
To effectively detect and respond to anomalies that appear in both cyber and physical components of digital twin systems, a variety of advanced machine learning and deep learning models have been introduced~\cite{luo2021deep, qin2024machine}. These models enable the continuous monitoring and analysis of complex data, identifying patterns and predicting potential issues to ensure the reliability, integrity, and efficiency of the systems.

\header{Machine Learning-based Anomaly Detection}
Machine learning-based anomaly detection methods in digital twin systems employ a range of traditional algorithms to identify irregularities across various domains. These methods encompass clustering and distance-based techniques~\cite{Shetve2022ada, Zhang2019fau, Sarris2023tow, Abirami2023dig}, probabilistic approaches~\cite{Ademujimi2022dig, Ruah2022ab, Shi2018ear, Shi2018spa, Yu2020ad, Maleh2019mac}, as well as discriminative models~\cite{Gaikwad2020tow, Yin2016rec, Wang2021ano, noureen2019anomaly, almajed2022using, jeffrey2024using}, each offering unique strengths in handling high-dimensional data, capturing complex patterns, and providing robust solutions for real-time anomaly detection and fault diagnosis. a) \textit{Clustering and distance-based methods} are effective for applications requiring quick and adaptive responses to anomalies. Shetve et al.~\cite{Shetve2022ada} propose an adaptive N-step technique that integrates DBSCAN, Isolation Forest, and Local Outlier Filter to achieve high accuracy in detecting anomalies in smart manufacturing environments. Zhang et al.~\cite{Zhang2019fau} enhance k-nearest neighbors (kNN) by introducing weighted distances to improve fault detection in multimodal processes. In healthcare, Sarris et al.~\cite{Sarris2023tow} develop a K-means-based algorithm for brain tumor detection from MRI scans, while Abirami and Karthikeyan~\cite{Abirami2023dig} propose an optimized fuzzy k-nearest neighbor classifier for early Parkinson's disease identification.
b) \textit{Probabilistic-based methods} provide robust solutions for managing uncertainty and complex data relationships. Ademujimi and Prabhu~\cite{Ademujimi2022dig} utilize Bayesian Networks trained through a co-simulation approach for fault diagnostics in smart manufacturing systems. Ruah et al.~\cite{Ruah2022ab} present a Bayesian framework for anomaly detection in wireless systems, which addresses model uncertainty to enhance detection and data optimization. In distribution networks, Shi et al. apply a random matrix theory (RMT) based method for early anomaly detection using SCADA data~\cite{Shi2018ear} and a spatio-temporal correlation analysis approach for locating anomalies~\cite{Shi2018spa}. Yu et al.~\cite{Yu2020ad} propose a nonparametric Bayesian network for health monitoring, employing an improved Gaussian particle filter (GPF) and Dirichlet process mixture model (DPMM) for real-time updates. Maleh~\cite{Maleh2019mac} demonstrates the effectiveness of machine learning models in cybersecurity for IoT systems, achieving high accuracy even in constrained environments. c) \textit{Discriminative models}, such as support vector machines (SVM), are effective in defining decision boundaries for classification tasks. Gaikwad et al.~\cite{Gaikwad2020tow} integrate thermal simulations and sensor data in a machine learning framework using SVM to detect process faults in additive manufacturing. Yin and Hou~\cite{Yin2016rec} highlight the advantages of SVM in fault monitoring and diagnosis for complex industrial processes, emphasizing its generalization performance and suitability for small sample scenarios.

\header{Deep Learning-based Anomaly Detection}
Deep learning-based anomaly detection techniques have significantly advanced the capabilities of digital twin systems by leveraging sophisticated neural network architectures to detect complex patterns and irregularities in various domains.
These methods include autoencoders~\cite{Castellani2020rea, Hu2023am, Xu2019ad}, generative adversarial networks (GANs)~\cite{Xu2021dig, Hasan2023was, Lian2023ano}, convolutional neural networks (CNNs)~\cite{Danilczyk2021sma, Li2023an}, recurrent neural networks (RNNs)~\cite{Feng2021tim,Kumar2023blo}, and hybrid methods~\cite{Darvishi2023dee, Hao2021hyb}, each providing unique advantages in modeling high-dimensional data, capturing temporal and spatial dependencies, and ensuring robust anomaly detection and fault diagnosis. a)~\textit{Autoencoders} and their variants are particularly effective for unsupervised anomaly detection and feature extraction. Castellani et al.~\cite{Castellani2020rea} introduce a weakly supervised approach using Siamese Autoencoders (SAE) for industrial anomaly detection, outperforming state-of-the-art methods in various settings. Hu et al.~\cite{Hu2023am} propose a masked one-dimensional convolutional autoencoder (MOCAE) for bearing fault diagnosis, demonstrating superior performance on real bearing datasets. Similarly, Xu et al.~\cite{Xu2019ad} utilize deep transfer learning with a Stacked Sparse Autoencoder (SSAE) model to achieve high-fidelity fault diagnosis in dynamic manufacturing processes.
b)~\textit{Generative adversarial networks (GANs)} have proven effective in capturing complex data distributions and generating realistic data for anomaly detection. Xu et al.~\cite{Xu2021dig} present a GAN-based approach called ATTAIN for cyber-physical systems, utilizing GCN-LSTM modules to enhance anomaly detection capabilities. Hasan et al.~\cite{Hasan2023was} propose a Wasserstein GAN-based model for early drift fault detection in wireless sensor networks, achieving high accuracy in detecting sensor faults. Lian et al.~\cite{Lian2023ano} introduce a digital twin-driven MTAD-GAN for multivariate time series anomaly detection in oil and gas stations, leveraging attention mechanisms to improve detection performance. 
c)~\textit{Convolutional neural networks (CNNs)} are leveraged for their ability to handle spatial data and extract hierarchical features. For example, Danilczyk et al.~\cite{Danilczyk2021sma} utilize a CNN within the ANGEL Digital Twin environment to detect and classify faults in power systems using high-fidelity measurement data. Li et al.~\cite{Li2023an} employ a multidimensional deconvolutional network with attention mechanisms for real-time anomaly detection in industrial control systems. d)~\textit{Recurrent neural networks (RNNs)}, particularly LSTM networks, are adept at capturing temporal dependencies in sequential data. For instance, Feng and Tian~\cite{Feng2021tim} propose the NSIBF method combining neural system identification and Bayesian filtering for robust anomaly detection in cyber-physical systems. Kumar et al.~\cite{Kumar2023blo} integrate LSTM-SAE and BiGRU with self-attention mechanisms for secure communication in digital twin-empowered IIoT networks, enhancing intrusion detection capabilities. e) \textit{Other hybrid methods} like deep recurrent graph convolutional architectures have also been utilized to address sensor fault detection and accommodation. Darvishi et al.~\cite{Darvishi2023dee} present a deep recurrent graph convolutional architecture for sensor fault detection, isolation, and accommodation in large-scale networked systems. Additionally, hybrid approaches combining deep learning with other methods have been explored. Hao et al.~\cite{Hao2021hyb} develop a hybrid model integrating SARIMA and LSTM for real-time anomaly detection in ICS networks, providing high detection accuracy with low computational complexity.

\subsection{Optimizing and Controlling}

AI-enhanced optimization and control methods represent a significant advancement in managing complex systems, merging the capabilities of artificial intelligence with the detailed modeling and simulation afforded by digital twins~\cite{leng2021digital}. This integration results in smarter, more responsive, and predictive optimization and control mechanisms~\cite{olabi2023application,bost2024smart}. The enhancement of AI in digital twins for optimization and control systems can be seen in the following aspects, including AI-enhanced optimization and adaptive control.

\subsubsection{Optimization Strategies} 
Traditional mathematical programming-based optimization methods for digital twins often struggle to handle scenarios requiring rapid response or complex environmental changes that demand continuous adaptivity~\cite{chades2017optimization}. AI-enhanced optimization methods can address these issues to some extent~\cite{wang2020ai}. This section will be divided into two parts: the first part will discuss how AI improves response speed to enhance real-time optimization for digital twins, thereby enhancing system performance and reducing resource consumption; the second part will summarize how AI enhances adaptivity, leading to better robustness and improved handling of uncertainties in adaptive optimization tasks for digital twins. 

\header{Real-time Optimization}
Real-time Optimization refers to the immediate adjustment of physical systems based on real-time sensor data and simulation results to optimize system performance. In related tasks that require immediate decision-making and fast response, AI enhances response speed and decision efficiency, improving the quality of decisions. For instance, early work~\cite{powell2020real} presents a reinforcement learning-based real-time optimization (RL-RTO) methodology for process systems, which integrates optimal decision-making into a neural network, contrasting with traditional repeated process model solutions. The RL-RTO approach, demonstrated using a chemical reactor, shows potential by improving annual profit by 9.6\%, though it still lags behind the conventional first principles plus nonlinear programming method which achieved a 17.2 \% improvement. Further, Dong-Hoon Oh et al.~\cite{oh2021actor} introduce an actor-critic reinforcement learning strategy for optimizing hydrocracking unit operations, developed from a validated mathematical model with less than 2\% error. The approach achieved optimal operating conditions with 97.86\% and 98.5\% accuracy, demonstrating quick response times, low computational burden, and high customizability, suitable for practical online optimization and adaptable to other chemical industries. More recently, researchers~\cite{li2024ai} have addressed the challenges of optimization in fed-batch biopharmaceutical processes by proposing an RL framework that incorporates human knowledge. Verified through the domain-specific simulator, the RL scheme enhances batch yields by 14\% with minimal online computation time, demonstrating significant potential over existing methods. Additionally, a series of studies~\cite{muller2020dynamic,qi2024real,pan2023real} have explored using RL to generate real-time optimal decisions for digital twins. In summary, RL provides the benefit of training an optimal policy rather than optimizing actions at each time step. Once the optimal policy is identified, a cost-effective forward pass through the function can quickly produce an online solution, enabling faster decision-making. 

\header{Adaptive Optimization}
Adaptive optimization is a method that dynamically adjusts optimization parameters or strategies in response to real-time feedback and changing conditions to continually improve performance and handle uncertainties. However, traditional mathematical programming methods often involve relatively static models that struggle to repeatedly reconfigure or adjust parameters to handle complex and changing scenarios~\cite{wang2020adaptive}. By collecting real-time data and making corresponding adjustments, AI can better address these tasks. In the early days, research efforts focused on using traditional machine learning methods to achieve better adaptive optimization~\cite{reker2020adaptive,joly2019machine}. Although the above results and methods can address optimization of process parameters to a certain extent, their dynamics are poor and they cannot well adapt to changing environments~\cite{wang2020adaptive}. Liu et al.~\cite{liu2022digital} proposes an AI-based adaptive optimization method for predicting surface roughness and adaptively adjusting process parameters in parts machining, addressing the limitations of traditional optimization methods in handling real-time and uncertain factors. By constructing a digital twin and utilizing a combination of  Particle Swarm Optimization and Generalized Regression Neural Networks, the method enables real-time monitoring, prediction of tool wear and surface roughness, and adaptive optimization of cutting parameters, thereby improving both quality and efficiency in the machining process. More recently, Yang et al.~\cite{yang2024adaptive} present an adaptive optimization method combined with federated learning, which leverages the advantages of federated learning in handling heterogeneous data. This method improves convergence speed by over 60\% and reduces traffic consumption by over 60\%. By integrating various AI technologies, it is possible to more effectively overcome the limitations of traditional optimization methods in terms of dynamism, thereby enhancing adaptive optimization and improving this task for digital twins.

\subsubsection{Adaptive Control}
Adaptive control is a technique used to improve the efficiency or reduce the resource costs of physical systems by leveraging real-time simulation results and sensor data. Specifically, through enhancements of reinforcement learning, adaptive control plays a crucial role in various applications, including manufacturing equipment regulation~\cite{he2019data}, robotics~\cite{kuts2019digital}, and autonomous driving~\cite{veledar2019digital}. In this section, we focus on introducing the most representative AI technology utilized in control strategies - Reinforcement Learning (RL), and summarizing its advantages over traditional control techniques like Proportional-Integral-Derivative (PID) Control and Model Predictive Control.

\header{Reinforced Learning and PID Control}
In the control tasks for digital twins, the most traditional and widely used controller is the PID controller. However, PID control struggles to handle high-dimensional, highly nonlinear, and time-varying systems. By integrating with Reinforcement Learning techniques, these challenges can be addressed to a certain extent. One approach involved using RL agents for parameter tuning of PID controllers. Early attempt~\cite{carlucho2017incremental} used an incremental Q-Learning strategy to tune the PID controller in an online fashion. This algorithm dynamically grows a Q-value table in both the action space and state space direction through discretization techniques, in order to create a discrete yet accurate tuning model. Other methods have employed the Continuous Action RL Automata algorithm and a Radial Basis Function Actor-Critic network to tune a PID controller. Nevertheless, these methods do not fully resolve the linearity issues inherent to PID controllers. Furthermore, efforts have been made to entirely replace PID controllers with RL. One study~\cite{hu2019intelligent} successfully implemented the DDPG algorithm to develop an intelligent control strategy for the transient response of a variable geometry turbocharger system. Similarly, other researchers~\cite{kafkes2021developing} have proposed RL models as substitutes for PID controllers.  

\header{Reinforced Learning and Model Predictive Control}
Model Predictive Control (MPC) is also a widely used control technique, particularly suitable for control systems that require anticipatory actions. It is extensively employed in applications such as autonomous driving and robotic control. While MPC can handle complex, multivariable systems, it faces challenges with computational complexity and model accuracy. Integrating RL with MPC offers promising solutions to these issues. One approach involves using RL to improve the performance of MPC controllers. A digital twin-based adaptive controller has been proposed that integrates software-in-loop (SIL) and hardware-in-loop (HIL) simulations~\cite{zhang2021digital}. This method allows for real-time optimization of the MPC parameters, enhancing its ability to handle nonlinear and time-varying systems. Another strategy employs RL to directly replace traditional MPC algorithms. Researchers have developed DQN policy models capable of regulating complex systems against realistic time-varying perturbations~\cite{gros2020data}. These RL-based controllers can adapt to changing system dynamics without requiring explicit model updates, a significant advantage over traditional MPC. Digital twins play a crucial role in the development of RL-MPC systems. By creating accurate virtual representations of physical systems, digital twins provide safe environments for training and testing RL algorithms before deployment. For instance, an LSTM network has been used to capture the full dynamics of a control system, serving as a digital twin for RL training~\cite{wang2020digital}. The integration of RL and MPC has shown promising results for the control tasks of digital twins.

\section{Towards Autonomous Management of Digital Twins}\label{sec:6}
% \subsection{Cognitive Foundations for Autonomy}\
\subsection{Cognitive Capabilities for Autonomous Management}
% \header{Semantic Understanding of Digital Twin States}
% \header{Natural Language as a Control Interface}
% \header{LLMs as Autonomous Coordinators}

Autonomous management relies on the cognitive layer of the digital twin, which allows it to understand human intentions and interpret the dynamic state of the system it manages. These cognitive capabilities form the foundation of intelligent control: the twin must first comprehend what needs to be achieved and perceive how the system is currently behaving. Large language models provide the mechanism for understanding and translating natural language commands into management operations, while foundation models enable perception across multiple data modalities. Together, they establish the semantic bridge between human intent, environmental understanding, and autonomous decision-making. It should be noted that current LLM- or diffusion-based world models do not guarantee physical fidelity or closed-loop stability, and their role in digital twins remains largely exploratory.

\subsubsection{Natural Language Interaction with LLMs}
Natural language interaction enables digital twins to understand and respond to human instructions in an intuitive way~\cite{openai2023gpt4}. Instead of relying on predefined scripts or manual configuration, operators can issue management goals directly using ordinary language, such as “reduce energy consumption without slowing production” or “check whether the network is operating normally.” The large language model interprets these instructions, extracts actionable entities, and translates them into formal goals that the twin can execute~\cite{wei2022chain}. This natural interface transforms digital twin management from a highly technical process into an accessible and collaborative activity~\cite{bommasani2021opportunities}.

\header{From Language to Management Decisions}
The ability to transform human language into system-level actions represents a central cognitive advance for autonomous management. Large language models can parse free-form text into structured intents, identifying the relevant variables, constraints, and objectives involved in a task. They then map these structured goals to internal control modules or simulation tools through schema alignment and function invocation~\cite{schick2023toolformer}. For example, when a user requests “optimize throughput while keeping temperature below 70°C,” the model converts the phrase into measurable objectives that define optimization targets and boundary conditions for the planning module.
This process is reinforced through retrieval-augmented reasoning, where the model grounds its decisions in real-time system data and prior management records~\cite{lewis2020retrieval}. By incorporating contextual information from sensor logs, historical performance data, and configuration files, the model avoids speculative reasoning and ensures that generated management actions are aligned with the current state of the physical system. As a result, language ceases to be a vague or ambiguous form of communication; it becomes a direct, interpretable interface between human expertise and machine execution.
More importantly, this transformation also enables adaptability. When the system receives partially defined or conflicting goals, the model can request clarification, negotiate constraints, or infer missing details based on historical patterns. This bidirectional reasoning loop ensures that human intent is faithfully translated into precise operational instructions, supporting both efficiency and safety in autonomous control.

\header{Conversational Management Interfaces}
Beyond one-time instructions, large language models enable ongoing dialogue between humans and digital twins. Through conversational interfaces, the system can provide summaries of its current state, report ongoing operations, and justify its planned decisions~\cite{amodei2016concrete}. For instance, after executing a maintenance optimization, the twin might respond: “Cooling power has been reduced by 10\%, energy consumption decreased by 12\%, and no overheating detected.” Such transparency allows users to remain informed without micromanaging the process.
Continuous conversation also allows for dynamic collaboration. Users can refine goals in natural language, such as “reduce the temperature faster but avoid overshoot,” and the system immediately adjusts its plan, re-evaluating trade-offs through its internal reasoning module~\cite{bommasani2021opportunities}. This adaptability turns management into a continuous negotiation between human and machine, where language serves as the medium for shared understanding.
In addition, conversation provides a mechanism for accountability. The twin can explain why it made a particular decision, cite the data used, and quantify uncertainty in its predictions. These dialogue-based explanations not only enhance trust but also provide a foundation for regulatory transparency and auditability, which are essential in safety-critical domains such as manufacturing and infrastructure management.

\subsubsection{Multimodal Perception with Foundation Models}
A digital twin’s ability to manage autonomously depends not only on understanding instructions but also on perceiving its environment accurately. Foundation models extend perception beyond individual sensors by learning joint representations across images, signals, and text~\cite{radford2021learning}. This multimodal capability allows the twin to interpret complex operational contexts-identifying patterns, recognizing anomalies, and inferring hidden conditions that may not be apparent from a single data stream~\cite{alayrac2022flamingo}. In doing so, it bridges the gap between physical phenomena and digital awareness.

\header{Multimodal Data Understanding}
Foundation models learn to interpret diverse data modalities by encoding them into a shared semantic space. They integrate information from cameras, vibration sensors, acoustic signals, thermal arrays, and textual logs to build a holistic understanding of the system’s state~\cite{li2023blip2}. For example, a model may correlate a faint noise pattern with an abnormal vibration frequency, identifying a potential mechanical imbalance before traditional diagnostics detect it. This enables a form of perceptual intelligence that moves beyond static thresholds toward contextual reasoning.
By abstracting multimodal signals into interpretable embeddings, the model can recognize system configurations, operating modes, and early indicators of degradation. This capability enhances situational awareness and provides a richer foundation for prediction and control. As environments evolve, the model continues to adapt through incremental fine-tuning, allowing perception to remain accurate even under new conditions or sensor configurations~\cite{bommasani2021opportunities}.
Multimodal data understanding supports semantic consistency between the physical and digital layers. When the system observes discrepancies, such as when sensor readings suggest stability but video data reveals irregularities, it can flag these inconsistencies for review. This capacity to cross-validate signals across modalities strengthens reliability and prevents blind spots in autonomous management.

\header{Multimodal Fusion for System Awareness}
Fusion mechanisms combine diverse streams of information into a unified situational representation that reflects the system’s overall status~\cite{bommasani2021opportunities}. Through spatial and temporal alignment, the twin synthesizes localized signals into a coherent global view. For instance, data from distributed sensors across a power grid can be fused with satellite imagery and maintenance logs to infer both operational performance and environmental stress patterns. Such integrated awareness enables the twin to reason about interactions among components rather than treating each subsystem in isolation.
This fusion-driven awareness also supports proactive behavior. The digital twin can detect emerging patterns, anticipate cascading effects, and adjust management strategies in advance. By tracking correlations across modalities, it learns causal relationships that link physical states to operational outcomes. The result is not just richer perception but an embodied understanding of the system’s dynamics.
Finally, maintaining awareness requires the twin to represent uncertainty transparently. Foundation models quantify prediction confidence and associate each inference with traceable evidence~\cite{kuleshov2018accurate}. This information is passed to the planning and decision modules, ensuring that high-risk judgments are handled conservatively or deferred to human oversight. In this way, multimodal fusion transforms raw data into actionable awareness, serving as the perceptual counterpart to the reasoning and planning processes that define autonomous management.

\subsection{Enabling Autonomous Decision and Planning}
While cognitive capabilities enable a digital twin to understand intentions and perceive the system state, autonomous management further requires the capacity to make decisions, plan actions, and adapt to changing conditions without explicit human intervention. This section discusses how agent-based reasoning and adaptive learning provide the mechanism for decision-making, and how self-optimization and closed-loop control complete the cycle of autonomous management. Together, these elements operationalize the MAPE-K paradigm, enabling digital twins to continuously monitor, analyze, plan, execute, and refine their management strategies.

\subsubsection{Agent-based Reasoning and Adaptive Learning}
Agent-based reasoning endows the digital twin with a modular structure capable of acting intelligently in complex environments~\cite{wooldridge2009introduction}. Each agent embodies autonomy, perception, reasoning, and learning, functioning as both a decision-maker and an executor of management tasks. Within a digital twin system, agents perceive environmental inputs, analyze contextual information, generate management plans, and execute actions through interactions with the underlying physical or simulated systems. Reinforcement and continual learning further allow these agents to refine strategies from experience and coordinate with others in multi-agent settings~\cite{silver2021reward,ma2021continual,zhang2025lamma}.

\header{Agent Architecture for Autonomous Management}
The agent architecture forms the operational backbone of autonomous management. Typically organi2zed according to the MAPE-K loop, an agent continuously monitors system states, analyzes patterns, plans interventions, executes actions, and updates its knowledge base~\cite{kephart2003vision}. This loop transforms management from reactive to proactive, allowing the digital twin to maintain stable operation even under uncertainty.  
The architecture usually comprises four layers: a perception layer that collects multimodal signals, an analysis layer that diagnoses conditions or predicts outcomes, a planning layer that formulates adaptive strategies, and an execution layer that interfaces with actuators or simulation modules~\cite{russell2020artificial}. These layers interact bidirectionally, ensuring that every action is grounded in current observations and that outcomes feed back into learning and optimization.
A practical example is the autonomous operation of industrial plants. Agents continuously assess performance metrics, detect bottlenecks, and coordinate actions such as adjusting temperature or resource allocation~\cite{parunak2001applications}. When the system deviates from desired performance, the planning layer proposes new configurations and the execution layer implements them automatically. The knowledge component stores contextual rules, such as causal relationships between control variables and outcomes, enabling the system to reason about “why” certain strategies work and reuse them in future scenarios.
Furthermore, agent-based frameworks support hierarchical and distributed management. Local agents handle subsystem optimization, while higher-level agents coordinate global objectives, ensuring alignment between individual actions and system-wide performance. This organization enables scalability and robustness, as local failures can be compensated by neighboring agents without central control~\cite{wooldridge2009introduction}.

\header{Learning from Management Experience}
Adaptive learning transforms the digital twin from a static control system into a self-improving entity. Through reinforcement learning (RL), agents learn management policies that maximize long-term rewards rather than immediate performance~\cite{mnih2015human}. Each experience, including success, failure, or anomaly, contributes to refining these policies, allowing the twin to anticipate the consequences of its actions in similar future contexts. Over time, this continuous improvement leads to resilient and efficient management strategies.
For instance, consider an agent managing energy consumption in a data center. Initially, it may explore different cooling strategies through simulation. As it accumulates feedback on temperature stability, cost, and latency, it gradually learns an optimal control policy that balances energy efficiency and performance~\cite{silver2021reward}. This process mirrors human expertise acquisition—learning not only from positive outcomes but also from mistakes that inform better decisions.
In addition to single-agent learning, collaborative learning across multiple digital twins enhances global intelligence. When experiences from one system are shared with another through federated or transfer learning, the collective knowledge base grows~\cite{zhu2021federated}. This enables rapid adaptation to new environments without starting from scratch, promoting a form of “organizational memory” for complex infrastructure networks.  
Ultimately, learning from management experience allows digital twins to evolve beyond pre-programmed behaviors, enabling self-adaptation in dynamic, uncertain, and data-rich environments.

\subsubsection{Self-Optimization and Closed-Loop Control}
Self-optimization is the culmination of autonomous management, where digital twins no longer rely on external commands but continuously refine their performance through closed-loop feedback~\cite{qin2022survey}. The system observes its own behavior, identifies inefficiencies, and implements corrective actions automatically. When combined with predictive and cognitive capabilities, closed-loop control transforms the twin into an autonomous entity capable of sustaining optimal performance with minimal supervision.

\header{Autonomous Closed-Loop Management}
Closed-loop management completes the autonomous control cycle by continuously connecting perception, reasoning, and execution~\cite{astrom2021feedback}. In this paradigm, the digital twin monitors real-time data, detects deviations from expected performance, and triggers self-corrective actions. The MAPE-K loop becomes operational in a real-time context: Monitor to capture state data, Analyze to detect anomalies or predict outcomes, Plan to generate interventions, Execute to apply adjustments, and Knowledge to update future strategies~\cite{kephart2003vision}.  
For example, in smart manufacturing, when vibration sensors indicate potential equipment fatigue, the digital twin predicts the failure horizon and autonomously schedules maintenance before a breakdown occurs~\cite{jiang2021digital}. The system then evaluates the effectiveness of its intervention, learning from the outcome to improve future responses. This cycle of detection, action, and refinement ensures continuous adaptation to changing operational conditions.
The advantage of autonomous closed-loop management lies in its capacity for sustained performance optimization. Unlike open-loop systems that depend on periodic calibration or manual adjustment, the closed loop enables continuous learning, model recalibration, and policy adaptation. Over time, the twin becomes more proficient in managing itself, bridging the gap between simulation-based optimization and real-world operational autonomy~\cite{qin2022survey}.

\header{Human-AI Collaborative Management}
Despite advances in autonomy, complete independence from human oversight is rarely desirable or safe. Human-AI collaboration remains essential for balancing efficiency with accountability~\cite{seeber2020machines}. Collaborative management frameworks define different autonomy levels—from fully manual control to fully autonomous operation—allowing human intervention when uncertainty or risk exceeds predefined thresholds.  
At intermediate autonomy levels, the digital twin acts as an intelligent assistant that recommends actions, explains its reasoning, and executes tasks upon approval. This cooperative workflow enhances human decision-making by offloading complex computations while preserving transparency and trust~\cite{gunning2019darpa}. For instance, in power grid management, the twin may autonomously adjust voltage to stabilize supply but defer load-shedding decisions to human operators when the consequences are ethically or economically sensitive.
The collaboration also extends to learning. Human experts can guide the twin by providing feedback on system responses or labeling exceptional cases for model retraining~\cite{seeber2020machines}. This hybrid feedback accelerates learning convergence while ensuring that autonomous management aligns with human values and regulatory constraints. Ultimately, the goal of human-AI collaborative management is not to replace human judgment but to augment it-creating digital twins that are not only self-managing but also self-explaining and accountable within human-centered systems. 

\section{Applications}\label{sec:7}
\subsection{Healthcare System}
% Hospital System, Clinical Trials
The rapid adoption of electronic health records (EHRs), along with the emergence of digital and smart healthcare, has accelerated the integration of diverse technologies aimed at optimizing healthcare operations, improving patient outcomes, and reducing healthcare costs~\cite{tian2019smart,baker2017internet,catarinucci2015iot}. Among these innovations, the digital twin technology plays a significant role by enabling the simulation of complex systems and the integration of virtual representations of real-world entities with AI to transform healthcare~\cite{sun2023digital,liu2019novel, sahal2022personal,croatti2020integration}. In the following sections, we ourlined several major applications of digital twin (DT) technology across key healthcare domains, including clinical practice, clinical research, drug discovery, disease modeling, and precision medicine~\cite{machado2023literature, khan2022digital}. 

\header{Clinical Workflow Optimization} As hospitals transition from conventional practice to technology-driven information systems, digital twins are becoming essential for monitoring operations and managing resources~\cite{erol2020digital,peng2020digital}. By integrating real-time and historical EHR data to simulate clinical resource usage and system-level processes, digital twins serve as adaptive virtual environments for operational decision-making. These models go beyond simple prediction by enabling healthcare administrators to test intervention scenarios and prescribe data-driven strategies to optimize staffing, bed utilization, and equipment deployment.~\cite{erol2020digitaltrans,aluvalu2023novel}. This simulation-based decision support ensures that resources are used optimally, reducing waste and operational costs. For example, Siemens Healthineers utilized digital twins in the Mater Private Hospital in 2018 to simulate MRI and CT workflows, identifying potential improvements and significantly enhancing the patient experience~\cite{SiemensHealthineers_2018}. Moreover, digital twins facilitate scenario planning and decision-making, allowing hospitals to prepare for various contingencies such as sudden influxes of patients or changes in healthcare policies~\cite{han2023digital,zhong2022multidisciplinary}. This leads to enhanced flexibility, resilience, and overall quality of care within the hospital environment. Karakra et al. introduced a digital twin model based on discrete event simulation to evaluate the efficiency of current healthcare delivery systems and assess the impact of service changes without disrupting daily operations~\cite{karakra2018}. In 2019, Karakra et al. further developed HospiT’Win, a virtual hospital replica that enables healthcare providers to track patient pathways, monitor behaviors, and predict future outcomes~\cite{karakra2019}.

\begin{figure}[H]
    \centering
\includegraphics[width=0.85\linewidth]{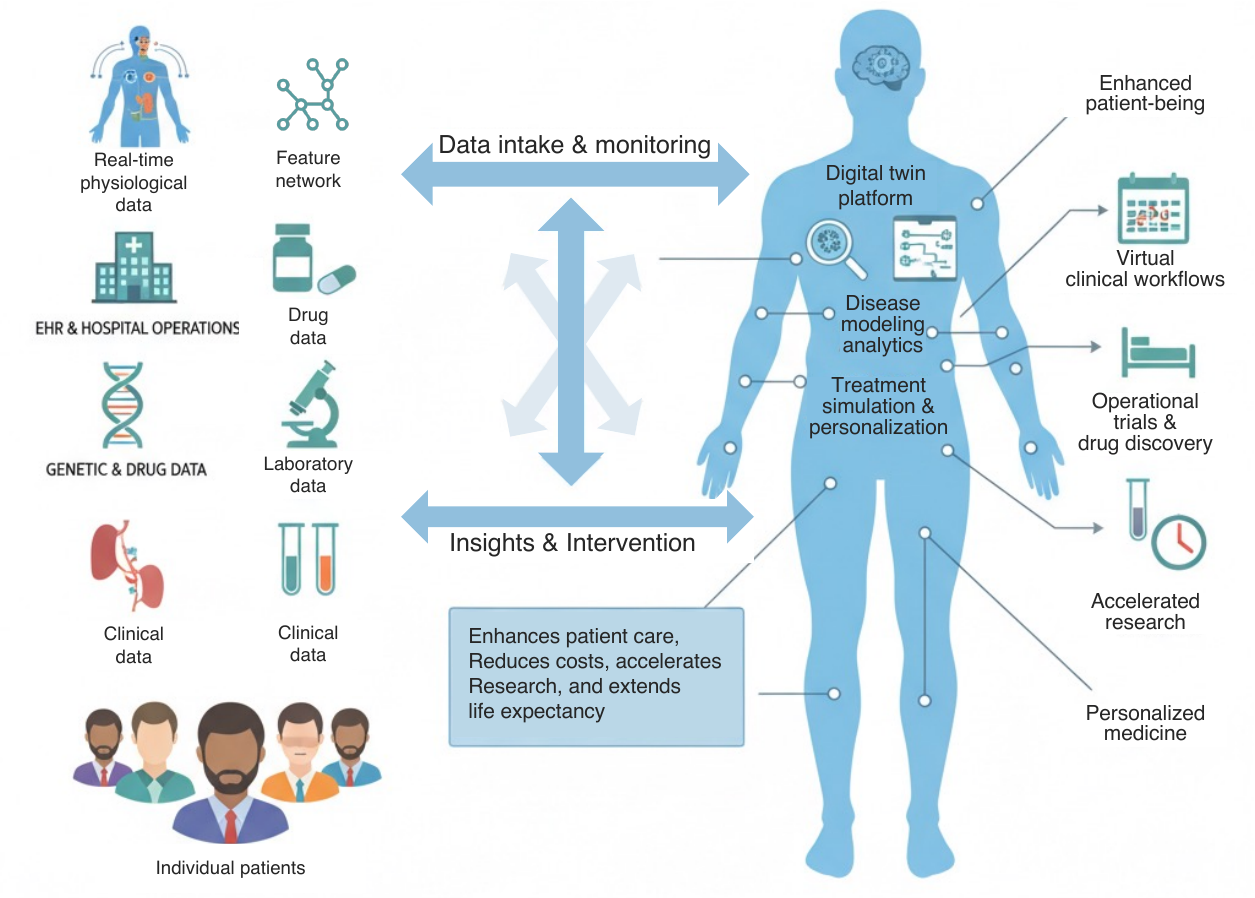}
    \caption{\textbf{Patient-centric digital twin framework for healthcare.}
Multi-source patient data, including real-time physiological signals, electronic health records, clinical measurements, laboratory results, genetic and drug data, are continuously ingested into a digital twin platform through data intake and monitoring. The digital twin enables disease modeling and analytics as well as treatment simulation and personalization, generating actionable insights that support clinical decision-making and intervention. By closing the loop between data, modeling, and intervention, the framework facilitates virtual clinical workflows, operational trials and drug discovery, and accelerated research, ultimately improving patient care, reducing costs, and enabling personalized medicine.}
    \label{fig:app-healthcare}
\end{figure}

\header{Clinical Trial} In addition to streamlining healthcare workflows, the digital twin also significantly contributes to advancements in clinical research~\cite{barbiero2021graph}. It can offer transformative potential to augment the design and emulation of clinical trials, addressing persistent barriers such as absent control arms, small cohorts, and limited generalizability, and thereby advancing evidence generation in medical research~\cite{venkatesh2022health}. For example, Digital twins use advanced AI and machine learning models to create virtual patients, simulating counterfactual outcomes and allowing trials to maintain statistical power with fewer participants~\cite{thangaraj2024novel,grieb2023digital}. This approach has potential to reduce the duration and cost of clinical trials by enabling concurrent simulations of multiple scenarios, thus reducing the need for extensive human trials and accelerating the development of new treatments~\cite{li2024construction}. For example, in 2021, Qian et al. proposed SyncTwin, a method that learns a patient-specific time-constant representation from pre-treatment observations to create a digital twin that closely matches the target patient, enabling accurate counterfactual predictions~\cite{qian2021synctwin}. Similarly, in 2023, Das and Wang et al. introduced TWIN, a model that enhances clinical trials by creating personalized virtual models of patients using large language models~\cite{das2023twin}. These digital twins simulate patient health trajectories and predict clinical outcomes based on real-world data, demonstrating the significant advancements in trial efficiency and precision that digital twin technology can offer.

\header{Drug Discovery and Development} Drug discovery involves identifying potential new medications, understanding their mechanisms, and optimizing their efficacy and safety before they can proceed to human testing~\cite{dugger2018drug, mak2019artificial, tamimi2009drug}. The drug discovery and development process using digital twin technology involves creating virtual models of biological systems to simulate disease mechanisms and drug interactions~\cite{an2022drug,mariam2024unlocking}. It usually begins with the identification and validation of therapeutic targets by modeling cellular processes and interactions. Drug candidates are then screened through simulated interactions with these targets, allowing for optimization before preclinical testing~\cite{talukder2022drugomics, chen2020digital, li2022dynamic}. Digital twins in silico trials can help predict drug behavior and potential side effects, enhancing trial design and patient stratification in clinical stages~\cite{young2024design,sinisi2020artificial}. This method accelerates drug development, reduces costs, and improves success rates by providing a detailed understanding of drug behavior within a virtual biological system. In 2023, Moingeon highlights the utilization of digital twin in drug development against autoimmune disease, where digital twins of human cells are created using multi-omics data, including detailed information at the single-cell level~\cite{moingeon2023artificial}. This data can be combined with machine learning to predict how cells will respond to different drug candidates, enabling rapid testing of billions of drug combinations. This not only accelerates target identification and optimization but also facilitates drug repurposing for novel therapeutic applications. As another example, Subramanian utilized digital twin to create a comprehensive virtual model that simulates liver function, disease progression, and treatment responses~\cite{subramanian2020digital}. This model integrates data from experimental studies and mathematical frameworks, allowing researchers to predict drug-induced liver injury and optimize drug candidates before clinical trials~\cite{subramanian2020digital}. 

\header{Disease Modeling, Diagnosis and Prognosis} With its ability to create dynamic, patient-specific models, digital twin technology plays a critical role in advancing precision medicine by supporting early disease detection and trajectory forecasting~\cite{ferdousi2021digital}. Digital twins can simulate a wide range of physiological and pathological processes by integrating data from multiple sources, such as medical imaging, electronic health records, genetic information, and real-time sensor data~\cite{mourtzis2021smart,abirami2023digital}. They can replicate the functioning of organs, track the development of diseases, and predict how a disease will progress in a specific patient. The ability to simulate and analyze various medical scenarios without invasive procedures represents a significant advancement in medical technology and patient care~\cite{hussain2021healthcare, sun2023digital2, sasikaladevi2024digital}. By running simulations, healthcare providers can test various scenarios, such as the impact of different medications or lifestyle changes, without any risk to the patient~\cite{kulkarni2024hybrid}. This helps in early detection of potential health issues and proactive intervention. In 2024, Wang et al. applied digital twin technology to simulate the inhomogeneous optical properties of multi-core fiber, achieving unpixelated, high-resolution tumor imaging to enhance cancer diagnosis~\cite{wang2024resolution}. Surian et al. utilized clinical and physiological biomarker data from various cohorts to develop a digital twin that simulates metabolic health profiles~\cite{surian2024digital}. This model accurately predicts the onset of chronic kidney disease (CKD) over a three-year period and effectively stratifies patients into different risk categories, facilitating early detection and improving management strategies for CKD. Similarly, Venkatapurapu et al. integrated a mechanistic model of Crohn's disease (CD) and digital twin of each patient to predict the temporal progression of mucosal damage and healing, which provides detailed and chronological predictions of disease dynamics to support treatment decisions~\cite{venkatapurapu2022computational}.

\header{Personalized Treatment} Personalized treatment involves tailoring medical interventions to a patient's specific genetic, environmental, and behavioral characteristics, offering a more targeted and effective approach to care.~\cite{kosorok2019precision,ashley2016towards}. Similar to applications in disease diagnosis, digital twin technology enhances personalized treatment by creating virtual replicas of individual patients using data from various sources, but it focuses on comprehensive health insights and the detailed customization of treatment plans~\cite{corral2020digital,coorey2021health}. By incorporating a patient’s genetic information, digital twins can simulate how different medications will interact with the individual's unique genetic makeup, predicting potential side effects or adverse reactions before administering the treatment, thus increasing the efficacy of treatments and minimizing adverse effects~\cite{vallee2024envisioning, venkatesh2022health,dedomenico2024challengesopportunitiesdigitaltwins}. Moreover, digital twins support continuous monitoring and adjustment of treatment plans, ensuring that care remains optimal over time~\cite{elshaier2022proposed}. For example, Martinez-Velazquez et al. constructed Cardio Twin, an architecture based on digital twin technology that integrates data from smartphone sensors, external devices, social networks, and medical records for real-time analysis~\cite{martinez2019cardio}. It optimizes personalized treatment for patients with cardiovascular diseases by analyzing patient-specific data to predict drug interactions and customize therapies, while autonomously managing lab data transmission and emergency service requests. Similarly, Wickramasinghe et al. adopted digital twin technology in personalized care for patients with uterine endometrial cancer, analyzing extensive patient data to develop predictive models for cancer progression and personalized treatment responses rather than relying on predefined theoretical frameworks~\cite{wickramasinghe2021vision}. The clinician support system based on digital twins addresses the complexities and cognitive limitations associated with traditional cancer care by leveraging AI and machine learning to process large datasets, thereby enhancing the precision and efficacy of cancer treatment.

\subsection{Biological System}

The application of digital twin technology in biological systems represents a groundbreaking advancement with the potential to revolutionize various fields within life sciences~\cite{barnabas2020human,quinn2023digital}. By creating precise virtual replicas of biological entities, digital twins enable researchers to simulate, analyze, and predict the behavior of complex biological processes in real-time~\cite{sigawi2023using,herwig2021digital,moller2021digital,zhao2023biodt}. This innovative approach facilitates a deeper understanding of molecular interactions, cellular dynamics, and anatomical functions~\cite{chude2021conceptual,baumgartner2022world,kalozoumis2022towards}. It enhances research and development, improves diagnostic techniques, optimizes experimental and therapeutic interventions, and streamlines production processes in biomanufacturing~\cite{yang2023sensing,viceconti2023position}.

\header{Molecular Research} Application of digital twin technology can expand to various levels in biological systems~\cite{li2021digital2, sigawi2023using}. At the molecular level, digital twins can simulate the interactions between proteins, nucleic acids, lipids, and other biomolecules with high precision~\cite{kearns2024application, miehe2021reprint}. These simulations encompass detailed molecular dynamics, including protein-protein interactions, DNA-RNA transcription and translation processes, lipid bilayer formation and behavior, and complex molecular assemblies~\cite{topolsky2022information,zifarelli2023multivariate,rahman2022explore, gerogiorgis2021digital}. Consequently, digital twins provide comprehensive insights into complex biochemical pathways and molecular mechanisms, facilitating the development of targeted therapies, personalized medicine, and biotechnological innovations~\cite{udugama2021digital, appl2021development}. For example, Hengelbrock et al. utilized digital twins for virtual replicas of the physical mRNA transcription process to simulate and optimize production conditions~\cite{hengelbrock2024digital}. By determining key kinetic parameters and using a plug flow reactor for high-throughput screening, the researchers significantly enhanced mRNA yield and reduced impurities like truncated mRNA. This approach enabled the efficient production of 20 vaccine candidates in a short time, a tenfold increase in productivity. The integration of process analytical technologies and digital twin within a Biopharma 4.0 framework facilitated continuous and automated production, ensuring a scalable and resilient supply of mRNA therapeutics. Similarly, Silva et al. leveraged digital twins to optimize the chromatographic process used for the purification of monoclonal antibodies (mAbs)~\cite{silva2024digital}. The digital twin framework facilitated the screening of Cation-Exchange (CEX) resins, model calibration, and the prediction of chromatographic behavior under different conditions, thereby streamlining process development, reducing experimental workload, and accelerating timelines.

\header{Cell Manufacturing and Metabolic Pathway} 
Digital twins at the cellular level create precise, real-time virtual models of biological processes, integrating data from various sources to simulate and optimize conditions for cell growth, production, and metabolic activity~\cite{park2021bioprocess, kolokotroni2024multidisciplinary,helgers2022towardshiv}. These digital replicas enable researchers and manufacturers to monitor, predict, and adjust parameters in real-time, significantly enhancing efficiency and ensuring consistent product quality~\cite{lim2020state,portela2021silico}. In cell manufacturing, digital twins optimize bioreactor conditions by fine-tuning parameters such as temperature, pH, and oxygen levels, managing nutrient supply to maintain optimal growth conditions, and predicting product quality through continuous monitoring and feedback loops~\cite{gargalo2021towards,fontanili2020engineering, an2021application}. For metabolic pathways, they simulate intricate biochemical processes, including enzyme activities, metabolic fluxes, and the effects of genetic modifications, thereby improving the design and productivity of cell-based processes~\cite{mosquera2024digital,joshi2023digital,silfvergren2022digital}. For instance, Cheng et al. incorporated a digital twin model with Biological Systems-of-Systems (Bio-SoS) to simulate interactions in sub-models like single-cell and metabolic shift models. This approach enables real-time adjustments of cultivation conditions and optimal nutrient management, ensuring consistent and high-quality cell culture outcomes~\cite{cheng2024digital}. By simulating the interactions at various scales, from molecular to macroscopic levels, this digital twin model provides a comprehensive understanding of the cell culture process, facilitating more efficient and flexible manufacturing. Likewise, Helgers et al. employed a digital twin to enhance CHO cell-based antibody production using a dynamic metabolic model that simulates central metabolic pathways, such as glycolysis, the TCA cycle, and amino acid metabolisms~\cite{helgers2022towards}. This model incorporates detailed reaction kinetics and feedback mechanisms, allowing for precise control of metabolic activities and cell growth. The digital twin demonstrated high accuracy and precision, enabling the optimization of yield and product quality in monoclonal antibody manufacturing. By integrating real-time data and predictive modeling, this digital twin supports continuous process improvement and robust process control, ensuring high productivity and compliance with regulatory standards.

\header{Single Anatomical Structures} 
Single anatomical structures are individual components of the body with specific forms and functions, typically recognizable as distinct entities, such as organs, bones, and muscles~\cite{borner2021anatomical,chaney1992defining}. Digital twins for these structures provide detailed, accurate models that enhance research and education~\cite{ahmadian2022toward,subramanian2020digital,lal2020development}. They enable pre-surgical planning by allowing surgeons to simulate complex procedures, thereby reducing risks and improving outcomes~\cite{bjelland2022toward,ahmad2021unified}. Additionally, digital twins facilitate the simulation and testing of medical devices, ensuring safety and efficacy before application~\cite{gliszczynski2021digital,kalozoumis2022towards,rowan2024digital}. These virtual models are created using high-resolution imaging techniques like MRI, CT scans, and ultrasound, integrating anatomical and physiological data to enable advanced simulations of real-life scenarios through deep learning techniques~\cite{wang2022deep,sultanpure2024internet,shi2022synergistic,frossard2019vivo}. For example, Shu et al. introduced Twin-S, a digital twin model designed for skull base surgery, which simulates, monitors, and continuously updates all essential aspects of the procedure in real-time to replicate real-world conditions~\cite{shu2023twin}. The model utilizes high-precision optical tracking and real-time simulation to create detailed virtual models of surgical tools, patient anatomy, and surgical cameras. By integrating data from sources like CT scans and employing calibration routines, the system ensures accurate representation and updates the virtual model at a frame rate of 28 FPS. Evaluation of Twin-S shows an average drilling simulation error of 1.39mm, demonstrating its accuracy and potential to improve surgical planning and outcomes. 

One particularly notable advancement in this field is the development of digital twin models for the brain~\cite{amunts2022coming, stephan2010ten, kriegeskorte2018cognitive}. The Digital Twin Brain (DTB) technology integrates multimodal neuroimaging data, genomic data, behavioral data, and cognitive assessments to construct personalized brain models that accurately simulate anatomical structures, functional connectivity, and dynamic changes~\cite{deco2018whole, wang2024virtual, amunts2013bigbrain,xiong2023digital}. This virtual replica holds immense potential for neuroscience research, clinical practice, and brain-computer interfaces~\cite{xiong2023digital, sanz2013virtual}. In 2024, Park, Wang, Guan, and colleagues unveiled an integrated platform for multiscale molecular imaging and phenotyping of the human brain, creating a 3D atlas at subcellular resolution~\cite{park2024integrated}. This innovative approach simultaneously maps brain-wide structures and captures high-dimensional features. The platform includes "Megatome," a device that finely slices intact human brain hemispheres without causing damage, and "mELAST," which makes each brain slice clear, flexible, durable, expandable, and quickly, evenly, and repeatedly labelable. Additionally, the "UNSLICE" computational system seamlessly reunifies the slabs, reconstructing each hemisphere in full 3D with precise alignment of individual blood vessels and neural axons. This platform is expected to enable comprehensive analysis of numerous human and animal brains, enhancing our understanding of interspecies similarities, population differences, and disease-specific characteristics. It also facilitates the mapping of single-neuron projectomes integrated with molecular expression profiles, uncovering the organizational principles of neural circuitry and their alterations in diseases, thus advancing our understanding of disease mechanisms.

\header{Multi-Scale Biological Systems}
Digital twins at the multi-scale level aim to capture the interplay between biological processes that unfold across molecular, cellular, tissue, and organ systems~\cite{tang2024roadmap,viceconti2016virtual}. Unlike single-scale twins, which focus on local dynamics, multi-scale digital twins need to reconcile differences in spatial resolution, temporal dynamics, and data modalities across layers~\cite{pappalardo2019silico}. This creates fundamental modeling challenges: aligning scales with inconsistent data densities, integrating mechanistic models with statistical surrogates, and updating the system in real time while maintaining internal consistency~\cite{barros2024multiscale}. Additionally, many biological interactions are bidirectional: molecular perturbations may lead to organ-level changes, but systemic states such as inflammation or hormonal feedback can also rewire cellular behaviors~\cite{ladroue2009beyond, rabin1989bidirectional}. To address this complexity, multi-scale twins often rely on modular architectures, where distinct sub-models (e.g., gene regulatory networks, metabolic flux simulators, biomechanical solvers) are coupled through interface layers~\cite{agmon2022vivarium, neal2014reappraisal}. These models support high-fidelity simulation of emergent physiology and enable hypothesis testing for diseases whose behavior cannot be understood at a single biological level, such as cancer progression, organ failure, or complex neurological disorders~\cite{hunter2013vision, wang2024virtual}.

Multi‑scale digital twins put these frameworks into practice by enabling integrative modeling of complex disease systems that defy single-level explanation. For example, in oncology, TumorTwin constructs patient‑specific breast cancer twins by linking genomic profiles, histopathology features, tumor microenvironment data, and longitudinal imaging, allowing simulation of tumor evolution and therapy response across molecular, cellular, and anatomical scales~\cite{kapteyn2025tumortwin}. In organ failure, Gallo et al. developed a liver twin that integrates hepatocyte‑level metabolism, zonated lobular blood flow, and whole‑organ perfusion, enabling prediction of spatially localized drug-induced liver injury and systemic decompensation~\cite{gallo2024developing}. For neurological disorders, virtual brain twins for stimulation in epilepsy established personalized brain digital twins by combining patient‑specific structural MRI/DTI, network‑scale neural population models, and EEG/SEEG data~\cite{wang2025virtual}. This multi‑scale architecture captures how stimulation perturbs local excitability and triggers seizure propagation throughout the brain, supporting individualized estimation of epileptogenic zones and in silico testing of neurostimulation strategies. Across these domains, multi‑scale digital twins serve as unified platforms that integrate data and mechanisms from molecular or cellular levels to organ and system scales, enabling mechanistic insight and actionable simulations.

\subsection{Aerospace}
The aerospace sector integrates multiple disciplines of engineering, data analytics, and systems control to support the development and operation of aircraft and spacecraft~\cite{glaessgen2012digital}. Because of its complexity and strict safety requirements, the industry has rapidly adopted digital twin technology as a foundation for modernization~\cite{rathore2021role}. A digital twin creates a virtual model that mirrors the physical system, enabling continuous monitoring, simulation, and predictive analysis~\cite{xiong2022digital}. When integrated with artificial intelligence and physics-based modeling, these systems allow adaptive design, predictive maintenance, and autonomous decision-making, marking a major transformation in aerospace engineering~\cite{grieves2022foundations}. 

\header{Aircraft Design}
At the design stage, digital twins support virtual prototyping and early verification of aerostructures~\cite{glaessgen2012digital}. By linking computer-aided engineering models with computational fluid dynamics and finite element analysis, engineers can explore aerodynamic performance and structural integrity before physical testing~\cite{bazilevs2015isogeometric}. Li et al.~\cite{li2021digital} introduced a digital twin framework for composite part fabrication that synchronizes process data with simulation feedback to minimize deviation. In Europe, the Clean Sky 2 program used digital twins to model airframe deformation under load and refine lightweight materials, achieving measurable improvements in fuel efficiency~\cite{xiong2022digital}. These developments demonstrate how digital twins shorten design cycles and increase accuracy in aircraft development.

\header{Maintenance and Health Monitoring}
Maintenance has become one of the most mature and widely adopted applications of digital twins in aerospace~\cite{tuegel2012airframe}. By integrating real-time sensor data with high-fidelity physical models, engineers can estimate the remaining useful life of key components and identify early signs of fatigue~\cite{liu2021predictive}. The U.S. Air Force developed the Airframe Digital Twin project to simulate cumulative damage and schedule inspections more effectively~\cite{tuegel2011reengineering}. Siemens has also applied digital twin systems to engine monitoring, combining vibration and temperature data to predict performance degradation and prevent unplanned failures~\cite{turner2021digital}. With advances in AI-based diagnostics, these predictive models are helping the aerospace industry transition from reactive maintenance to proactive asset management.

\header{Flight Simulation and Mission Planning}
Digital twins are also reshaping flight simulation by creating dynamic digital environments for training and mission testing~\cite{rossmann2016simulation}. Through integrated aerodynamic and structural modeling, engineers can evaluate flight stability and control under varying atmospheric and operational conditions~\cite{bazilevs2015isogeometric}. NASA incorporated digital twin environments into its simulation platforms to test autonomous flight algorithms and fault recovery procedures before live deployment~\cite{nasa2021digitaltwin}. These virtual testbeds have reduced the reliance on costly physical experiments and improved pilot decision support, particularly in missions that require real-time adaptability.

\header{Space Missions and Satellite Systems}
In space operations, digital twins assist mission controllers in managing spacecraft health and predicting system behavior in orbit~\cite{grinshpun2016virtual}. NASA’s Artemis program applies digital twins to simulate propulsion, communication, and thermal subsystems, allowing mission planners to evaluate performance under extreme conditions~\cite{nasa2021digitaltwin}. For satellite constellations, twin-based simulators reproduce orbital dynamics and sensor interactions, helping teams predict fuel usage and detect anomalies before they escalate~\cite{xiong2022digital}. The European Space Agency has used similar architectures for ground control validation, improving reliability and reducing mission delays. These applications demonstrate the growing importance of digital twins in maintaining the safety and longevity of space assets.

\header{Autonomous Space Robotics}
Digital twin technology is equally transformative for robotic systems used in orbit and planetary exploration~\cite{schluse2016simulation}. By constructing high-fidelity models of robotic manipulators and rovers, engineers can test control strategies and failure responses under microgravity and harsh terrain conditions~\cite{grinshpun2016virtual}. NASA employs digital twin frameworks to monitor robotic arms on the International Space Station, predicting mechanical wear and optimizing motion planning~\cite{nasa2021digitaltwin}. In research by Grinshpun and Rossmann, twin-based virtual testing enabled autonomous robots to rehearse debris removal and satellite servicing tasks, ensuring operational safety before deployment~\cite{rossmann2016simulation}. These developments highlight how digital twins are enabling more resilient, self-learning robotic systems that can operate independently in remote environments.

In summary, digital twin technology is redefining aerospace engineering by fusing data, physical modeling, and intelligent analytics into continuous feedback systems. From early aircraft design to autonomous space missions, these technologies improve precision, reliability, and efficiency throughout the aerospace lifecycle.

\subsection{Smart City}
The rapid pace of urbanization, coupled with the growing demand for livelihoods among urban dwellers and advancements in technology, has significantly accelerated the development of smart cities~\cite{albino2015}. Smart cities integrate a range of innovative technologies, including digital twins, the Internet of Things (IoT), blockchain, and AI, to deliver smart services and enhance the quality of life for local residents~\cite{Zanella6740844, bibri2024, farsi2020digital,joshi2016}. 
Among these technologies, digital twins offer virtual replicas of real urban environments, providing a robust platform for enhancing smart city capabilities~\cite{farsi2020digital}. Additionally, AI further strengthens the data-driven foundations of smart cities by improving data accuracy, predictive analytics, and decision-making~\cite{andrews2022, koumetio2023, marasinghe2024, sanchez2023, son2023, sanchez2022}.
This segment explores the applications of digital twin and AI technologies within the context of smart cities.
% \begin{figure}[htbp] \centering \includegraphics[width=0.8\linewidth]{figDTAI/urban planning.pdf} \caption{Digital twins in urban planning and management.} \label{fig:app-urban} \end{figure}

% more sensor data

\header{Building Efficiency} 
Buildings account for the majority of energy consumption in cities~\cite{IPCC2007}, making them a critical area of attention for the development of smart city initiatives~\cite{El-Agamy2024}.
In the context of urban building energy systems, digital twins significantly enhance the capabilities of smart metering infrastructure. This infrastructure facilitates the recording of electricity usage at granular levels, with meter readings taken at intervals of less than one hour~\cite{EIA2018}. 
Beyond electricity metering, building-level sensing infrastructures increasingly integrate Internet of Things sensors to capture environmental and operational variables such as occupancy presence and indoor temperature, which are critical drivers of building energy demand~\cite{gray2021internet}. Recent studies highlight that such IoT-enabled sensing infrastructures constitute the primary source of large-scale urban and building data used in data-driven energy management systems~\cite{HAJJAJI2021100318}. These heterogeneous sensor streams form the data foundation for data-driven building energy modeling and control.
When integrated with machine learning techniques, this rich energy data can support a wide range of applications, including energy load analysis, forecasting, management, and real-time assessments of energy consumption~\cite{8322199}. For instance, decision tree and support vector machine (SVM) classifiers can be employed to detect anomalous consumption patterns in real time~\cite{7434588}. Additionally, a k-nearest neighbors classifier can be utilized to analyze the energy behaviors of occupants in commercial buildings~\cite{RAFSANJANI2018317}.
For building energy prediction, Fan et al. leveraged deep learning techniques for data augmentation, significantly improving the accuracy of short-term energy forecasts for buildings~\cite{fan2022}.

\header{Microgrids}
Microgrids and smart grids are among the most important and well-established application domains of digital twin technology. Unlike building-level energy systems, microgrids operate at the intersection of local energy generation, distributed loads, and the public power grid, extending beyond the scope of building energy efficiency alone. A typical microgrid integrates distributed energy resources, interconnected loads, energy storage systems, and control mechanisms, enabling flexible operation under both grid-connected and islanded modes.
Within the digital twin framework, microgrids have been extensively investigated for their potential to enhance system-level intelligence, robustness, and operational resilience. Existing studies have focused on forecasting tasks, including renewable generation and load prediction~\cite{din2017,he2017}, as well as system management and real-time monitoring~\cite{xu2019deep,park2020}. Digital twin models have also been applied to fault detection and predictive maintenance, facilitating the early identification of abnormal behaviors and component failures~\cite{nowocin2017,goia2022}. In addition, cybersecurity and system protection have emerged as critical research topics, particularly given the increasingly tight cyber–physical coupling in microgrids~\cite{huang2021iop}.
A prominent research direction involves using digital twins to proactively enhance grid robustness against multifaceted uncertainties. Advanced frameworks employ hybrid stochastic-robust optimization methods to determine optimal schedules for both normal and resilient operation, effectively modeling uncertainties related to grid costs, renewable generation, and loads~\cite{CAO2024103628}. Beyond standalone resilience, digital twins are crucial for deepening the operational and supportive integration of microgrids with the main public grid. This involves not only islanding during outages but also providing active support to the main grid under normal conditions. For example, digital twin-enabled coordinated control strategies can orchestrate distributed energy resources such as wind and solar to provide rapid frequency support to the main grid, a critical service as renewable penetration increases~\cite{chang2024frequency}.
At the local distribution level, digital twins play an indispensable role in coordinating complex, modern energy consumption, particularly within building clusters and electric vehicle charging stations. The spatiotemporal flexibility of electric vehicles, including their ability to move and shift charging times, presents both a challenge and a unique resource for local energy management. Recent research proposes digital twin-based strategies for coordinating electric vehicle charging and discharging across multiple temporary microgrids formed after a blackout~\cite{Zaiets_Koskina_Drozhzhyn_2024}.
Beyond these aspects, Sun et al. emphasize the role of digital twins in improving microgrid robustness and resilience against various sources of uncertainty, including renewable intermittency, load fluctuations, and grid disturbances~\cite{resilience2021}. Lasseter et al. highlight the growing research attention toward the integration of microgrids with the public grid, especially when considering localized energy consumption scenarios such as building clusters and electric vehicle charging stations~\cite{5768104}. Furthermore, Palensky et al. demonstrate that digital twin-enabled co-simulation and control strategies provide a promising approach for coordinating energy flows across multiple subsystems and operational scales~\cite{7883974}.

\header{Urban Planning}
Infrastructure—including buildings, bridges, roads, railways, and associated pipeline networks—plays a critical role in the development of smart cities. However, planning, constructing, and maintaining such infrastructure presents significant challenges due to high costs, extensive resource requirements, long project timelines, and the unique characteristics of each project~\cite{YuZheng2022}. Digital twin technology offers a promising approach to address these challenges.
A fundamental aspect of digital twin technology is the representation of infrastructure. Digital twins of cities can be organized into multiple layers~\cite{Castelli2019}, one of which is the geometric layer~\cite{scalas2022}. This layer captures the morphology and physical characteristics of infrastructure in 3D~\cite{catalano2011,scalas2022}. For example, 3D digital twin modeling software like CityEngine can be used to manually reconstruct urban infrastructure, as demonstrated in Matera, Italy~\cite{scalas2022}. The accuracy of these models can be further enhanced using AI and data augmentation techniques~\cite{Baduge2022}, while deep learning methods enable the generation of complex, high-dimensional digital representations of urban infrastructure~\cite{Abdollahi2020}.
Beyond modeling, digital twin technology supports infrastructure design through simulation and analytical testing. By relying on virtual models, designers can perform 3D design iterations with greater precision and receive more accurate feedback compared to traditional methods~\cite{WANG202033}. This virtual approach extends to fabrication and installation, where digital twins combined with virtual reality allow for systematic data storage, precise positional coding, and accurate measurement of components~\cite{ZhouSE2019}.
Integrating data-driven digital twin models with intelligent systems further enables real-time predictive maintenance, diagnostics, and informed decision-making~\cite{LUO2020101974}. Machine learning techniques, for instance, can support Structural Health Monitoring (SHM) of civil infrastructure~\cite{LiBetti2023}. These methods allow for independent generation of new data samples, enhancing infrastructure safety by predicating potential failures.

\header{Public Safety and Environment}
Digital twin technology plays an increasingly important role in enhancing public safety and environmental governance by offering a dynamic, data-driven framework for simulating, monitoring, and managing complex urban systems. In this context, its primary applications encompass emergency management, environmental risk assessment, and the protection of critical infrastructure.
In the field of emergency management, digital twins serve as effective tools for disaster preparedness, real-time response, and post-event recovery. By integrating real-time data streams from IoT sensors with geospatial information systems and historical incident records, digital twins generate a continuously updated virtual representation of a city’s physical state. This living model enables authorities to visualize the evolution of natural hazards, such as floods, earthquakes, and wildfires, in real time, assess their potential impacts on populations and infrastructure, and evaluate alternative evacuation plans or resource allocation strategies through scenario-based simulations~\cite{dougan2021digital, Deng2020, WHITE2021103064}. During flood events, for instance, digital twins can model hydrodynamic processes, predict inundation extents, and identify vulnerable critical assets, thereby supporting early warning dissemination and targeted emergency interventions~\cite{ge2025urban}.
Beyond reactive disaster response, digital twins also support proactive environmental monitoring and public health protection. By assimilating data from air quality sensors, traffic flows, and meteorological observations, digital twins can model and forecast urban air pollution dynamics, enabling the identification of pollution hotspots and the evaluation of mitigation measures such as traffic restrictions or emission control policies~\cite{farsi2020digital}. In a similar manner, digital twins can be applied to monitor the spatial and temporal distribution of noise pollution and assess its effects on urban livability and community well-being.
Digital twin technology further contributes to public safety through the continuous monitoring and protection of critical infrastructure. As demonstrated by Khan et al., machine learning techniques embedded within digital twin frameworks enable the early detection of structural anomalies or performance degradation in assets such as bridges, tunnels, and power grids~\cite{Khan2019}. By continuously comparing real-time sensor data from physical structures with the expected behavior predicted by their digital counterparts, potential failures can be anticipated before they occur. This enables predictive maintenance strategies and reduces the likelihood of catastrophic incidents. Such capabilities are particularly valuable for advanced composite materials used in aerospace and transportation systems, where digital twins facilitate structural health monitoring (SHM) to ensure long-term integrity and operational safety~\cite{Khan2019, LiBetti2023}.

\subsection{Mobility and Transportation}
The growing trend towards connectivity and automation in the transportation sector is increasingly leveraging advanced technologies to optimize traffic management and improve overall system efficiency~\cite{irfan2024towards}. A particularly promising technology is the digital twin, which can be employed to create dynamic and real-time models of various elements within the transportation ecosystem. This includes the ability to monitor and simulate traffic flow, evaluate vehicle performance, and assess road conditions, among other critical factors that collectively impact the effectiveness and safety of transportation systems~\cite{dureja2024combining}. By providing a comprehensive and continuously updated virtual representation of the physical world, digital twin technology stands as an essential tool at the forefront of mobility and transportation innovation.
Given the rapid advancements in Connected and Automated Vehicles (CAV) and the Internet of Vehicles (IoV) technologies, the integration of digital twins into traffic management platforms is becoming increasingly feasible and advantageous. This integration promises to further revolutionize mobility and transportation systems by enabling safer and more efficient traffic management solutions~\cite{10443037}.

\header{Connected and Automated Vehicles}
%Autonomous driving is defined by vehicles' capability to perceive their surroundings, plan routes, and navigate safely through advanced technologies. Although these vehicles are still in the testing phase and have not yet been widely adopted, they are expected to dominate the global automotive market soon due to their many benefits. Autonomous driving incorporates digital twin and AI technologies to enhance the digitization and collaboration of interconnected autonomous vehicles~\cite{ali2023review}. 
The rise of CAV technology introduces another platform to implement digital twins beyond traditional autonomy. Since the level of automation and connectivity within our vehicles has greatly improved, these equipped vehicles can not only sense their surroundings using their onboard perception sensors (e.g., camera, LiDAR, radar) but also ``talk'' with other agents such as vehicles, infrastructure through vehicle-to-everything (V2X) communications~\cite{wang2020asurvey, xu2022v2x, xu2024v2xv2, gao2025langcoop, gao2025stamp}.
Various studies leverage cloud computing to empower digital twins to serve for CAVs. For example, Wang et al. proposed a mobility digital twin framework that empowers CAVs with various micro-services based on a device-edge-cloud architecture~\cite{wang2022mobility}. Alam and Saddik designed a digital twin framework reference model for the cloud-based CPS, where a telematics-based driving assistance application was proposed for the vehicular CPS with three parts: 1) computation, 2) control, and 3) sensors and services fusion~\cite{alam2017c2ps}. Kumar et al. developed a digital twin-centric approach with machine learning, edge computing, 5G communication, and data lake, aiming for driver intention prediction and traffic congestion avoidance~\cite{kumar2018novel}. Wang et al. proposed a vehicle-to-cloud paradigm for an advanced driver-assistance system (ADAS) of CAVs~\cite{wang2020adigital}, and this paradigm was further experimented with by Liao et al. in a cooperative ramp merging scenario~\cite{liao2021cooperative}. The challenge of properly visualizing the digital twin information received from the cloud was studied by Liu et al., where a sensor fusion method that combines onboard camera data was proposed to facilitate the decision-making of CAVs~\cite{liu2020sensor}. Gao et al. proposed a multi-tiered Carla-SUMO-AirSim co-simulation approach to bridge ground and air V2X collaboration, enabling broader embodiment for CAV research~\cite{gao2025airv2x}.
Testing and evaluation are essential for the progress and implementation of CAVs. Digital twin technology is particularly useful for facilitating closed-facility testing by integrating virtual components generated by computer systems with real-world road conditions~\cite{9540143}. Fully testing automated driving systems poses significant challenges and requires extensive testing that cannot be accomplished without simulation support. As a result, recent research has concentrated on creating simulation frameworks for testing~\cite{wang2021digitaltwinsimulation, chen2024opencda}. The digital twin framework proposed by Ge et al. identifies three testing levels: entirely virtual, based on actual sensor data, and vehicle-based~\cite{ge2019research}. A similar approach involves using digital twin technology within specific frameworks to record vehicle responses in various simulated environments, which helps generate a large dataset for training and testing autonomous vehicle control systems, thus establishing a solid foundation for accurate system development~\cite{rassolkin2019digital}. Digital twins are also used to create statistical models that forecast a vehicle's future movements based on historical data~\cite{yang2017digital}. Additionally, researchers have applied digital twin technology to dynamically compute motion parameters, improving vehicle trajectory planning~\cite{zhang2019time}. The modeling of human behaviors on CAVs has also been advanced with digital twin technology, where the Driver Digital Twin (DDT) concept was first proposed by Chen et al. to simulate human driver behavior on CAVs~\cite{chen2018digital}. This concept has been further researched by various studies, which involve digitizing human drivers to link current autonomous driving systems with fully digital systems and predicting drivers' future decision-making process with machine learning approaches~\cite{hu2022review,liao2023driver, ma2024driver}. This development contributes to the creation of a comprehensive Human-Cyber-Physical System (H-CPS) that integrates human driving behaviors~\cite{9847095}.

\header{Internet of Vehicles}
IoV is a sophisticated network of vehicles outfitted with sensors, software, and technology, all of which follow standardized protocols to connect and share data over the Internet~\cite{lee2016internet}. This network includes not only the connections between vehicles~\cite{wang2020v2vnet} but also the links between vehicles and various road infrastructures and the cloud~\cite{xu2022v2x,xu2022cobevt,li2024comamba,wang2025cocmt,wang2025cmp}. By exchanging traffic information, IoV systems can collaboratively optimize vehicle movements and traffic control, leading to a significant improvement in overall traffic efficiency~\cite{ji2020survey}. As a result, IoV is seen as a crucial factor in the future of autonomous driving, as well as connected, electrified, and shared mobility~\cite{lauridsen2017lte}.
Digital twin technology is widely utilized in IoV for resource allocation, sharing, and traffic forecasting. For example, a method for real-time traffic data prediction using digital twins has been developed, which relies on monitoring traffic flow and speed data transmitted via 5G through IoV sensors, greatly enhancing the system's accuracy and response time~\cite{9440709}. Experiments were carried out using traffic data collected in Nanjing, China~\cite{9939166}. Additionally, W. Sun et al. have explored a digital twin model aimed at dynamic resource allocation in aerial-assisted IoV networks, allowing for coordinated resource scheduling and distribution~\cite{9351542}.
In terms of offloading strategies within IoV, a study has proposed a digital twin network framework that maintains digital twins in cyberspace, facilitating the synchronization of real-world vehicle activities~\cite{xu2023v2v4real}. An innovative IoV framework has also been introduced, utilizing digital twins to create a digital representation of the IoV environment~\cite{electronics13071263}. This enables real-time updates to a vehicle's driving route based on current data, significantly improving navigation and operational efficiency.
Moreover, a digital twin design that incorporates consortium blockchain technology has been developed, focusing on remote resource sharing within the IoV framework to effectively track and safeguard resources~\cite{9625367}. In aerial IoV networks, the RADiT framework has been introduced as a digital twin–driven resource allocation model that leverages real-time network state representations to enable fast and efficient resource sharing, thereby significantly enhancing network throughput, latency performance, and overall connectivity~\cite{10234627}.
As a prime example of a fully connected scenario, IoV enables intelligent vehicle operations through the use of artificial intelligence (AI). The future of vehicular networks will require a wide range of services that demand considerable computing resources. To tackle the resource shortage, neural networks are utilized to optimize the use of excess computing capacity.
Federated Learning (FL) has also been applied extensively in the IoV sector~\cite{chellapandi2023federated}. For instance, a novel asynchronous FL method has been created to ensure secure and efficient data sharing within IoV networks~\cite{11015654}. Additionally, I. Ullah et al. introduced a blockchain-supported FL algorithm aimed at enhancing knowledge sharing in IoV networks, promoting collaborative learning and data integration~\cite{Ullah2024}. To address the challenge of efficiently recognizing license plates in 5G-enabled IoV environments, a new FL model has been designed to improve recognition accuracy and processing speed~\cite{9381655}. Furthermore, FL-based collaborative positioning technology has been demonstrated in IoV networks to support autonomous driving and collision avoidance, thereby enhancing the safety and efficiency of vehicular movements~\cite{2022ITCSS...9..197K}.

\header{Traffic Flow}
The application of digital twin technology in traffic flow presents considerable opportunities for enhancing mobility and transportation systems. Digital twins can effectively visualize traffic patterns within urban environments, providing a detailed and dynamic representation of traffic flow~\cite{zhang2017spatio,wang2025uniocc}. VISSIM, a microscopic traffic flow simulation software, can be integrated into a digital twin framework for smart traffic corridors by utilizing real-world data, allowing for highly accurate simulations and analyses~\cite{Abhilasha2021}. 
A digital twin focused on mobility management has been created, leveraging a cloud-based microservices architecture to manage and optimize traffic flow in a scalable and flexible manner~\cite{XuHaowen2023}. Additionally, a radar-camera fusion approach has been proposed to develop a digital twin for specific sections of roadway, enhancing the precision and reliability of traffic monitoring and management~\cite{LiYanbing2023}. 
A traffic flow prediction model based on digital twin architecture and neural networks has also been established, enabling the accurate prediction of inflow and outflow at nodes within the Beijing Subway network~\cite{toque2016destination,Wu2020LSTM,TuZhen2022}. Another line of approaches for predicting network traffic and behavior of traffic participants involves the use of reinforcement learning or generative models, which have demonstrated effectiveness in providing long-term forecasts~\cite{NieLaisen2022,wang2025deployable,LangeB-RSS-25,li2024adaptive,li2024interactive,toyungyernsub2024predicting,li2020evolvegraph,zhao2025trajevo,girase2021loki,li2019coordination,lange2024scene,xie2023cognition,toyungyernsub2022dynamics,ma2021multi}. 
Once traffic flow is predicted and visualized, the functionality of digital twins can be further enhanced by integrating artificial intelligence to classify traffic congestion. For instance, neural network models can leverage extensive datasets to improve the precision of traffic congestion classification, helping to identify and respond to congestion issues in real time~\cite{Jilani2022}.
Moreover, digital twins can provide actionable guidance to drivers based on predicted traffic conditions. For example, advisory messages generated by a digital twin at an intersection can instruct drivers to adjust their speeds, promoting smoother vehicle movement without the need for frequent stops. This approach significantly reduces travel time and energy consumption, thereby improving overall mobility and reducing environmental impact~\cite{WangZiran2022}.
Furthermore, digital twins can play a critical role in optimizing traffic signal timings. By analyzing predicted traffic flow data, digital twins can dynamically adjust signal timings in a smart connected corridor testbed. Such optimizations have been shown to achieve a 20.81\% reduction in travel times compared to traditional actuated traffic control systems, underscoring the potential of digital twins to enhance traffic efficiency and reduce congestion~\cite{Saroj2022}.

\subsection{Smart Manufacturing}
We are currently in the midst of the Fourth Industrial Revolution, also known as Industry 4.0. Industry 4.0 can be defined as the integration of intelligent digital technologies into manufacturing and industrial processes~\cite{sapindustry40}. With the maturity and application of new-generation information technologies, the advancement of Industry 4.0 is in full swing~\cite{mourtzis2024}.
A key component of Industry 4.0 is smart manufacturing~\cite{8119409}, which represents the primary application of ``manufacturing intelligence" across the production and supply chain~\cite{Davis2012145}. Smart manufacturing refers to a new manufacturing paradigm where manufacturing machines are fully connected through wireless networks, monitored by sensors, and controlled by advanced computational intelligence to improve product quality, system productivity, and sustainability while reducing costs~\cite{WANG2018144}. This area has garnered widespread attention and experimentation~\cite{Frank201915, Cohen20194037, carla2020}. The implementation of smart manufacturing is driven by digital twins, which abstract physical entities in factories into their digital forms within cyberspace~\cite{Ghosh2019317, Errandonea2020, 8247583, Josifovska2019398, Shakhnov2020864}. This abstraction enables the monitoring, control, diagnosis, and prediction of the states of these entities~\cite{SCHLEICH2017141, Aheleroff2020228, Guerra-Zubiaga2021933, 9024430}.
In addition to digital twin technology, smart manufacturing leverages a set of technologies that include industrial IoT networks, AI, Big Data, robotics, and automation to enhance system efficiency and improve outcomes~\cite{LI2022100289, JAVAID202371, lee2020integration, Jiang2021, Howard20211, Melesse2020267, singh2021appl}. Among these, AI is a critical technology for digital twins in smart manufacturing. AI enables data processing and real-time prediction of manufacturing processes and component performance, thereby optimizing the performance of digital twins~\cite{TANG2023100753} and improving product design and manufacturing efficiency~\cite{bartsch2021}.
Here are some applications that illustrate the use of digital twins and AI to achieve smart manufacturing in the background of Industry 4.0.

% \begin{figure}[htbp]
%     \centering
%     \includegraphics[width=0.9\linewidth]{figDTAI/manufacture.pdf}
%     \caption{Digital twins in smart manufacture and industry 4.0.}
%     \label{fig:app-manufacture}
% \end{figure}

\header{Manufacture Visualization}
Visualizations, as an important component of manufacturing, can effectively transform vast amounts of information into knowledge and insights, thus facilitating system control for staff~\cite{WANG202212}. In smart manufacturing, digital twins primarily drive graphical visualization, 3D interactive visualization, and augmented reality (AR) among the visualization technologies.
Graphical visualization refers to the real-time mapping of machine status through visual representations, enabling operators to remotely view manufacturing data and better adjust systems. For instance, dashboards can display machine health status, production efficiency, and order scheduling~\cite{Kung-Jeng2021}. Additionally, Tong et al. presented HMIs and applications for visualizing and analyzing machining trajectories, machining statuses, and energy consumption~\cite{Tong20201113}.
3D interactive visualization involves 3D simulations of manufacturing equipment, processes, and products. It provides operators, technicians, and process planners with a more intuitive perspective to quickly identify issues and determine root causes. 3D interactive visualization can create virtual models of manufacturing equipment based on physical and kinematic models. For example, a virtual model of a grinding machine based on 3D, physical, and kinematic models can accurately reflect the equipment's real-time status and processes~\cite{Qi_2020}. Moreover, 3D interactive visualization enables remote access to physical systems, permitting experts to provide remote support and problem-solving without needing to be on-site~\cite{8421851}. Manufacturers can gain a comprehensive view of manufacturing process data and analysis by combining 3D interactive visualization with other methods, such as dashboard monitoring. For instance, Zhao et al. integrated dashboards and 3D interactive visualization for the programmable logic controller (PLC) of a milling machine~\cite{10.1145/3351917.3351979}.
The vision of Industry 4.0 is to construct cyber-physical production systems (CPPS) that seamlessly connect the physical and digital worlds, making manufacturing increasingly intelligent~\cite{EGGER2020106195}. AR applications enable real-time access to the vast data generated by CPPS~\cite{yaozhou2019}, aligning with the needs of Industry 4.0 and smart manufacturing. Compared to 3D interactive visualization, AR can overlay virtual information onto the real world, further enhancing human-machine interaction. Operators can wear AR devices during inspections, using actual production scenes as the background to visualize the operating data of the equipment~\cite{10.1145/3411764.3445330}. It is worth saying that AR devices can be used to display possible defects on the product being inspected~\cite{RUNJI2020101957}. 

\header{Production}
% \header{Production Monitoring}
Production monitoring is a crucial aspect of manufacturing. In smart manufacturing, monitoring includes the condition monitoring of machines as well as the quality monitoring of products~\cite{BOTTJER2023162}.
Manufacturing machines often experience failures as a result of degradation or abnormal operating conditions, leading to increased operational costs, reduced productivity, higher rates of defective parts, and even unexpected downtime. Hence, the implementation of condition monitoring is imperative. This involves monitoring and tracking machine status, detecting early defects, diagnosing the root causes of failures, and integrating this valuable information into manufacturing production and control processes~\cite{Park2016303}. With the help of digital twins, condition monitoring of industrial equipment can be achieved. A digital twin architecture was suggested and put into practice for a pneumatic robotic gripper to identify anomalies like pneumatic cylinder leaks and bearing malfunctions~\cite{Redelinghuys20201383}. Miao et al. demonstrated a digital twin framework using multidimensional time series data for anomaly prediction and equipment state monitoring for computer numerical control(CNC) machines~\cite{Miao2021246}.
Increasingly, deep learning techniques are being extensively researched for condition monitoring, offering higher accuracy and timeliness~\cite{ZHAO2019213}. CNNs integrate feature learning and defect diagnosis into a single model and have been applied in many areas, such as bearings~\cite{Janssens2016331,Lu2017139}, gearboxes~\cite{Wang2017310}, wind generators~\cite{Dongqi}, and rotors~\cite{7790137}. In addition, deep Belief Networks (DBNs) have been investigated for fault diagnosis of aircraft engines~\cite{Tamilselvan}, chemical processes~\cite{Yuhong}, reciprocating compressors~\cite{Tranvan}, rolling element bearings~\cite{Shaohaidong,Ganmeng}, high-speed trains~\cite{Yin2016250}, and wind turbines~\cite{en10010006}.
Quality monitoring involves monitoring the quality of products and then identifying potential product defects to improve overall product quality. Compared to traditional quality monitoring, which is time-consuming, labor-intensive, and unable to detect subtle defects, digital twins offer a more efficient quality monitoring method. Using manufacturing and sensor data, a digital twin-based predictive model for surface roughness quality in machine tools was developed~\cite{Cai20171031}. To tackle optical quality control challenges in additive manufacturing, multiple digital twins were developed to streamline the inspection process~\cite{MORETTI2021101609}.
Deep learning, especially CNNs, has been applied to various textures or hard-to-detect defect cases. CNNs, initially designed for image analysis, are particularly suited for automatic defect identification in surface integration inspection. Max-pooling CNNs perform feature extraction directly from pixel representations of steel defect images~\cite{7864335}, facilitating automatic inspection of dirt, scratches, burrs, and wear on surface parts~\cite{6252468}.

% \header{Production Optimization}
Optimization in industrial manufacturing refers to improving performance and efficiency while meeting specific objectives within manufacturing operations. Digital twin optimizes manufacturing processes by providing accurate, real-time data to enhance performance, reduce waste, and increase sustainability through virtual commissioning and parameter optimization ~\cite{BOTTJER2023162}.
Virtual commissioning involves using the virtual entities created by digital twins to optimize and debug processes, replacing some physical operations. This helps reduce operator fatigue and safety risks. For example, simulation tools can be used to retrofit traditional machine tools~\cite{AYANI2018243}.
In industrial manufacturing, continuous trial and error are required to find parameters that yield the highest quality, efficiency, and benefits, which is known as parameter optimization~\cite{BOTTJER2023162}. However, traditional physical methods of trial and error are often insufficient, costly, and time-consuming. Digital twins offer a significant advantage by simulating virtual scenarios and predicting processes, allowing for low-cost testing of various parameters to find the optimal values. Consequently, many applications of digital twins focus on parameter optimization. For instance, a digital twin system of a cutting machine tool can be developed to optimize machining dynamics and estimate and compensate for contour errors~\cite{Tong20201113}. Furthermore, Balderas et al. applied ant colony optimization for manufacturing hole patterns on printed circuit boards using minimal trajectory and tool change time~\cite{Balderas20211295}. Moreover, dynamic programming can also be used to optimize a grinding process regarding processing time, feeds, and product quality requirements~\cite{pereverzev2020}.
Machine learning captures critical process parameters with greater accuracy~\cite{Arinez2022}. In laser manufacturing, an Artificial Neural Network (ANN) can be used to predict laser cutting quality, represented by explicit nonlinear functions, to optimize parameters related to laser power, cutting speed, and pulse frequency~\cite{Chaudhari2012ArtificialIA}.

\header{Supply Chain Management}
Recent major disruptions, including natural disasters, geopolitical tensions, and the COVID-19 pandemic, have prompted supply chain managers to seek technologies that enhance sustainability and resilience in order to better address these challenges. In this context, the Supply Chain Digital Twin (SCDT) has emerged as a promising concept, showcasing extensive applications across various sectors. It has the capability to replicate the physical supply chain and identify potential issues before they arise~\cite{CIMINO2024100154}.
For instance, the digital twin based on the temperature and quality data can be used to simulate the cooling process in real time. This improves refrigeration processes and reduces food losses, thereby making the refrigerated supply chain greener~\cite{DEFRAEYE2019778}. Moreover, the port of Rotterdam has been working with International Business Machines Corporation (IBM) to create a digital twin that helps the port test scenarios and understand how to improve operational efficiency~\cite{boyles2019port}. SCDT can also be used to help logistics players manage container fleets more efficiently~\cite{dohrmann2019}.
AI techniques, including machine learning, evolutionary algorithms, big data analytics, and reinforcement learning, enhance the capabilities of digital twins in supply chains by leveraging historical data to improve real-time data analysis and predictions, thereby expanding the functionalities of SCDT~\cite{Marmolejo-Saucedo2020,Azevedo2024}. 
Anomaly detection is essential for supply chain managers, as it enables the identification of potential issues, allowing for proactive measures to mitigate negative impacts and maintain operational continuity. Machine learning substantially enhances these detection capabilities~\cite{Landauer2022}. Additionally, artificial neural networks can aid in identifying and mitigating risks associated with supply chain operations, such as supply disruptions, demand fluctuations, and shifts in market conditions~\cite{NEZAMODDINI2020107569}.
Within the SCDT framework, reinforcement learning can optimize real-time decision-making in transportation and logistics, including routing, scheduling, and vehicle dispatch~\cite{SHIUE2018604}. Furthermore, the automation capabilities inherent in SCDT can be improved through reinforcement learning, empowering stakeholders to make well-informed decisions based on specific scenarios~\cite{Abideen2021}.
The quest for optimal inventory policies involves fine-tuning inventory parameters and evaluating inventory costs in relation to service levels. This process can be effectively supported by evolutionary algorithms~\cite{SARACOGLU20148189}.

\subsection{Robotics} 
Robotics fundamentally revolves around transforming ideas into action, which translates abstract goals into tangible physical outcomes and guiding processes toward their objectives~\cite{halperin2017robotics}. The use of digital twins in robotics enables a tight integration between physical systems and their virtual counterparts~\cite{9367549,8049520}, supporting more precise sensing, modeling, and control. In this regard, digital-twin technologies offer exceptional opportunities for advancing robotics development~\cite{Mazumder2023}. Current literature highlights five key domains where digital twins are prominently applied: space robotics, medical and rehabilitation robotics, soft robotics, industrial robotics, and human-robot interaction. Moreover, the integration of artificial intelligence into robotics further empowers systems to become more intelligent, autonomous, and efficient across diverse applications~\cite{andras2020artificial,SOORI202354}.

% Robotics fundamentally revolves around transforming ideas into action. It translates abstract goals into tangible physical outcomes, guiding processes toward their objectives~\cite{halperin2017robotics}. The application of digital twin in robotics enables the effective merging of physical systems with their virtual counterparts~\cite{9367549,8049520}, facilitating precise sensing and control of robotic systems. In this regard, the application of digital twins in this field offers exceptional opportunities for advancing robotics development~\cite{Mazumder2023}. 
% Five key domains have been identified where digital twins are significantly implemented: Space Robotics, Medical and Rehabilitation Robotics, Soft Robotics, Human-Robot Interaction, and Industrial Robotics.
% Furthermore, the integration of artificial intelligence in robotics empowers robots to become more intelligent, autonomous, and efficient in a diverse array of applications~\cite{andras2020artificial,SOORI202354}.

\header{Space and Aerial Robotics}
As outer space presents a harsh environment characterized by extreme temperatures, vacuum conditions, radiation, gravity challenges, and vast distances, human access remains both difficult and hazardous. Consequently, human activities in outer space are significantly limited. In this context, space robotics have become essential for assisting these activities~\cite{5306922}. 
A Virtual Testbed (VTB) specifically designed for optical sensors in space robotics has been developed, which represents a common approach in digital twin applications within space robotics~\cite{sondermann2017virtual, 8289327}. The researchers in this study aimed to integrate digital twins into the VTB, allowing for the simulation of space robotics while simultaneously controlling the robot actuators in a virtual environment. In practical applications, the digital twin concept has demonstrated its significance by notably enhancing mission efficiency~\cite{martin2020osiris}. 
This improvement stems from its ability to facilitate complex decision-making through the evaluation of simulation outcomes conducted on the digital twin within the VTB~\cite{lockheed2020}. Additionally, a novel simulation-based methodology known as Experimentable Digital Twins (EDT) was proposed~\cite{8289327}. The study also introduced a practical approach for integrating these EDT infrastructures into simulation environments, referred to as virtual test bed (VTB). This integration is believed to hold substantial potential for advancing space robotics, particularly regarding the development and testing of simulation and component algorithms.
One study highlighted in~\cite{weber2018space} presented a haptic telerobotic system that utilized digital twins to seamlessly assemble miniature modular satellites in space. Another research focused on methods for processing the telemetry data associated with the DT-integrated telerobotic system described earlier~\cite{10.1115/DETC2019-97151}. 
The construction, repair, refurbishment, and maintenance of aerospace components, such as spacecraft bodies and systems, have emerged as critical aspects of digital twin adoption in contemporary space robotics. For instance, a robotic grinding system incorporated with digital twin technology was developed for aerospace maintenance, repair, and overhaul. This system employs a 6-DoF robotic arm to perform grinding operations, utilizing the digital twin to analyze and determine essential grinding parameters, such as the required grinding force~\cite{oyekan2020}.
Beyond outer space applications, digital twin technologies have also become increasingly essential in the field of aerial robotics and unmanned aerial vehicle (UAV) systems. As UAV swarms grow in scale and operational complexity, digital twins enable high-fidelity airspace modeling, mission rehearsal, and adaptive swarm coordination. Several studies have illustrated these capabilities. For instance, T. Souanef st al. introduced a digital-twin-supported flight safety framework for multi-UAV operations, enabling real-time risk assessment in dynamic airspace~\cite{drones7070484}. Research presented in ~\cite{article} developed a digital twin platform for UAV swarm mission planning, demonstrating improved coordination and reduced collision risks through virtualized testing. In the context of inspection tasks, a digital-twin-enhanced UAV system for infrastructure monitoring was proposed, where the virtual replica was used to evaluate sensor placement strategies and flight trajectories before field operation~\cite{10.1007/978-3-031-69626-8_102}. These examples demonstrate that integrating digital twins into aerial robotics parallels their use in space robotics, providing a unified simulation-driven framework that enhances autonomy, safety, and mission reliability across both domains.

\header{Medical and Rehabilitation Robotics}
Although the concept of medical robotics is not new, its integration with digital twins has gained significant attention in recent years~\cite{Mazumder2023}. 
One of the fastest-growing areas in this field is DT-aided medical telerobotics, which has key applications in robotic surgeries~\cite{8632888,Olivas-Alanis2020,Hagmann2021,Shi2021,LOPOMO2022533}. Medical telerobotic approaches often utilize immersive virtual reality (VR) interfaces or environments~\cite{8632888,9659460,Yiming2022}. These include VR-assisted telerobotic medical and laboratory equipment management~\cite{9659460}, telemedical service robots~\cite{9435999}, and RDT-VR-assisted e-skin and soft actuator developments for telerobotic bio-sample collection in contagious environments~\cite{Yiming2022}. Notably, the emergence of the COVID-19 pandemic prompted the development of disaster management-centric telerobotic approaches aimed at preventing infections in such environments~\cite{9659460,Yiming2022}. 
In recent years, another rapidly growing trend in medical robotics has been rehabilitation, which encompasses the development of DT-aided prosthetics, exoskeletons, and other robot-assisted measures. For instance, DT-aided neuromusculoskeletal modeling and simulation can be constructed using CT scans to aid in the development of biomimetic robot prototypes. Additionally, digital twins can simulate and optimize the creation of patient-specific prosthetics~\cite{Pizzolato2019}, robotic hexapod external fixators focused on correcting bone deformities~\cite{9476181}, and Triboelectric sensor-based exoskeletons~\cite{Zhuzhong2021}. Moreover, a digital twin for a haptic hand exoskeleton can replicate the rehabilitation process in a VR environment~\cite{Topini2022}. An Automatic Gait Data Control System (AGDCS) has been developed for self-activating DT-aided lower limb exoskeletons~\cite{wanghe2023}. 
The COVID-19 pandemic accelerated the adoption of robotic assembly for medical equipment to address the escalating demand for such supplies. The digital twin of a robotic system was created for the contactless distribution of medicines and essential supplies in response to the pandemic~\cite{9133423}. Additionally, a DT-optimized Human-Robot Collaboration (HRC) system was designed to meet the increasing needs for medical equipment, particularly alternators and ventilators~\cite{bilberg2019}. 
The implementation of AI in conjunction with digital twins has also been on a steady rise. For example, R-CNN has been employed to detect various medical equipment and their corresponding digital twins' positions within a virtual environment~\cite{9659460}. That same year, a DT-integrated robot platform was developed utilizing deep learning to automatically collect bio-samples from patients' nasal vestibules~\cite{parak2022intelligent}. 
Lastly, micro-nano medical robotics represents an emerging trend in the field. In 2022, a digital twin of a micro-bot was developed, which employed AI to predict system outputs.

\header{Soft Robotics}
As we know, sensors are employed to monitor both their movements and external stimuli, particularly in fields such as surgical and micro-nano manipulation, where accurate motion detection and tactile sensing are crucial~\cite{Bandari2019,dong2020wearable,gul2019retracted,fujiwara2014flexible,dahroug2018review}. Soft robotics, characterized by high compliance and dexterity with muscle-like actuators made from materials such as silicone rubber and thermoplastic polyurethanes (TPUs)~\cite{WangH2018,rus2015design,yap2016high,majidi2014soft}, is well-suited for collaboration with sensors to enhance monitoring and control in robotic applications. Although soft robotics is a relatively new field, it holds considerable potential. The digital twin concept in soft robotics generates digital information that accelerates development~\cite{Mazumder2023}.
Recent research trends in the digital twin realm of soft robotics are increasingly focused on augmented and extended reality~\cite{Borges2022,zhang2022artificial}. Soft robots can serve as virtual humans within human-centered production systems, where infrastructure efficiency can be further enhanced by leveraging digital twins in virtual reality (VR) environments~\cite{ali2020}.
A digital twin of a soft robot utilizing pneumatic muscles has been developed, offering numerous applications, including cellular production, where it optimizes the work environment and reduces space requirements~\cite{Sokolov2024}. Traditional data processing techniques are insufficient for managing the heterogeneous big data necessary to establish human-machine interfaces (MMI) for the digital twin. This necessitates the use of machine learning approaches in conjunction with advanced communication protocols like 5G to maximize the efficiency of digital twin-soft robotics and related systems~\cite{9436090}. For example, machine learning enhances data interpretation for improved manipulation or detection, such as accurate gesture recognition~\cite{zhu2020haptic}. 
This approach is used to explore the capabilities of specially designed triboelectric nanogenerator (TENG) sensors. A tri-actuator soft gripper, fabricated through three-dimensional (3D) printing and integrated with TENG sensors, perceives gripping status and identifies objects by leveraging machine learning for data analysis. Additionally, a digital twin framework is established to create a duplicate digital representation of the aforementioned manipulation within a virtual reality (VR) environment, often referred to as cyberspace~\cite{jin2020triboelectric}.

\header{Industrial Robotics}
The industrial robotics sector is one of the fastest-growing divisions, offering standardized technologies suitable for various automation processes~\cite{dzedzickis2021advanced}. A comprehensive analysis in~\cite{errandonea2020digital} revealed that the digital twin could be effectively utilized for five types of industrial plant maintenance: reactive, prescriptive, condition-based, predictive, and preventive maintenance, with the latter two being particularly effective. These approaches are also applicable to robot-integrated systems.
Vachálek et al. demonstrated a digital twin that employed a genetic algorithm for DT-optimized predictive analysis and plant augmentation in a robot-assisted production line~\cite{7976223}. Subsequently,~\cite{AIVALIOTIS2019417} proposed a Robot Digital Twin (RDT) modeling methodology utilizing a physics-based approach and virtual sensors to collect and generate data from industrial robots for predictive analysis.
In 2018, a reinforcement learning approach was applied to enable a robot to learn to lift various weights autonomously. The robot model could be visualized and controlled through its digital twin~\cite{verner2017robot}. More recent developments include DT-integrated machine vision assessments of industrial robotic skills~\cite{9766459} and enhanced industrial robot programming using machine learning with point cloud information and RDT~\cite{Enes2022}.
A significant limitation of the data-driven RDT approach, as highlighted by~\cite{AIVALIOTIS2021102177}, is the lack of historical data. However, this challenge can be addressed by generating synthetic data, which can be further improved using AI algorithms. In~\cite{Kosmas2020}, the authors presented a deep learning approach for generating additional synthetic image data from digital models, achieving a 100\% success rate when the developed model was retrained and utilized for detecting real-world objects in various orientations.
Developing efficient AI algorithms for physical robots poses a critical challenge due to the excessive time consumption, power supply, and component constraints for long-term repetitive tasks, as well as the absence of suitable virtual testbeds (VTBs). An effective solution lies in developing algorithms through extensive simulations, where digital twins have proven to play a vital role~\cite{LIU2022102365,HU2022102371}. Notable examples include DT-aided deep reinforcement learning policy transfer from simulation to physical robots~\cite{LIU2022102365}, and training robots within their digital twin for intelligent grasping using a grasp-generation-and-selection convolutional neural network, which achieved 96.7\% and 93.8\% success rates for gripping single items and mixed objects, respectively~\cite{HU2022102371}. 

\header{Field and Service Robotics}
Field and service robotics span diverse application domains—including agriculture, construction, mining, underwater exploration, search-and-rescue, logistics, and public service—where robots must operate in dynamic, unstructured, and often hazardous environments. Such settings introduce substantial uncertainties arising from terrain variability, environmental disturbances, adverse weather, sensor noise, and unpredictable human interactions. Digital twins have emerged as a powerful tool to mitigate these challenges by providing high-fidelity virtual counterparts of physical systems for analysis, simulation, planning, and control~\cite{agriculture15090903,Escriba-Gelonch2024DigitalTwins}.
In agriculture, Melesse et al. reviewed DT-assisted systems that model crop growth, soil moisture evolution, and canopy structures to support autonomous weeding, fruit picking, and targeted spraying~\cite{Melesse2025DigitalTwinCropMonitoring}. L. Yining et al. further introduced DT-aided robotic harvesters equipped with multispectral sensing, in which the digital twin supports fruit-state estimation and enables adaptive manipulation of delicate crops~\cite{LANG2025110451}.
In construction robotics, DT-enabled frameworks integrating building information modeling (BIM), real-time site scanning, and autonomous machinery control have been demonstrated~\cite{Wang_2024building,202508.1387}. DT-based models have been applied to generate optimized excavation trajectories for autonomous earthmoving systems, while DT-supported robotic construction workflows improve placement accuracy through continuous synchronization between virtual and physical models~\cite{articlefu}. Additional efforts include DT-driven autonomous rebar tying~\cite{MOMENI2022103990}, concrete finishing robots developed and validated in DT-based simulation environments~\cite{KajimaCFR2025}, and site-inspection robots exploiting virtual replicas for collision-aware navigation in cluttered and evolving workspaces~\cite{Byers2022LayoutModeling,yao2025towards,li2024multi,li2024interactive,arief2024importance,lee2023robust,ma2021reinforcement}.
In mining and underground environments, DT technologies support higher-risk operations. Lee et al. introduced digital twins of subterranean tunnels that integrate geological models, environmental sensing, and operational records to enhance situational analysis and planning~\cite{app13169137}. Complementary studies show that DT-assisted systems can simulate rock deformation processes or ventilation dynamics prior to execution, enabling improved safety assessment and operational efficiency~\cite{articlejacobs}.
Underwater robotics imposes additional constraints, including limited visibility, dynamic water currents, and severe communication restrictions~\cite{Zaiets_Koskina_Drozhzhyn_2024}. Orjales et al. developed DT-based underwater simulation environments with high-fidelity hydrodynamic modeling, allowing autonomous underwater vehicles (AUVs) to refine navigation and manipulation strategies under realistic virtual conditions~\cite{app15137085}.

\header{Human-Robot Interaction}
The study of Human-Robot Interaction (HRI) has been a focal point of extensive scientific research over the past decades~\cite{goodrich2008human}. The enhanced capabilities provided by digital twins significantly expand the potential for leveraging human-robot systems effectively. Consequently, digital twins in HRI are drawing significant attention for future innovations~\cite{elbasheer2023shaping}. 
Early efforts in this field aimed to develop reliable and interactive robot control interfaces~\cite{10.1145/3359997.3365727} and affordable 3D reconstructions of digital twins for robot control in factories~\cite{8571914}. Digital twin-aided robots in virtual reality (VR) and mixed reality (MR) have emerged as fast-growing research trends in recent years. Applications include HRI workspace design and optimization using VR~\cite{9319116}, robotic construction supervision in MR~\cite{RaviKaushik2021}, and robot programming in VR and MR~\cite{burghardt2020,9196965,garg2021}. Other notable applications of digital twin-aided robots include warehouse and indoor automation using Autonomous Mobile Robots (AMRs)~\cite{Kuts2020,9659133}, and digital twin-integrated, energy-efficient smart manufacturing~\cite{Ali2021}.
Gesture control methods have also been notably applied in various telerobotic applications~\cite{HORVATH2017184,8470634,8577081,CSERTEG201851,9395580}. The Leap Motion method, due to its easy implementation in VR and digital twin integration, has become a popular choice for hand gesture-controlled robotic systems~\cite{8470634,8088140,Cichon2017}. Robots are made aware of collisions with workplace objects and human operators through real-time simulation and processing in a virtual test bed (VTB), where digital twins represent the robots, human operators, and workspace~\cite{HORVATH2017184}. Many approaches utilize the versatility of MR/VR interfaces to further augment digital twin-aided telerobotics~\cite{9659460,GHOSH202010223,9382763,Ordile2021,LOTSARIS2021301}. These augmentations include establishing immersive user interfaces for robot control~\cite{GHOSH202010223,Ordile2021,LOTSARIS2021301}, operations in hazardous environments like nuclear facilities~\cite{GHOSH202010223}, kinesthetic aid to robot operators~\cite{9382763}, and AI-based VR data processing for improved robot control~\cite{9659460}.
Clone digital twins, which collect extensive data from their physical counterparts, represent a trendy application in human-robot systems. They collect near-real-time sensorial data from humans, robots, and their co-environment~\cite{liu2020digital,gualtieri2022development,de2021autonomous}. The scalable data exchange characteristic of clone digital twins allows for incorporating intelligent features using AI and ML for planning, optimizing, and automating human-robot interactions~\cite{liu2020digital}. Furthermore, emerging haptics technology provides an additional data source for HRI~\cite{malik2021digital}. This technology, when combined with Extended Reality (XR), enables real-time sensory experiences for humans, significantly improving joint human-robot task execution.
Recent studies on digital twin-AI applications in human-robot interaction (HRI) have focused on various areas, including machine learning and deep learning techniques for robot trajectory estimation and obstacle detection, which ensure safe and collision-free HRI workspaces~\cite{DRODER2018187}. Additionally, Fast Fourier Transform (FFT) and machine learning-based digital twins are utilized in human-robot interactive welding and welder behavior analysis~\cite{9269521}. Reinforcement learning-based autonomy in complex assembly environments aims to reduce operator fatigue~\cite{ZHANG2022102227}. Another notable development involves lifecycle frameworks and the optimization of pick-and-place robots for virtual product development~\cite{9274698}. Furthermore, DT-aided deep learning is being applied for human action recognition~\cite{gupta2021, PITKAAHO2021540,li2023pedestrian,ma2022multi,li2021rain,dax2023disentangled,zhou2022grouptron}, alongside data augmentation techniques for VR-assisted tele-manipulation~\cite{9659460} and eye-gaze and head gesture recognition systems that facilitate gesture control in robot tele-manipulation~\cite{9395580}. For instance, Mobile ALOHA features a low-cost full-body teleoperation system that collects data from high-quality human demonstrations, alongside a novel imitation learning algorithm that effectively learns from these demonstrations~\cite{zhao2023learning,fu2024mobile,yan2025rdd,ghoshrobust,ghosh2025preference}. Finally, significant advancements have been made in robots capable of self-learning assembly processes~\cite{ahmad2021intelligent}.

\subsection{Natural System and Environment}
%%% a devider, above is re-arranged material %%%
% \newpage
%%% to be deleted after rearrangement %%%

% \begin{figure}[t]
%     \centering
%     \includegraphics[width=0.9\linewidth]{figDTAI/application-nature.pdf}
%     \caption{Digital twins in nature systems and environment.}
%     \label{fig:app-nature}
% \end{figure}

The natural environment is crucial for sustaining life on Earth and maintaining ecological balance, exerting a profound influence on human activity~\cite{sutton2017nature}. Environmental concerns have sparked numerous important topics of discussion, including climate change~\cite{li2023big,national2023opportunities}, biodiversity~\cite{srivastava2005biodiversity, sharef2022applications}, conservation efforts~\cite{qiu2023construction, afsar2024prototype}, and sustainable development~\cite{geisinger1999sustainable,tzachor2023digital}. Digital twin models, benefiting from their unique ability to create virtual replicas of environmental entities and simulate cyber-physical systems, are gradually gaining attention as a promising solution to these environmental challenges~\cite{buonocore2022proposal, chen2023toward}. While the digital twin technology is well-established in industrial sectors, its applications in environmental sciences remain nascent due to the complexity of modeling environmental systems~\cite{dale2023environment}. Traditional physics-based dynamical models from earlier eras are complex and computationally expensive, failing to support high-resolution, interactive digital twin systems for environmental simulations~\cite{singh2023downscalebench}. However, by adopting machine learning and artificial intelligence models, digital twin systems are evolving to offer greater scalability, interoperability, and high-fidelity environmental cyber representations~\cite{jones2020characterising, Mihai2022Digital, sun2022review, cui2023advances, liu2023machine}. This integration not only facilitates rapid interactions among physical environments but also accelerates simulation processes, promoting academic and industrial innovations in fields such as weather~\cite{hoffmann2023destination}, ocean~\cite{bronner2023digital}, geology~\cite{cheverda2019digital2}, wildlife~\cite{sharef2022applications}, and forestry~\cite{jiang2022forestry}. A frontier view of digital twins proposed by researchers, organizations, and companies envisions the creation of a digital Earth, which is portrayed as a dynamic, interactive replica of Earth's past, present, and future states~\cite{le2022earth, national2023opportunities}.

\header{Weather Prediction} Weather prediction aims to reduce associated losses and improve societal benefits by providing accurate information on weather conditions~\cite{national2010weather}. Digital twin models offer precise, high-resolution predictions by combining machine learning and deep learning techniques~\cite{singh2024leveraging, li2023big}, making them pivotal in various applications. Research has demonstrated that incorporating digital twin models in weather prediction can efficiently support decision-making processes in natural disaster risk mitigation~\cite{broccaplaying, zhong2023reduced}, weather services~\cite{hoffmann2023destination, swope2023using}, and renewable energy grids~\cite{sehrawat2023solar, stadtmann2023standalone}.
To predict natural climate disasters, researchers have developed numerous digital models. Brocca envisioned the Hydrology model, a digital twin of Earth's hydrological processes, to predict flooding by utilizing real-time data from both terrestrial and satellite measurements of rainfall and soil moisture, aiding in disaster mitigation~\cite{broccaplaying}. Similarly, a digital twin model based on both unsupervised and supervised learning was constructed to monitor ice storms~\cite{swope2023using} dynamically. Efforts to improve weather services through digital twin models are also underway. Singh et al. proposed a framework for generating high-resolution urban precipitation data to revolutionize urban climate services, providing city planners with timely and precise climatic information~\cite{singh2023downscalebench}. Koldunov and Jung demonstrated how LLMs can deliver localized climate services by combining the strength of LLMs and model simulations, making manipulations in simulations more accessible~\cite{koldunov2024local}. Chen et al. incorporated vision information into LLMs, proposing a novel Vision-Language Model approach~\cite{chen2024vision}. This integration considerably enhances the analysis of weather heatmaps, improving the speed and accuracy of extreme weather event detection based on the predictive capability of LLMs. Furthermore, weather conditions play a crucial role in energy production. Sehrawat et al. developed a digital twin system for predicting solar irradiance~\cite{sehrawat2023solar}, while Stadtmann et al. designed a digital twin aimed at enhancing wind energy production efficiency~\cite{stadtmann2023standalone}. Savage et al. focused their digital twin research on facilitating collaboration between energy and climate research~\cite{savage2022universal}.

\header{Ocean System} The ocean system is essential to life on Earth, making the understanding of ocean dynamics profoundly significant~\cite{portner2014ocean}. Digital twin models for the ocean cover diverse areas, showcasing their vast potential in understanding the state and changes of the ocean, such as ocean sustainability~\cite{tzachor2023digital}, ocean observation~\cite{chattopadhyay2023oceannet, barbie2021developing}, and coastal flood predictions~\cite{jiang2021digital}. Digital twins play a crucial role in ocean sustainability by reducing and preventing overfishing, modeling and predicting marine pollution and facilitating marine spatial planning~\cite{tzachor2023digital}. Rakotonirina et al. conducted a digital twin study for Ocean Cleanup systems, optimizing system design, predicting fleet performance, and estimating operational costs~\cite{rakotonirina2023digital}. In the application of observing and simulating the ocean, an underwater network of ocean observation systems has been deployed in the Baltic Sea, demonstrating the feasibility of digital twins in extreme underwater environments~\cite{barbie2021developing}. Similarly, Zheng et al. enhanced wave height predictions for ocean energy generation by integrating recurrent neural networks~\cite{zheng2023multivariate}. OceanNet, a physics-informed neural network-based digital twin, models and predicts ocean circulation in regional seas using a Fourier neural operator, promoting stability and mitigating autoregressive error growth over long-term forecasts~\cite{chattopadhyay2023oceannet}. Coastal flood prediction is another notable application. A local digital twin earth project on the Italian coast integrates environmental, demographic, and marine data into a high-precision digital model to monitor, simulate, and predict natural and human activities~\cite{duque2021towards}. Thiria et al. developed an advanced deep learning method for enhancing the spatial resolution of ocean current estimates from satellite observations by merging low-resolution geophysical ocean fields with high-resolution data using convolutional neural networks~\cite{thiria2023downscaling}. Physics-informed machine learning techniques have also been used to create fast and accurate models for coastal flood predictions, significantly accelerating the simulation process while maintaining high accuracy in predicting sea surface height~\cite{jiang2021digital}.

\header{Geological Simulation} Simulation in geology is a vital and long-lasting topic that has gained prominence due to the inherent complexities of underground processes, which are often unobservable directly~\cite{van1989geological, wellmann20183}. Digital twin models have transformed traditional geological simulations due to their precision and efficiency~\cite{elmo2020disrupting, defilipe2022towards}, stimulating the development of areas such as geological modeling~\cite{wang2021complex, wu2022multi}, underground operations~\cite{wu2022multi,nagovitsyn2021digital}, and geological disaster monitoring~\cite{zhang2019monitoring, cheverda2019digital}. In the aspect of geological modeling, Wang et al. employed a digital twin framework to construct complex 3D geological models, incorporating real-time data to refine these models and adapt to new geological conditions~\cite{wang2021complex}. Yang et al. introduced the EdGeo toolkit, a physics-guided generative AI tool for subsurface exploration, which enhances the fidelity of velocity maps through diffusion-based models~\cite{yang2024physics}. Besides, digital twin models substantially aid underground operations. In tunnel construction, dynamically updatable digital twin models are used to manage the continuously evolving geological conditions encountered in long-term projects~\cite{wu2022multi}. In the mining sector, the digital twin system provides updated and precise representations of deposits, supporting informed decision-making in mining operations~\cite{nagovitsyn2021digital}. Additionally, digital twin models are instrumental in monitoring geological disasters. Zhang et al. utilized digital twins for real-time, dynamic predictions of complex and frequent geological disasters, such as landslides, earthquakes, and mudslides, in their monitoring and early warning systems~\cite{zhang2019monitoring}. For seismological research, Vladimir et al. implemented numerical methods and high-performance computing in a digital twin system to create detailed 3D models of geological structures, simulating seismic wave propagation to improve subsurface understanding~\cite{cheverda2019digital}.

\header{Wildlife Protection} Biodiversity possesses inherent value for the natural world and is crucial for maintaining ecosystem functionality~\cite{srivastava2005biodiversity, sharef2022applications}. Wildlife plays a vital role in supporting biodiversity~\cite{mawdsley2009review}. Digital twins can simulate wildlife and provide dynamic, predictive insights that enhance understanding of ecological systems~\cite{ qiu2023construction}. These insights include monitoring animal behaviors and population status, thereby strengthening conservation efforts. Sharef et al. developed an interactive machine learning framework integrated with digital twins to improve biodiversity projection models~\cite{sharef2022applications}. A prototype simulating virus spread among wildlife, aiding in effective management strategies, has been created by Ingenloff~\cite{ingenloff2024prototype}. To mitigate wildlife-vehicle collisions, Moulherat et al. designed an integrated system using sensors and machine learning, leveraging real-time data from camera traps for ongoing wildlife management~\cite{moulherat2023biodiversity}. Fergus introduced an innovative system where digital transactions triggered by camera trap detections compensate local guardians for their conservation efforts, fostering community involvement in biodiversity preservation~\cite{fergus2023empowering}. Rolph expanded the application of digital twins to the management of cultural ecosystem services, merging biodiversity and community recreation to provide a comprehensive assessment of ecosystem services~\cite{rolph2024prototype}. Sakhri adapted digital twin technology to optimize energy use in monitoring waterbirds, demonstrating significant improvements in real-time monitoring capabilities~\cite{sakhri2024digital}. Teschner introduced a novel digital twin-enhanced approach using unmanned aerial vehicles to protect agricultural fields from wildlife intrusion, showcasing the technology's adaptability to various conservation needs~\cite{teschner2022digital}.

\header{Forest Management} Forests contain necessary functionality in maintaining the health of Earth's ecosystems and climate~\cite{putz2010importance}. Digital twin models have the ability to utilize historical data to simulate and forecast changes in forest ecosystems over time~\cite{jiang2022forestry}. Hence, the advent of digital twin technology has brought transformative changes to the field of forest management, marking a significant advancement in how forest ecosystems are monitored, analyzed, and managed~\cite{buonocore2022proposal}. Dynamic ecosystem modeling by Qiu et al. utilizes remote sensing and 3D parametric modeling for real-time forest ecosystem monitoring~\cite{qiu2023forest}. At the same time, Li et al. have developed a robust database for virtual plantation management~\cite{li2023framework}. These technologies enable a highly interactive and dynamic digital twin model for forests that allows for real-time monitoring and decision-making, enhancing the management of forest ecosystems by enabling synchronization between virtual and real interactions. On a national scale, Li et al. utilized a multi-task deep learning network to accurately map and characterize every single tree within and outside forests across Denmark, identifying the location, crown area, and height of trees, thereby enhancing national forest management and conservation efforts~\cite{li2022digital}. Expanding further, Mõttus et al. introduced a high-precision global model under the Digital Twin Earth initiative, enhancing vegetation mapping and characterization across Europe~\cite{mottus2021methodology}. Urban forests also benefit from digital twin technologies, as demonstrated by Ozel and Petrovic. Their Green Urban Scenarios framework incorporates factors such as weather conditions, tree species, diseases, and spatial distributions into simulations, helping to predict the future trajectories and impacts of urban forests under different scenarios~\cite{ozel2023green}. Innovations continue with the introduction of adaptive management strategies using machine learning. Damavsevivcius have applied reinforcement learning algorithms within their digital twin models to optimize forest management~\cite{damavsevivcius2024reinforcement}. Additionally, Zhong~\cite{zhong2023reduced} and Sanchez-Guzman~\cite{sanchez2022modeling} both focus on wildfire management, utilizing digital twin models to predict and manage wildfire dynamics effectively, providing real-time data that helps prevent and mitigate forest fires.

\header{Digital Earth} While researchers have developed digital systems and machine learning models at national, regional, and local levels~\cite{henriksen2022new, sehrawat2023solar, singh2023downscalebench}, the arrival of more advanced deep learning models and powerful computational technology has made the vision of a global digital twin of Earth increasingly feasible. This ambitious project aims to realize global climate projections and assess local impacts simultaneously. It integrates advanced computer science, mathematics, and engineering with interdisciplinary knowledge from natural sciences to create global-scale interactive models of Earth's systems~\cite{li2023big}. This vision has sparked widespread global discussion~\cite{hoffmann2023destination, national2023opportunities}. The Earth-2 project~\cite{NVIDIA2024} exemplifies this ambition by simulating real-time, global-scale meteorological conditions and constructing a replica of the entire Earth's atmospheric system to predict weather conditions, natural disasters, and conduct meteorological explorations. Data-driven physics-informed machine learning and deep learning models strive to embed the rules of the physical world into their frameworks~\cite{karniadakis2021physics}. A notable success in this field is the FourCastNet model, which employs Adaptive Fourier Neural Operators to revolutionize global weather forecasting with high-resolution outputs~\cite{pathak2022fourcastnet}. Building on this, Kurth et al. have demonstrated the scalability and efficiency of FourCastNet on supercomputing systems, paving the way for large-scale, real-time, high-resolution global weather forecasting~\cite{kurth2023fourcastnet}.

\subsection{Agriculture}

As the global population surges and climate change impacts food security, there is a pressing need for agricultural systems to increase production efficiency while minimizing resource consumption. 
In response, digital twin technology has become increasingly essential~\cite{chaux2021digital, purcell2023digital} due to its advantages. 
Digital twins, which are synchronized virtual counterparts of physical objects or systems, offer significant potential to transform agricultural practices by providing detailed virtual replicas of farms and agricultural objects~\cite{peladarinos2023enhancing}. 
These virtual agriculture systems promote smart farming, livestock management, and agricultural facility optimization, thereby offering advantages such as cost savings, improved product quality, and enhanced operational efficiency to effectively address agricultural challenges~\cite{pylianidis2021introducing}. Machine learning and deep learning technologies are being implemented across all facets of agriculture, spanning preproduction, mid-production, and postproduction phases, enabling more accurate simulations~\cite{nie2022artificial}. Researchers highlight that artificial intelligence facilitates data processing and analysis in agricultural digital twin systems, supporting decision-making and providing feedback to virtual systems~\cite{nasirahmadi2022toward, pylianidis2023operationalizing, akshay2024iot}. Contemporary advancements in generative AI have the potential to augment digital twin technologies, addressing longstanding challenges such as farmer-system interactions and agricultural data synthesis~\cite{liu2023exploring}. The integration of artificial intelligence with digital twin technology is propelling significant advancements in agriculture, enhancing the potential for increased crop yields while simultaneously reducing environmental impact and optimizing resource utilization~\cite{shamia2023digital, ubina2023digital, rahman2024digital}.

\header{Smart Farming} Smart farming, a paradigm shift in agricultural practices, leverages an array of modern innovations, including cloud computing, the Internet of Things, machine learning, augmented reality, and robotics to revolutionize agricultural production~\cite{walter2017smart}. As farms increasingly embrace digitalization, the concept of digital twins has emerged as a comprehensive framework that virtualizes every object within the farm, creating digital replicas of crops that mirror their behavior and states throughout their lifecycle in a virtual space, enabling farmers to optimize operations~\cite{verdouw2021digital}. In the realm of irrigation management, Alves et al. have pioneered a sophisticated digital twin system that integrates soil, weather, and crop data to generate daily irrigation prescriptions~\cite{alves2023development}. Nitrogen application, another critical aspect of farming, has also benefited from digital twin technology. Stefano et al. discovered that employing digital twins to simulate diverse scenarios and analyze data empowers farmers to make more informed decisions regarding nitrogen application~\cite{cesco2023smart}. Smith emphasizes the applications of AI in enhancing precision when detecting and measuring farm activities to enable reliable alerts for farmers~\cite{smith2018getting}. Visual neural network techniques have proven particularly effective in monitoring plant health. Nasirahmadi et al. demonstrated the power of coupling these techniques with other smart farming technologies to detect and assess plant health with remarkable accuracy~\cite{nasirahmadi2021sugar}. Integrating multiple smart farming technologies, Angin et al. proposed AgriLoRa, an innovative digital twin framework, which utilizes cloud-based computer vision algorithms to identify plant diseases and nutrient deficiencies, providing farmers with actionable insights to improve their crop management strategies~\cite{angin2020agrilora}.

\header{Livestock Management}
The field of livestock management is another important area in agriculture, facing increasing challenges due to finite resources, the need to reduce greenhouse gas emissions, and a declining global workforce~\cite{akhigbe2021iot}. Digital twin technology has emerged as a promising solution in livestock farming. This technology creates a digital replica that simulates the physical, biological, and behavioral states of animals based on real-time data input, enabling optimal and sustainable livestock management operations~\cite{neethirajan2021digital}. The integration of digital twins with augmented reality (AR) and virtual reality (VR) technologies further enhances the training of veterinary professionals and breeders by providing immersive, interactive learning environments that simulate real-world scenarios, ultimately improving animal welfare and operational efficiency~\cite{petrov2021digital}. In the realm of aquaculture, AI-powered IoT digital systems are revolutionizing fish farming practices by employing sensors and smart devices to collect real-time data on fish metrics, environmental conditions, and health status, enabling automated fish feeding, water quality monitoring, and disease detection~\cite{ubina2023digital}. Advanced research in this area includes replicating the hydrodynamic behaviors of living organisms in aquatic environments using visual neural networks that extract features from video data, combined with computational fluid dynamics to analyze and predict the hydrodynamic cues for fish navigation~\cite{lagneaux2024automatic}. Urban beekeeping has also benefited from digital twin technology. A multiagent model integrating bee populations, beekeepers, non-beekeepers, and the environment has been developed to monitor and analyze bee colony health and behavior, providing a robust tool for decision support for small-scale farmers and urban planners~\cite{johannsen2021digital}. For cattle management, an innovative digital twin model powered by LSTM neural networks has been developed within a farm IoT system to monitor and track cattle's physiological and behavioral states in real time, leveraging the power of artificial intelligence to predict future behaviors and physiological cycles of cattle~\cite{han2022ai}.

\header{Controlled Environment Agriculture}
Controlled Environment Agriculture (CEA) stands in stark contrast to traditional farming methods, focusing instead on artificially constructed agricultural facilities. These facilities, which include physical structures and installations, are designed to foster optimal conditions for enhanced farming quality and efficiency, thereby optimizing resource utilization~\cite{wang2021introductory, skobelev2020development}. Serving as the cornerstone of modern large-scale agriculture, these systems employ advanced monitoring and control technologies~\cite{ amitrano2020crop}. They show immense potential when integrated with Digital Twin (DT) technology, promising to invigorate agricultural practices~\cite{ gonzalez2022monitoring}. Digital twin systems excel in supervising conditions within enclosed structures such as greenhouses or indoor facilities. These sophisticated systems manage and manipulate various environmental factors, including temperature, humidity, light intensity, CO2 levels, and air flow~\cite{skobelev2020development}. By constructing digital replicas of physical farms, a digital twin system can simulate and optimize environmental conditions, enhancing animal welfare and improving farm management~\cite{jo2018smart}. Further advancements in CEA are geared towards increasing energy efficiency and operational effectiveness, particularly in livestock management. A proposed digital twin framework, for instance, replicates physical pig houses in a virtual environment. This enables simulations that optimize heating, ventilation, and air conditioning systems~\cite{jeong2023digital}. In commercial greenhouse production, digital twin technology optimizes energy usage and streamlines production processes. Howard et al. have described a digital twin system that integrates climate control, energy management, and production processes to simulate and refine operational strategies without disrupting ongoing cultivation~\cite{howard2020data}. Prawiranto et al. conducted a comprehensive study on optimizing solar drying processes for fruits. Their approach utilizes a physics-based Digital Twin that incorporates mechanistic fruit drying models, quality models, and weather data to evaluate different improvement strategies~\cite{prawiranto2021physics}.

\subsection{Commerce}

Traditional live commerce, while rapidly growing, faces numerous challenges such as crowding, limited operating hours, and long queues. It also suffers from several limitations, including unengaging content and limited interactivity~\cite{jeong2022innovative,ahmed2023framework}. Digital twinning enables the creation of virtual replicas of physical objects, allowing consumers to interact with products or services in a virtual space that closely mimics real-world conditions~\cite{ahmed2023framework}. This innovative technique can transform both e-commerce and retail stores by constructing digital twins of products and customers, thus enhancing the shopping experience~\cite{liu2024digital, pous2023showing, jenkins2022immersive}. Additionally, its capacity to facilitate collaborative design permits multiple stakeholders such as customers, designers, and manufacturers to engage in the product development process through the convergence of physical and virtual data~\cite{tao2019digital, li2020sustainable, kuzmichev2022application}. Furthermore, with a digital representation of financial data and activities integrated with the capabilities of Large Language Models, a digital twin system can also provide financial services to individuals, institutions, and governments~\cite{anshari2022digital, li2024econagent, gao2024simulating}. Unlike traditional commerce, digital twins utilize real-time data and employ machine learning models~\cite{vijayakumar2020digital}. Deep learning algorithms, such as image recognition, 3D reconstruction, and even generative AI, are used to categorize products, optimize shop layouts, and predict customer needs more accurately~\cite{tao2019digital, jenkins2022immersive, yao2022webshop}. These technologies enhance operational efficiencies in customization and responsiveness, promoting informed decision-making and improving customer experiences in both virtual and physical stores.

\header{E-commerce} 
In the evolving landscape of e-commerce, the integration of digital twin technology has introduced novel applications ranging from immersive shopping experiences to advanced fraud detection~\cite{wang2023sequence} and realistic simulations of consumer behavior~\cite{yao2022webshop}. Jeong and colleagues have pioneered the MBUS platform, combining metaverse elements with live commerce to create a virtual shopping environment, enhancing user engagement by allowing real-time interaction with products~\cite{jeong2022innovative}. Simultaneously, Wang et al. have advanced fraud detection by modeling user behaviors as genetic sequences in the SAGE framework, which employs principles from genetics to enhance the detection of fraudulent activities~\cite{wang2023sequence}. Yao and his team have contributed to WebShop, a large-scale interactive environment that trains language agents to perform web-based tasks by navigating and customizing real-world product interactions based on textual instructions~\cite{yao2022webshop}. Terán and his collaborators have developed an agent-based simulation model that incorporates word-of-mouth dynamics and endorsement theory to more accurately represent market behaviors~\cite{teran2023modeling}. Furthermore, Kuzmichev explores the application of digital twins in the fashion industry, focusing on 3D digital garment design~\cite{kuzmichev2022application}. This process significantly reduces labor and material costs by using virtual human models and virtual garments to simulate the interaction between the body and clothing. Fu et al. developed an innovative Augmented Reality try-on system for virtual clothing digital twins, featuring real-time interactions and realistic cloth simulation through an optimized framework, which has the potential to significantly enhance customer experiences in virtual fashion retail environments~\cite{fu2023hanfu}.

\header{Retail} The collection and analysis of real-time data has become the engine for continued growth in the retail industry~\cite{raji2024real}. Digital twin technology can serve as a framework for data integration, providing intelligent analysis for various aspects of the retail industry and facilitating decision-making~\cite{kanaga2024review}. Vijayakumar explores the concept of a behavioral digital twin, which models consumer behaviors to tailor interactions and predict future purchases, aiming to significantly enhance customer satisfaction through personalized experiences~\cite{vijayakumar2020digital}. In a different application, Shoji et al. focus on the postharvest life of imported fruits, utilizing physics-based digital twins to control and monitor hygrothermal conditions, thus improving fruit quality from packhouse to retail stores~\cite{shoji2022mapping}. Sengupta and Dreyer demonstrate how digital twins can facilitate sales and operations planning in grocery retail, helping to predict and manage variabilities across the value chain to minimize waste~\cite{sengupta2023realizing}. Additionally, Liu et al. present a multi-modal approach that employs AI-driven real-time product recognition and 3D store reconstruction to create a precise virtual representation of physical retail spaces, enhancing both operational efficiency and customer interaction~\cite{liu2024digital}. Pous et al. introduce an innovative use of robots that scans item locations within a store, integrated into a digital twin, providing customers with real-time product information~\cite{pous2023showing}. Stacchio et al. discuss the social acceptance and potential benefits of Human Digital Twins in fashion retail, highlighting how 3D models of customers can improve service and customer satisfaction in brick-and-mortar stores~\cite{stacchio2022will}.

% \header{Product design} Digital twin technology is revolutionizing various sectors, including product design and manufacturing, by facilitating more efficient and innovative processes. Tao et al. introduce the concept of digital twin-driven product design (DTPD), which supports collaborative design and enables real-time interaction between designers and the digital twin~\cite{tao2019digital}. These interactions foster iterative improvements by leveraging feedback from both virtual simulations and real-world usage to expedite and innovate design practices. In a practical application of digital twin platforms, Li and colleagues discuss the Digital Twin framework that connects smart home devices, facilitating real-time data exchange and personalized services~\cite{li2020sustainable}.This framework exemplifies how digital twins can support mass customization and efficient resource utilization, enhancing stakeholder engagement and driving community-led innovation.

\header{Financial Service and Analysis} The integration of digital twin technology in financial management and economic simulations is reshaping the capabilities of existing tools by incorporating real-time data, LLMs, and more human-like decision-making processes. Anshari and colleagues discuss the application of DT technology within robo-advisors, transforming them from static tools into dynamic financial advisory platforms. This integration allows for continuous improvement through interaction and real-time data updates, providing users with more tailored and responsive financial management solutions~\cite{anshari2022digital}. Li et al. present EconAgent, an agent-based modeling framework that leverages LLMs to simulate macroeconomic phenomena, enabling more realistic decision-making based on work, consumption, and past economic trends~\cite{li2024econagent}. In another innovative approach, the Agent-based Simulated Financial Market utilizes LLMs to mimic real human traders and simulate stock market behavior. The ASFM framework includes a realistic order matching system and simulates various industry sectors, creating agents with diverse profiles and strategies that can understand market dynamics and respond to economic news~\cite{gao2024simulating}. Veshneva formulated a novel approach for constructing digital twins of socio-economic systems using status function-based mathematical models, enhancing predictive analytics and decision-making in complex, uncertain environments~\cite{veshneva2022construction}. Barkalov et al. envisioned an innovative distributed forecasting information system architecture integrating digital twins and recurrent neural networks for predictive asset maintenance in socio-economic systems, demonstrating improved accuracy in predicting remaining useful life and potential asset failures~\cite{barkalov2021application}.

\subsection{Education and Training}

Over the past four years, the digital twin technique in education has gained popularity, driven by safety concerns associated with traditional offline education during the COVID-19 pandemic~\cite{sepasgozar2020digital, han2022intelligent}. As virtual modules that replicate real-world construction processes, digital twins can blend online and offline learning, emerging as a solution to ensure the continuity and quality of education~\cite{kartashova2020digital}. Some researchers argue that due to the rapid acceleration of knowledge doubling, traditional education methods are becoming limited~\cite{kinsner2021digital, nikolaev2018implementation}. The dynamic digital twin model facilitates an integrated educational environment, merging physical and digital spaces to provide comprehensive educational resources, such as traditional knowledge, experience, and societal wisdom. It also enhances tailored learning experiences by integrating various aspects of a student's academic and personal life~\cite{kartashova2020digital, akhmedov2023prospects}. The digital twin system enhances learning outcomes by enabling learners to experience efficient data communication and interaction, facilitating the understanding of complex, practical knowledge through virtual reality integration, as demonstrated by existing results~\cite{zhang2022influences}. Furthermore, by leveraging advanced technologies such as the Internet of Things, artificial intelligence, virtual reality, and 5G, educational institutions can create more effective and personalized learning environments~\cite{akhmedov2023prospects, shuguang2020holographic}. Recently, more articles have demonstrated the potential of generative AI, especially LLMs and agents, as revolutionary tools to boost simulations and enhance educational methods~\cite{wang2024large,chu2025llm,yue2024mathvc, zhang2024simulating}.

\header{Virtual Learning}
Applications of digital twins in educational settings, ranging from immersive virtual environments to hands-on learning tools, significantly enhance student engagement and understanding across various disciplines~\cite{nikolaev2018implementation, sepasgozar2020digital, madni2019exploiting}. For instance, Nikolaev et al. implemented a digital twin system that enables MSc students to create virtual models of tunnel boring machines, enriching the learning experience and sparking increased interest in the program~\cite{nikolaev2018implementation}. Madni et al. have engaged students in developing digital twins by instrumenting physical vehicles with sensors, helping students understand vehicle dynamics and control techniques, and moving beyond traditional lecture-based learning~\cite{madni2019exploiting}. Furthermore, Han et al. have developed a comprehensive digital twin of campus using UAV tilt photography and 3D modeling, integrated via Unity3D to facilitate real-time monitoring and decision-making~\cite{han2022intelligent}. Razzaq et al. have pioneered DeepClassRooms, a digital twin framework that utilizes convolutional neural networks for enhanced attendance tracking and content delivery monitoring in public sector schools, demonstrating the adaptability of digital twins in addressing educational challenges in resource-limited settings~\cite{razzaq2023deepclassrooms}. Zhang and colleagues have introduced SimClass, a cutting-edge framework that simulates traditional classroom settings using LLMs. This framework improves educational experiences by allowing LLMs to assume various classroom roles, which promotes cognitive and social development as well as collaborative behaviors, enriching overall classroom dynamics~\cite{zhang2024simulating}.

\header{Interactive Learning}
Digital twin technology in education is advancing through innovative applications to foster collaborative and interactive learning environments~\cite{lan2024teachers, zhang2024simulating}. Digital twin systems use data analytics and AI to adjust the content difficulty, pacing, and learning paths in real-time, personalizing the learning experience based on individual student performance and needs~\cite{ yue2024mathvc, murtaza2024transforming}. Sepasgozar has employed these technologies to allow students to safely explore complex construction activities through the VTBM module, an immersive virtual environment~\cite{sepasgozar2020digital}. Innovatively, Lee and colleagues have introduced game-like elements within the digital twin framework to demystify mathematical concepts and engage students in problem-solving activities~\cite{lee2023digital}. In the holographic classroom envisioned by Liu and Ba, digital twins provide an immersive and interactive 3D learning space, merging physical and virtual worlds to enhance educational experiences and introduce dynamic teaching methods~\cite{shuguang2020holographic}. Lan and Chen have developed a team-teaching framework where human educators are paired with AI agents, designed to support personalized learning and real-time feedback. This approach enhances pedagogical effectiveness while preserving the dynamics of human teaching~\cite{lan2024teachers}. In mathematics education, Yue et al. have developed MATHVC, a virtual platform where LLMs simulate student interactions in a multi-character setup. This platform encourages collaborative problem-solving and offers a scalable method to practice mathematics without continuous teacher supervision, thereby enhancing student engagement and learning autonomy~\cite{yue2024mathvc}. Murtaza and his team explore the application of ChatGPT in driver education, comparing traditional instructional methods with interactive, ChatGPT-based learning. Their findings indicate that participants trained with ChatGPT exhibit significantly better learning outcomes, demonstrating the effectiveness of LLMs in practical applications~\cite{murtaza2024transforming}.

\subsection{Quantum Computing for Digital Twin}
\header{Motivation: Why Digital Twins Need Quantum Computing}
% As digital twin (DT) technologies evolve toward modelling ever more ambitious systems—multi-physics, high-dimensional, real-time, and highly coupled—the computational burden of simulation, inference, optimisation and control escalates rapidly. Classical digital twin architectures (sensor-data ingestion + AI/ML + numerical simulation) begin to hit bottlenecks in three key areas:
% 强调了QC对DT任务在复杂度层面的写法，把“难”具体落实到高维线性代数、随机模拟和组合优化这几个方面，新加了Montanaro的量子算法综述来作为解决复杂度问题的support literature。
As digital twin (DT) technologies evolve toward modelling ever more ambitious systems—multi-physics, high-dimensional, real-time and tightly coupled—the computational burden of simulation, inference, optimisation and control escalates rapidly. Many core DT workloads can be cast as high-dimensional linear algebra, stochastic simulation and large-scale combinatorial optimisation problems, whose classical complexity typically grows polynomially or even exponentially with the relevant state, parameter or decision-space dimension. Even with modern high-performance computing (HPC) and accelerator hardware, these scaling limits constrain both the achievable spatial/temporal resolution and the breadth of scenario exploration within operational time budgets~\cite{montanaro2016quantum}.

% 原来的写法是classical architectures+bottlenecks，现在在此基础上新加上：即便有HPC，仍存在复杂度层面的bottleneck，而不是单纯的算力不够。
Classical digital twin architectures—typically comprising sensor-data ingestion, AI/ML-based surrogate modelling and numerical simulation—therefore begin to hit intrinsic bottlenecks in three key areas, even when backed by large-scale HPC infrastructure:

% 1. \textbf{Scale and Coupling:} Systems such as whole smart-cities, biological networks, aerospace systems or large manufacturing plants involve high-dimensional state-spaces, strong non-linearities, multi-domain physics (mechanical, thermal, electrical, biological), and real-time feedback loops. The complexity of modelling, simulating and controlling such systems is daunting for purely classical computing methods.
1. \textbf{Scale and Coupling:} Systems such as whole smart-cities, biological networks, aerospace systems or large manufacturing plants involve high-dimensional state-spaces, strong non-linearities, multi-domain physics (mechanical, thermal, electrical, biological), and real-time feedback loops. In practice, this translates into repeatedly solving large coupled PDEs/DAEs, high-dimensional Bayesian inference problems, and global or mixed-integer optimisation problems over many control and design variables. The complexity of modelling, simulating and controlling such systems at high fidelity is daunting for purely classical methods and leads to severe trade-offs between resolution, coverage and latency.

% 2. \textbf{Real-time/Near-real-time Operation:} Digital twins are increasingly used not only for offline analysis but for operational decision-making, fault detection/prediction, autonomous control and adaptation. Achieving low-latency, high-fidelity simulation and inference across large systems is challenging for classical methods.
% 新加了实时性+量子硬件时延的现实限制，根据我们的经验来说，目前的量子概念更可能用于离线或者批处理sub tasks，直接嵌入低时延的闭环控制不太现实。
2. \textbf{Real-time/Near-real-time Operation:} Digital twins are increasingly used not only for offline analysis but for operational decision-making, fault detection/prediction, autonomous control and adaptation. Achieving low-latency, high-fidelity simulation and inference across large systems is challenging for classical methods, especially when decisions must be updated at time scales comparable to measurement rates. In the near term, any quantum acceleration is more realistically applied to computationally intensive but latency-tolerant subroutines (e.g., periodic re-optimisation, scenario generation, policy improvement), which indirectly support real-time DT operation rather than sitting directly in millisecond-level feedback loops.

% 3. \textbf{Complex Optimisation and Inference Under Uncertainty:} Typical DT tasks include combinatorial optimisation (scheduling, routing, control), probabilistic inference with uncertainty quantification, inverse problems (e.g., deducing system state or faults from sensor data), and high-fidelity simulation of physical phenomena. Many of these are computationally intense and scale poorly in classical approaches.
% 原本只是简单对复杂优化与不确定性推理进行讨论，但是肯定要深入到具体的quantum algorithms，新加了映射到QAOA、量子线性方程、量子蒙特卡洛等算法族的讨论，而且需要说明目前end-to-end优势仍是开放问题，不然很容易被审稿人攻击。
3. \textbf{Complex Optimisation and Inference Under Uncertainty:} Typical DT tasks include combinatorial optimisation (scheduling, routing, control), probabilistic inference with uncertainty quantification, inverse problems (e.g., deducing system state or faults from sensor data), and high-fidelity simulation of physical phenomena. These tasks map naturally onto classes of quantum algorithms that promise asymptotic speedups in query or sample complexity under suitable assumptions: quantum approximate optimisation algorithms (QAOA) and related variational methods for combinatorial problems, quantum amplitude estimation and quantum-accelerated Monte Carlo for uncertainty quantification, and quantum linear solvers and PDE solvers for large-scale simulation and inverse problems~\cite{farhi2014quantum,childs2021high,bravo2023variational,montanaro2016quantum}. However, rigorous end-to-end quantum advantages for full DT workflows remain an open research question, particularly on noisy intermediate-scale quantum (NISQ) hardware.

% In this landscape, quantum computing (QC) presents a compelling potential: leveraging superposition, entanglement and quantum parallelism, QC offers opportunities for significant speed-ups or capability enhancements in simulation, optimisation and learning. The emergent concept of a Quantum Digital Twin (QDT) is defined as the integration of quantum processing into digital twin frameworks, to address classical bottlenecks. According to recent work, the QDT paradigm is not a mere buzzword but a hybrid approach combining classical DT infrastructure with quantum modules, aimed at enabling ``real-time simulation of highly complex, interconnected entities. \cite{amir2022can, zhang2023quantum, patsnap2025quantum}" Hybrid quantum–classical architectures are highlighted as the realistic near-term pathway: classical systems maintain sensor interface, control logic and standard ML, while quantum modules tackle the “hard core” tasks such as large-scale simulation, optimisation and inference \cite{zhang2023quantum}.
% 把原来的QDT定义部分重写了一下，减少“superposition/entanglement/parallelism”这种用法（比较泛泛），改为强调“算法与复杂度优势”；同时把QDT明确为混合架构。
In this landscape, quantum computing (QC) is not a monolithic replacement for classical DT infrastructure but a prospective source of algorithmic acceleration for specific computational kernels. Rather than focusing on physical phenomena such as superposition and entanglement, it is more useful at the DT level to view QC through the lens of quantum algorithms for simulation, optimisation and learning, many of which admit provable or conjectured complexity-theoretic advantages over the best-known classical counterparts~\cite{montanaro2016quantum,cerezo2021variational}. We use the term \emph{Quantum Digital Twin} (QDT) to denote a hybrid quantum–classical DT architecture in which classical components manage sensor interfaces, data pre-processing, domain-specific logic and most control tasks, while quantum processing units (QPUs) are invoked as accelerators for carefully selected ``hard-core'' subproblems (e.g., large-scale optimisation, high-dimensional inference, or fine-grained physical simulation)~\cite{zhang2023quantum,otgonbaatar2024quantum}. Recent position papers and industrial case-studies have started to articulate such QDT architectures and early prototypes in domains including manufacturing, smart-city operation and supply chains~\cite{amir2022can,zhang2023quantum,patsnap2025quantum}.

% 单独加了QML/QAI在QDT中的角色，perception-prediction-decision-making，并新加了QML综述和feature map工作；同时说明目前NISQ限制和VQA在混合架构中的表现。
Beyond simulation and optimisation, QDTs may incorporate quantum machine learning (QML) or quantum AI (QAI) modules as high-capacity function approximators for perception (feature extraction from high-dimensional sensor data), prediction (state and parameter forecasting) and decision-making (policy learning and planning). QML approaches based on variational quantum circuits and quantum feature maps are being actively explored as a way to trade classical sample or computational complexity for quantum circuit depth and qubit resources~\cite{biamonte2017quantum,schuld2019quantum}. At the same time, NISQ-era devices impose strict limitations in terms of qubit count, noise and connectivity, so near-term QDTs are likely to deploy quantum modules in offline or batch modes and to rely heavily on hybrid schemes such as variational quantum algorithms (VQAs), which combine parameterised quantum circuits with classical optimisation~\cite{cerezo2021variational}.

% 新加的这一段把后面11个领域abstract到4类通用量子模块（模拟、优化、QML/QAI、安全），从toolbox的角度展示应用，而不是11个割裂的case。
From an architectural perspective, it is useful to distinguish several generic types of quantum modules that can be reused across DT domains: (i) quantum simulation modules for high-dimensional linear systems, PDEs or quantum chemistry; (ii) quantum optimisation modules for combinatorial and continuous optimisation; (iii) QML/QAI modules for supervised, unsupervised and reinforcement learning; and (iv) quantum-enhanced cryptography and secure communication primitives. Each application domain below can then be viewed as instantiating different combinations of these generic modules according to its dominant computational bottlenecks.

\header{Quantum Digital Twins Across Application Domains}
% Below we follow the domain structure of Sections 4.1–4.11 and discuss how quantum computing can enhance digital twins in each domain. For each, we highlight the key potential and cite relevant literature where available.
% 新加的部分强调下面是问题到量子模块的映射，而非每个领域都已有成熟QDT，以降低hype。
Below we follow the domain structure of Sections 7.1–7.11 and discuss how quantum computing can enhance digital twins in each domain. For each, we highlight the key potential and map DT workloads to candidate quantum modules, citing relevant literature where available. In several domains, these mappings remain largely conceptual, reflecting the early stage of QDT research and deployment.

1. \textbf{Healthcare Systems}

% Digital twins in healthcare simulate patient-specific organs, physiological systems, treatment planning, and monitoring. When augmented with quantum computing, three main enhancements appear: (i) quantum molecular dynamics or quantum chemistry simulations for drug-target interaction and personalized medicine; (ii) quantum machine learning (QML) for patient‐specific disease progression prediction under uncertainty; (iii) quantum optimisation for treatment scheduling and resource allocation. Recent studies highlight that quantum digital twins enable modelling of complex biological interactions with greater dimensionality and coupling than conventional twins \cite{SAINI202537}.
% 把quantum chemistry/QML/优化具体化：（1）量子化学更偏中长期、更多是离线先验/参数；（2）QML关联具体数据模态和任务；（3）治疗排程等映射到NP-hard优化，需要注意得提及说明目前多为小规模的应用。
Digital twins in healthcare simulate patient-specific organs, physiological systems, treatment planning and longitudinal monitoring. When augmented with quantum computing, at least three main enhancements can be envisaged. First, in the longer term, quantum chemistry and quantum molecular simulation algorithms could improve the fidelity of drug–target interaction models and thereby supply more accurate mechanistic priors or parameter sets to patient-level DTs for personalised medicine~\cite{cao2019quantum}. In the near term, such quantum simulations are more likely to be used offline for molecular screening and model calibration rather than in real-time clinical workflows. Second, QML/QAI models may be used to analyse heterogeneous, high-dimensional patient data—such as longitudinal electronic health records, imaging and multi-omics—to predict disease trajectories and treatment responses under uncertainty, for example via quantum kernel methods or variational quantum classifiers~\cite{biamonte2017quantum,schuld2019quantum}. Third, treatment scheduling, radiotherapy planning and hospital resource allocation can be formulated as large-scale combinatorial optimisation problems potentially addressable by QAOA-like or annealing-based quantum optimisation heuristics~\cite{farhi2014quantum}. Recent studies on healthcare digital twins suggest that incorporating such quantum-enabled modules could expand the dimensionality and coupling that can be handled in patient-specific models, although practical QDT deployments in clinical settings remain at a very early stage~\cite{SAINI202537}.

2. \textbf{Biological System}

% For broader biological systems (metabolic networks, cellular communities, ecosystem–organism interactions), QDTs enable simulation of complex bio-reaction networks, stochastic heterogeneity and adaptive behaviours at scales that classical models struggle with. Hybrid quantum–classical digital twins incorporating uncertainty quantification are being explored \cite{otgonbaatar2024quantum}.
% 得减少与Healthcare的部分主旨上重复，主要具体突出系统生物学中的“网络+随机反应”的复杂性，并与Otgonbaatar等关于QDT+UQ的工作相呼应，得强调目前多为概念性和小规模探索。
For broader biological systems (metabolic networks, cellular communities, ecosystem–organism interactions), QDTs could target the computational core of stochastic reaction–diffusion models and network dynamics. For example, analysing gene-regulatory or signalling networks often reduces to dynamics on large, sparse graphs and to sampling from high-dimensional stationary distributions, which might benefit from quantum walk-based algorithms or quantum-accelerated sampling schemes. Single-cell omics data, on the other hand, naturally lead to very high-dimensional, sparse data analysis problems in clustering and trajectory inference, where QML-based representation learning could, in principle, provide more expressive embeddings. Early work on hybrid quantum–classical digital twins with explicit uncertainty quantification suggests that quantum-enhanced modules may be useful for propagating uncertainty through such complex networked models~\cite{otgonbaatar2024quantum}, although concrete large-scale biological QDT deployments are not yet available.

3. \textbf{Aerospace}

% Aerospace systems (aircraft, satellites, launch vehicles) include coupled structural/thermal/fluids/control subsystems, stringent safety requirements and real-time decision support. Quantum digital twins can enhance: (i) trajectory optimisation via quantum algorithms; (ii) high-fidelity simulation of aerodynamics, thermal loads and fatigue; (iii) predictive maintenance and fault diagnosis in complex avionics networks. Literature points to quantum digital twins as enabling real-time optimisation and autonomous decision making in space applications \cite{rfglobalnet2025quantum}.
% 具体得细化三类任务：（1）轨迹/星座配置→组合优化→QAOA/annealing；（2）CFD/结构分析→量子线性方程与PDE算法，需要声明目前仍与工程级网格有较大距离；（3）预测性维护→QML离线训练+经典部署。
Aerospace systems (aircraft, satellites, launch vehicles) include coupled structural, thermal, fluid-dynamic and control subsystems, stringent safety requirements and real-time decision support. In this setting, QDTs can, in principle, enhance several classes of tasks. Trajectory optimisation, constellation design and mission planning give rise to large discrete or mixed-integer optimisation problems, natural candidates for QAOA-type or annealing-based quantum optimisation modules~\cite{farhi2014quantum}. High-fidelity aerothermodynamic and structural simulations involve solving very large sparse linear systems or PDEs, for which quantum linear solvers and quantum PDE algorithms promise asymptotic speedups in certain regimes~\cite{childs2021high,tosti2022review}. However, bridging these theoretical advantages to industry-grade meshes, complex boundary conditions and certification constraints remains a long-term challenge. Finally, predictive maintenance and fault diagnosis in avionics and structural health monitoring could exploit QML models trained offline on telemetry and sensor streams, with distilled classical surrogates deployed in safety-critical, real-time DT instances. Emerging aerospace-oriented QDT concepts emphasise real-time optimisation and autonomous decision making in space applications, but so far remain largely at the roadmap or prototype stage~\cite{rfglobalnet2025quantum}.

4. \textbf{Smart City}
% Smart city DTs integrate transport, energy, buildings, environment, infrastructure at large scale. Quantum computing allows city-scale simulation (traffic/energy/thermal flows) with higher fidelity, quantum optimisation for scheduling (traffic signals, energy grid dispatch), and quantum cryptography for secure twin-to-twin communications \cite{patsnap2025quantum}.
% 重点是“多主体+多目标”的城市问题结构，将其拆解为network flow optimisation、unit commitment and demand response、multi-agent coordination，并将量子优化/QML对应到具体子任务，最后强调量子安全通信与关键基础设施数据的关联。
Smart city DTs integrate transport, energy, buildings, environment and infrastructure at urban scale. The resulting workloads combine network flow optimisation (for traffic and logistics), unit commitment and demand response (for energy systems), and multi-agent coordination (for mobility and services). Quantum optimisation algorithms could be used to tackle large combinatorial subproblems such as traffic signal timing plans, demand-response scheduling or distributed energy resource dispatch within hybrid planning frameworks~\cite{patsnap2025quantum}. QML models for spatio-temporal forecasting may aid in predicting traffic, energy demand and environmental conditions under uncertainty, improving the robustness of city-level DT predictions. Moreover, quantum-safe cryptography and, in selected settings, quantum key distribution can be integrated into smart-city QDT architectures to secure twin-to-twin and cross-organisational communication links involving critical infrastructure data, complementing conventional cybersecurity mechanisms.

5. \textbf{Mobility and Transportation}

% In transportation (vehicle fleets, network flows, autonomous mobility), digital twins support routing, real-time prediction, safety. Quantum enhancements include quantum-enabled combinatorial optimisation for vehicle routing or traffic flow, real-time decision making under uncertainty for autonomous systems, and large-scale scenario simulation for mobility infrastructure \cite{patsnap2025quantum}.
% 减少Smart City段的主旨相似程度，focus on车队/网络层面的路由与调度，强调VRP等NP-hard问题与QAOA/annealing的自然对应，同时得考虑真实应用情况，目前量子更多用于离线训练/场景生成而非直接控制。
In transportation (vehicle fleets, network flows, autonomous mobility), digital twins support routing, real-time prediction and safety-critical decision-making. QDTs may contribute quantum-enabled combinatorial optimisation for vehicle routing, crew and charging scheduling, and empty-vehicle repositioning problems, many of which can be expressed as large QUBO instances amenable to QAOA or annealing heuristics~\cite{farhi2014quantum,patsnap2025quantum}. For autonomous and connected vehicles, quantum reinforcement learning and planning methods might be explored for complex multi-agent decision-making under uncertainty, albeit primarily in offline training and evaluation loops rather than latency-critical control. Large-scale scenario simulation and rare-event analysis, potentially accelerated by quantum Monte Carlo techniques, could further enhance the predictive power of mobility DTs.

6. \textbf{Smart Manufacturing}

% Manufacturing is among the most mature DT domains. By integrating quantum computing, the digital twin can support production-line global optimisation, equipment predictive maintenance, process parameter fine-tuning, dynamic reconfiguration of manufacturing systems. For instance, a partnership between Bosch and Multiverse Computing is developing quantum digital twins for manufacturing processes \cite{zhang2023quantum}.
% 把目前应用的最成熟领域具体化到job-shop/flow-shop等经典NP难问题，然后新加入Bosch+Multiverse量子DT合作的真实应用作为产业案例，同时强调目前仍停留在PoC/原型阶段。
Manufacturing is among the most mature DT domains. Integrating quantum computing yields several concrete opportunities. Production planning, job-shop and flow-shop scheduling, and dynamic reconfiguration of manufacturing cells are classical NP-hard optimisation problems that can be encoded as QUBO or Ising models and targeted by QAOA-style or annealing-based quantum optimisation modules~\cite{farhi2014quantum,zhang2023quantum}. On the analytics side, QML models can be trained on high-dimensional process data, vibration and image streams for predictive maintenance and quality monitoring. Industrial collaborations between Bosch and Multiverse Computing, for instance, have begun to investigate quantum and quantum-inspired algorithms as accelerators within manufacturing DT simulation workflows, including prototype QDTs for factory processes~\cite{BoschMultiverse2022}. These efforts remain at the proof-of-concept stage, but they illustrate how quantum modules can be integrated into existing industrial DT pipelines.

7. \textbf{Robotics}

% For robotics (collaborative robots, autonomous drones/vehicles), digital twins must model kinematics, dynamics, sensing, environment interactions and control loops in real time. Quantum digital twins can deliver quantum reinforcement learning (QRL) or quantum optimisation for control policies, accelerated path planning, robust decision-making under uncertainty. Though research is less mature, the potential is substantial.
% 强调tight real-time constraints与量子硬件时延之间的矛盾，把QDT在机器人中的角色相对更合理地定位为离线策略优化/复杂规划，而非伺服级闭环控制。
For robotics (collaborative robots, autonomous drones/vehicles), digital twins must model kinematics, dynamics, sensing, environment interactions and control loops under tight real-time constraints. Quantum enhancement is therefore more realistic in computationally heavy but latency-tolerant components such as motion planning, task allocation and policy optimisation, rather than in low-level servo control. For example, multi-robot task allocation and path planning can be formulated as combinatorial optimisation problems for QAOA or annealing-based solvers, while QRL or quantum-enhanced policy search methods may be explored for learning control policies in complex, partially observed environments. In all cases, quantum modules are likely to operate offline or in supervisory loops, with policies or planners ultimately executed by classical controllers in the physical robots.

8. \textbf{Natural Systems and Environment}

% Environmental systems (climate, atmosphere, water/earth systems, pollution dispersion, ecosystem dynamics) are inherently high-dimensional, stochastic and multi‐scale. QDTs enable advanced simulation of climate interactions, pollution spread, ecological responses, and optimisation of environmental interventions (resource allocation, sustainability). A recent quantum-DT project in green hydrogen plant optimisation is one such example \cite{swayne2023multiverse}.
% 把环境/气候DT的计算核心明确为PDE+随机PDE+Monte Carlo UQ，并对应到已有量子线性方程/PDE/Monte Carlo算法文献，再保留green hydrogen QDT项目作为过程案例。
Environmental systems (climate, atmosphere, water/earth systems, pollution dispersion, ecosystem dynamics) are inherently high-dimensional, stochastic and multi-scale. DTs in this domain rely heavily on numerical solution of PDEs, stochastic differential equations and large sparse linear systems, as well as Monte Carlo sampling for uncertainty quantification and extreme-event analysis. Quantum algorithms for linear systems and PDEs, and quantum Monte Carlo via amplitude estimation, offer asymptotic advantages for some of these tasks~\cite{childs2021high,tosti2022review,montanaro2016quantum}. In practice, near-term applications are more likely to involve reduced-order or surrogate models whose most expensive components are offloaded to quantum accelerators. A recent quantum-DT project in green hydrogen plant optimisation exemplifies a process-level environmental QDT, where quantum optimisation is explored for improving operational efficiency under stochastic renewable input and demand conditions~\cite{swayne2023multiverse}.

9. \textbf{Agriculture}

% Agricultural digital twins model soil-crop-climate coupling, resource flows (water, nutrients), crop growth prediction, farm-operation scheduling. Quantum computing can support large-scale crop yield prediction under uncertainty, resource allocation optimisation, simulation of precision-agriculture systems at scale. Although concrete case studies remain rare, the domain holds large potential \cite{patsnap2025quantum}.
% 减少和环境/供应链部分的重复，因为是农业，所以DT的spatial-temporal、多目标（yield, cost, environmental impact）、多源数据（satellite remote-sensing, in-field IoT）都得突出；同时用已有供应链QML+DT/QDT工作的架构思路，说明农业目前的实际应用还未正式有案例，但是有前瞻性的选择。
Agricultural digital twins model soil–crop–climate coupling, water and nutrient flows, crop growth and farm-operation scheduling, typically over large spatial domains and long time horizons. The resulting decision problems (e.g., irrigation and fertiliser scheduling, machinery deployment, crop rotation planning) are high-dimensional, stochastic and subject to multiple, sometimes conflicting objectives (yield, cost, environmental impact). Quantum optimisation could be used to explore large combinatorial decision spaces in multi-objective planning under constraints, while QML models may help fuse satellite remote-sensing, in-field IoT and weather data to estimate latent soil–crop states and to forecast yields under uncertainty. Although concrete QDT case studies in agriculture are currently rare, recent work on QML–DT integration and quantum-enhanced uncertainty handling in supply-chain settings suggests that similar hybrid quantum–classical architectures could benefit agricultural planning and logistics~\cite{abdullah2024uncertainty,abdi2025quantum}.

10. \textbf{Commerce}

% In commerce, digital twins simulate supply chains, asset lifecycles, customer-behaviour, risk modelling. Quantum–classical hybrid DTs have emerged for uncertainty quantification in supply-chain digital twins; quantum feature transformations improve resilience under dynamic conditions \cite{otgonbaatar2024quantum}. Quantum optimisation also applies to pricing strategies, inventory management, secure transaction modelling.
% 供应链DT+量子的部分得细化，分解为network design、positioning、routing、risk and uncertainty management等子任务，并用供应链DT+量子特征变换/UQ文献作支撑，仍然声明目前结果多为小规模上的测试，尽可能的别制造应用规模上的误会。
In commerce, digital twins simulate supply chains, asset lifecycles, customer behaviour and risk. QDTs here naturally build on quantum optimisation and QML modules. Network design, facility location, inventory positioning and routing can be formulated as large-scale combinatorial optimisation problems suitable for QAOA-type or annealing-based quantum heuristics, possibly embedded in multi-stage stochastic optimisation frameworks. For risk and uncertainty management, recent work has proposed hybrid quantum–classical supply-chain digital twins in which quantum feature transformations and variational classifiers are used to improve demand forecasting, anomaly detection and uncertainty propagation, with the DT serving as a testbed for policy evaluation~\cite{abdullah2024uncertainty,abdi2025quantum}. These studies report potential gains in forecast accuracy and computational efficiency on small-scale testbeds, but emphasise that scalability and hardware noise remain significant challenges. Quantum-safe cryptography and secure multiparty computation techniques can further complement QDT-based financial and supply-chain systems, although these are orthogonal to the core simulation and optimisation roles of quantum computing. Quantum–classical hybrid DTs for uncertainty quantification in noisy quantum devices themselves represent another line of work, highlighting the use of QDT concepts to analyse quantum hardware performance~\cite{otgonbaatar2024quantum}.

11. \textbf{Education and Training}

% Educational/training digital twins (virtual labs, immersive learning, complex system simulations) can be enhanced by QDTs: quantum simulation of physics/chemistry experiments, virtual environments including high-fidelity quantum modules, adaptive learning pathways derived from QML. Though early stage, this is ripe for innovation and aligns with frontier educational methodologies.
% 把教培DT中的QDT角色拆成两类：（1）作为high fidelity（比如quantum acc）物理/工程实验平台；（2）作为学习过程DT+QML做个性化路径生成，需要声明目前多为概念与探索性工作，真实落地是未来的目标。
Educational and training digital twins (virtual labs, immersive learning, complex system simulators) can benefit from QDTs in two complementary ways. First, DTs that emulate complex physical or engineered systems can incorporate quantum-accelerated simulation modules—for example, in quantum physics, chemistry or materials science—to provide learners with interactive access to phenomena that are otherwise computationally or experimentally inaccessible~\cite{cerezo2021variational,cao2019quantum}. Second, DTs of the learning process itself may employ QML/QAI models to construct personalised learning paths and adaptive training scenarios based on rich learner interaction data, with the DT providing a sandbox for testing pedagogical policies before deployment. At present, concrete QDT deployments in education and training remain largely conceptual, but the combination of DTs with QML/QAI aligns with broader trends in data-driven, simulation-based education.

\section{Open Challenges and Future Directions}\label{sec:8}
While AI-empowered digital twins have demonstrated transformative potential across diverse domains, several fundamental challenges remain unresolved. This section identifies critical open problems and outlines promising research directions that require concerted efforts from the research community.

\header{Bridging Physics and AI}
Despite progress in physics-informed neural networks (PINNs) and hybrid modeling, a fundamental tension persists between data-driven AI and physics-based simulations~\cite{karniadakis2021physics}. Multi-scale integration remains challenging—cardiovascular digital twins, for instance, must seamlessly model from molecular biochemistry to organ-level hemodynamics while maintaining computational efficiency~\cite{rasheed2020digital}. Uncertainty quantification in hybrid models lacks principled methods for combining uncertainties from both physics-based and data-driven components, critical for high-stakes applications like aerospace and healthcare. Ensuring AI models preserve physical constraints such as conservation laws and causality during long-horizon predictions calls for architectures with embedded physical priors, including Hamiltonian neural networks and symplectic integrators.

\header{Scalability and Real-Time Performance}
As digital twin systems expand from single assets to interconnected networks spanning factories, supply chains, or cities, computational challenges grow exponentially~\cite{qi2021enabling}. Hierarchical and federated architectures are needed to orchestrate distributed twins while ensuring data consistency and minimizing communication overhead. Neural surrogate models provide speedups for simulation, but must balance computational efficiency and physical accuracy. Real-time adaptation to physical system changes, including degradation and failure, remains difficult. Edge-deployed models require efficient compression techniques that preserve physical consistency while meeting latency constraints~\cite{grieves2022foundations}.

\header{Trustworthiness and Ethics}
Digital twins increasingly inform safety-critical decisions, demanding rigorous assurance of transparency, robustness, and fairness~\cite{zheng2022digital}. Explainability should extend beyond feature attribution toward causal and physically grounded interpretations. Counterfactual reasoning and uncertainty quantification can support “what-if” scenario analysis, improving diagnostic reliability. Robustness to adversarial perturbations and cascading feedback failures remains an open problem. Formal verification techniques adapted from control theory could help certify system safety. Ethical governance frameworks that define accountability, auditability, and privacy protection are essential to ensure trust and societal acceptance~\cite{he2021digital}.

\header{Human-AI Collaboration}
Human oversight remains indispensable in digital twin ecosystems, especially for complex, uncertain, or high-stakes contexts. Effective collaboration requires intuitive interfaces where natural language queries are translated into simulation or optimization actions. Immersive visualization (VR/AR) and multimodal communication can make high-dimensional predictions interpretable to operators. Human-in-the-loop reinforcement learning and shared autonomy frameworks can integrate expert feedback dynamically, enabling systems that learn user preferences and calibrate trust levels through transparent uncertainty communication.

\header{Standardization and Cross-Domain Transfer}
The absence of universal data models, ontologies, and communication protocols limits interoperability between digital twin platforms~\cite{tao2022digital}. Community-wide standardization—led by efforts like the Digital Twin Consortium—is necessary to promote modular, interoperable architectures. Transfer learning offers an opportunity to reuse knowledge from mature industrial domains to accelerate deployment in emerging areas such as synthetic biology. Benchmark datasets covering diverse operational and failure modes, together with unified evaluation metrics balancing accuracy, robustness, and interpretability, are vital for reproducible research.

\header{Emerging Frontiers}
New frontiers for digital twins extend into global and human-scale systems. Climate, epidemic, and socio-economic digital twins demand integration of physical models with human behavioral and policy dimensions~\cite{wang2023multiagent}. Whole-human twins and quantum-level modeling represent long-term goals requiring breakthroughs in multi-scale data integration, biological modeling, and computational efficiency. Extending digital twin concepts to social systems raises profound questions about human agency, reflexivity, and ethics, underscoring the need for interdisciplinary collaboration among AI researchers, social scientists, and ethicists.

The convergence of AI and digital twin technologies is reshaping how we perceive, predict, and manage complex physical systems. Realizing this potential will require bridging physics and AI, scaling architectures for real-time autonomy, ensuring ethical governance, and fostering human-centered collaboration. We envision a future in which AI-powered digital twins serve as cognitive infrastructures—monitoring, reasoning, and adapting across scales from molecular to planetary—driving a new era of intelligent, sustainable, and trustworthy system management.

\section{Acknowledgement}
This work was in part supported by Lehigh University's CORE and RIG grants. In addition, we gratefully acknowledge Ruoxi Chen for her assistance in the early stages of figure preparation and for insightful discussions during her time as a visiting student at Lehigh University.
% This work was in part supported by NIH (RF1AG077820), NSF (IIS-2319451, IIS-2306791, MRI-2215789, CNS-2415209, CNS-2319343), DOE (DE-SC0025801), and Lehigh University (CORE and RIG).

\bibliographystyle{ieeetr}
\bibliography{reference}

\end{document}